%% file: acl_latex.tex
\documentclass[11pt]{article}

\usepackage[preprint]{acl}

\usepackage{times}
\usepackage{latexsym}
\usepackage{enumitem}
\usepackage{multicol}
\usepackage{multirow}
\usepackage{xspace}

\usepackage[T1]{fontenc}

\usepackage[utf8]{inputenc}

\usepackage{microtype}

\usepackage{inconsolata}

\usepackage{graphicx}
\usepackage{amsmath, amssymb}
\usepackage{xcolor}
\usepackage{array}
\usepackage{tabularx}
\usepackage{booktabs,makecell} 
\usepackage{tcolorbox}
\usepackage[table]{xcolor}

\newcolumntype{Z}[1]{>{\centering\arraybackslash}m{#1}}
\newcolumntype{Y}[1]{>{\raggedright\arraybackslash}m{#1}}
\newcolumntype{R}[1]{>{\centering\arraybackslash}m{#1}}
\newcommand{\imgbox}[1]{%
  \fbox{\includegraphics[width=0.98\linewidth]{#1}}%
}
\newcommand{\rubriccell}[3]{%
  \scriptsize
  \textbf{#1} (\texttt{#2}).\hspace{0.3em}#3\par
}

\newcommand{\ours}{RubSE\xspace}

\title{Rubrics as Visual-Repair Context for Self-Evolving UI-to-Code Generation}

\author{
\textbf{Tianyi Xiong}\textsuperscript{\textbf{1}}\thanks{%
\begin{tabular}[t]{@{}l@{}}
Work done during an internship at Microsoft.\\
Project: \href{https://tyxiong23.github.io/rubse}{https://tyxiong23.github.io/rubse}
\end{tabular}%
},
\textbf{Zhengyuan Yang}\textsuperscript{\textbf{2}},
\textbf{Xiaofei Wang}\textsuperscript{\textbf{2}},
\textbf{Chung-Ching Lin}\textsuperscript{\textbf{2}},
\textbf{Ruichun Ma}\textsuperscript{\textbf{2}}, \\
\textbf{Kevin Lin}\textsuperscript{\textbf{2}},
\textbf{Zhendong Wang}\textsuperscript{\textbf{2}},
\textbf{Linjie Li}\textsuperscript{\textbf{2}},
\textbf{Chenxi Liu}\textsuperscript{\textbf{1}},
\textbf{Ruibo Chen}\textsuperscript{\textbf{1}}, \\
\textbf{Ramani Duraiswami}\textsuperscript{\textbf{1}},
\textbf{Heng Huang}\textsuperscript{\textbf{1}},
\textbf{Lijuan Wang}\textsuperscript{\textbf{2}} \\[5pt]
\textsuperscript{\textbf{1}}University of Maryland, College Park,
\textsuperscript{\textbf{2}}Microsoft \\
\texttt{txiong23@umd.edu}
}

\begin{document}
\maketitle

\input{latex/sections/0_abstract}

\input{latex/sections/1_intro}

\input{latex/sections/2_related_work}
%
\input{latex/sections/3_method}

\input{latex/sections/4_experiments}

\input{latex/sections/5_analysis}
\input{latex/sections/6_conclusion}
\input{latex/sections/7_limitation}

\bibliography{custom}

\appendix
\input{latex/sections/x_appendix}

\end{document}

%% file: latex/sections/0_abstract.tex
\begin{abstract}

Large vision-language models have shown strong progress in UI-to-code generation, yet their test-time self-evolution remains unstable. We first identify a fundamental obstacle, termed \emph{visual repair coupling}: a local code edit may propagate through layout, style, and component dependencies, correcting one visual mismatch while degrading regions that were previously faithful. To address this issue, we present \ours, a \textbf{Rub}ric-guided \textbf{S}elf-\textbf{E}volution framework that uses rubrics to represent visual feedback as a structured visual-repair context. At each refinement round, \ours generates typed candidate rubrics, selects one prioritized repair target, and stores previously selected rubrics as history, thereby steering each revision toward a well-scoped visual repair while discouraging repeated or over-broad changes. Evaluations across six VLMs and three UI-to-code benchmarks demonstrate that \ours substantially outperforms naïve self-evolution in final-round and best-round settings, achieving more stable refinement trajectories and a higher trajectory-level performance ceiling. Further analysis shows that \ours mitigates trajectory collapse by improving recovery from severe visual regressions, and that stronger rubric generators can transfer effective visual-repair guidance to weaker code improvers.

\end{abstract}

%% file: latex/sections/1_intro.tex
\section{Introduction}
\label{sec:intro}

Recent progress in large vision-language models~\citep{hurst2024gpt,chen2024internvl,bai2025qwen3,chen2026eagle,li2024llava} has expanded their role from answering questions~\cite{liu2023visual,chen2023gpt,wang2025llava,xiong2025llava} about images to producing structured, executable outputs grounded in visual inputs~\cite{xiong2026phycritic,azzolini2025cosmos,ni2025viscoder2,zhao2025pyvision,yu2025minicpm}. UI-to-code generation~\cite{beltramelli2018pix2code,jiang2025screencoder,chen2025designcoder,zhou2025declarui} represents a concrete instance of this shift: given a screenshot, the model must generate web code that renders into a visually faithful reproduction of the target interface. This capability supports UI reproduction, design prototyping, and developer workflows by converting visual designs into executable implementation drafts. Despite its value, UI-to-code generation remains challenging because the quality of the generated code cannot be reliably judged from the code tokens alone. Instead, it must be rendered and evaluated against the target screenshot, requiring the model to align the resulting interface across multiple visual aspects~\citep{si2025design2code}, including spatial layout, color, typography, component hierarchy, and visual density.

\begin{figure}[!t]
    \centering
    \includegraphics[width=1\linewidth]{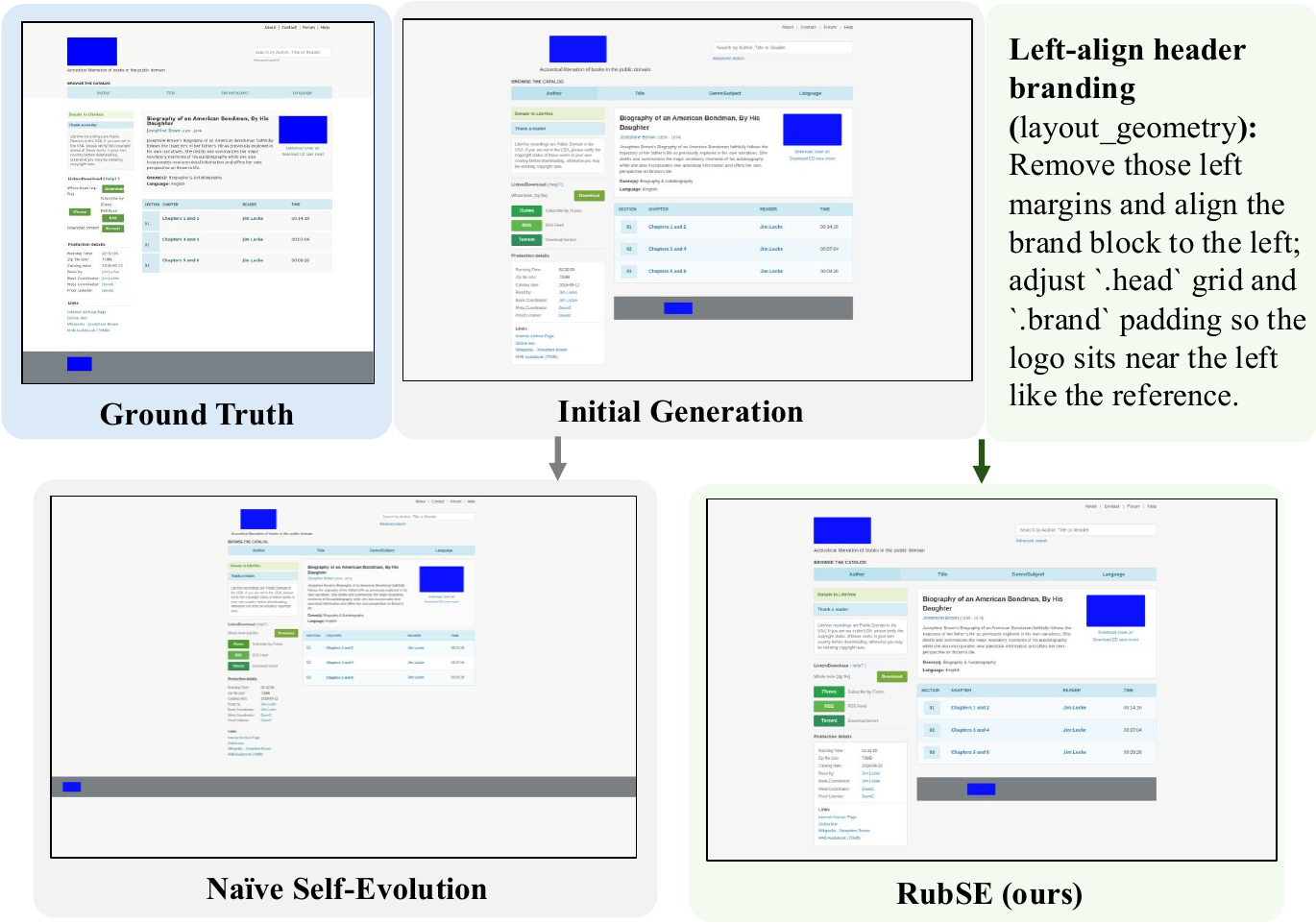}
    \caption{Example of visual repair coupling. Naïve self-evolution fixes the gray bar assignment, but introduces errors in page scale and bar length. In contrast, \ours correctly left-aligns the logo and tagline with the content edge while preserving the global layout.}
    \vspace{-5mm}
    \label{fig:teaser}
\end{figure}

Existing methods have primarily improved performance through training-time interventions, such as scaling instruction-tuning data~\citep{gui2025webcode2m,yun2024web2code,calo2026webui} or adopting stage-wise training pipelines~\citep{ni2025viscoder2,yang2025ui2code,zhao2025vincicoder,xu2025webvia}. In comparison, test-time improvement remains less explored. This gap is notable because UI-to-code provides a natural source of test-time feedback: after a code draft is generated, it can be rendered and visually compared with the target screenshot. A straightforward strategy is therefore self-evolution, where the model iteratively critiques the rendered output and revises the code based on observed discrepancies. Such generate--critique--revise loops have been widely studied for text-generation tasks~\citep{huang2023large,gou2024critic,khattab2023dspy,zhang2025agentic,zhang2025critic,liu2026prepare}, and are increasingly supported by general-purpose coding agents~\citep{claude_code,openai_codex} (see Appendix~\ref{appendix:addition_discussion} for discussion). However, in UI-to-code generation, naïve self-evolution does not reliably produce monotonic improvement.

The key obstacle is that UI code is visually coupled. Changing one CSS rule may shift the layout; modifying a parent container may affect its children; and adding a missing component may alter spacing, alignment, or visual density elsewhere. 
We refer to this phenomenon as \textbf{visual repair coupling}: code edits can induce non-local changes in the rendered interface through coupled layout, style, and component dependencies. 
Although we study this phenomenon in web UI coding, visual repair coupling reflects a broader challenge in structure--detail generation and agentic tasks, where local edits to implementation details can propagate through the global structure of the final artifact. 
%
As illustrated in Figure~\ref{fig:teaser}, naïve self-evolution is vulnerable to this coupling. An attempted repair may correct one visible error while damaging regions that were already faithful. Our preliminary analysis suggests that this leads to unstable refinement trajectories, where later iterations do not consistently improve over the initial draft and may even reduce overall visual fidelity.

This failure mode reveals a limitation of free-form visual feedback. A useful test-time critic should not merely tell the model what is wrong; it should also help constrain the repair so that correct regions are not unnecessarily disturbed. In other words, UI-to-code self-evolution must control both sides of visual repair: what should be changed and what should remain preserved. This requires feedback that is structured, localized, and persistent across iterations, rather than an unconstrained critique that may encourage broad code rewrites.

To this end, we propose \ours, a rubric-guided self-evolution framework for UI-to-code generation. The central idea is to represent visual feedback as structured repair context. Each rubric describes a specific visual failure, assigns it to a predefined visual aspect, and specifies a targeted correction direction. 
Compared with free-form critique, this rubric-based representation provides soft control over the refinement process: it makes the repair objective explicit while allowing the code generator to determine how to implement the edit.
To construct effective repair context across iterations, \ours uses three atomic operations: \textsc{Evolve}, \textsc{Select}, and \textsc{History}. \textsc{Evolve} generates candidate rubrics that describe possible visual repairs for the current rendering. \textsc{Select} chooses a single prioritized rubric as the focus of the next revision, reducing the risk of broad, entangled edits. \textsc{History} carries selected rubrics forward across iterations, reminding the model of previous repair decisions and discouraging repeated failures or regressions. Together, these operations separate three roles that are often conflated in naïve self-evolution—discovering errors, choosing a focused repair target, and preserving repair context over time—retaining the flexibility of open-ended code revision while steering the model toward more localized and stable visual improvements.

Across six models and three benchmarks, \ours improves both metrics over naïve self-evolution in 15/18 final-round settings, with average gains of +1.20 overall points and +0.11 aspect-mean score. It also achieves stronger best-round performance in 14/18 settings, with an average gain of +1.13 overall points, indicating both more stable refinement and a higher trajectory-level ceiling. 
Additional analysis reveals two key findings: (1) Rubric-guided repair stabilizes self-evolution by accelerating recovery from severe visual regressions and reducing collapse for stronger VLM executors. (2) High-quality rubrics transfer to weaker code improvers, as GPT-5.4-generated rubrics consistently improve Qwen self-evolution by providing perceptual, actionable, and well-scoped repair targets.

Our contributions are summarized as follows:
\begin{itemize}[leftmargin=1.4em,itemsep=1pt,topsep=2pt]
    \item We identify \emph{visual repair coupling}, where local code edits induce non-local rendered changes, as a central failure mode behind unstable UI-to-code self-evolution.

    \item We propose \ours, a rubric-guided self-evolution framework that turns visual feedback into structured visual-repair context through an \textsc{Evolve}--\textsc{Select}--\textsc{History} loop over typed rubrics, steering the model toward targeted and stable visual improvements.

    \item Experiments demonstrate that \ours improves both final-round stability and best-round performance over naïve self-evolution across benchmarks, execution models and metrics. Further analyses show that rubrics help recover from severe visual regressions and provide transferable guidance to weaker code improvers.
\end{itemize}

%% file: latex/sections/2_related_work.tex
\section{Related Work}
\label{sec:related_works}


%

\paragraph{UI-to-Code Generation.} Early task-specific methods train encoder--decoder architectures to translate GUI screenshots into intermediate UI descriptions~\citep{beltramelli2018pix2code} or HTML code~\citep{soselia2023learning}. Recent advances in vision-language models have enabled more flexible generation from visual inputs~\citep{ge2025advancing,jiang2025screencoder,zhang2025artifactsbench,liu2025modality,ni2025viscoder}.
A line of work constructs large-scale benchmarks from real-world or synthetic webpages. Design2Code~\citep{si2025design2code} introduces the first real-world benchmark and designs matching- and model-based evaluation metrics. Subsequent works extend this direction through improved data curation pipelines~\cite{yun2024web2code}, broader programming languages~\cite{ge2025advancing}, and fine-grained generation subtasks~\cite{lin2025webuibench}. 

Another line focuses on improving UI-to-code models. \citet{bhathal2025websight,gui2025webcode2m,jiang2025screencoder,calo2026webui} construct high-quality image--code pairs as instruction-tuning datasets. Method-wise, prior research filters self-generated UI programs using compiler and multimodal feedback~\citep{wu2024uicoder}, decomposes generation into hierarchy prediction and code synthesis~\citep{gui2025uicopilot}, and performs layout-aware reasoning and code assembly for better layout preservation~\citep{gui2025latcoder}. Recent studies further adapt reinforcement learning with verifiable rewards (RLVR, ~\citet{guo2025deepseek}) to UI-to-code generation task, by applying relative visual-quality signals as reference-based rewards~\citep{yang2025ui2code,deng2026visrefiner}. \citet{ni2025viscoder2} introduces an iterative self-debugging loop to improve generation quality, yet the role of structured context for self-improvement remains less explored. Our work studies what context enables test-time self-evolution, guiding targeted fixes while preserving overall visual fidelity.

\paragraph{Rubrics in LLM/VLMs.}
Rubrics have been widely adopted as structured evaluation criteria for LLM-as-a-Judge systems~\cite{zheng2023judging,zhang2023gpt}, particularly in open-ended domains where holistic reward signals are often insufficient, such as healthcare~\citep{arora2025healthbench} and academic tasks~\citep{akyurek2025prbench,yifei2025researchqa,wang2025profbench,xiong2025multi}. By decomposing evaluation into criterion-level decisions, rubrics improve the reliability of model-based judgment and better align automated scores with human evaluation~\citep{wang2025profbench,srivastava2025robust,liu2025openrubrics,wang2026unified}.
Rubrics have also been utilized as reward, extending RL post-training beyond domains with verifiable final answers. Rubric-based rewards provide fine-grained, instance-specific feedback, improving reasoning trajectories~\citep{yuan2025curing} and open-ended task performance~\citep{gunjal2025rubrics,shao2025dr,rezaei2025online}.
In contrast, our work treats rubrics as reusable visual repair context: they identify localized visual mismatches, guide subsequent code edits, and accumulate across refinement rounds to steer iterative UI revision toward the reference design.


%

%% file: latex/sections/3_method.tex
\begin{figure}[!t]
    \centering

    \includegraphics[width=0.99\linewidth]{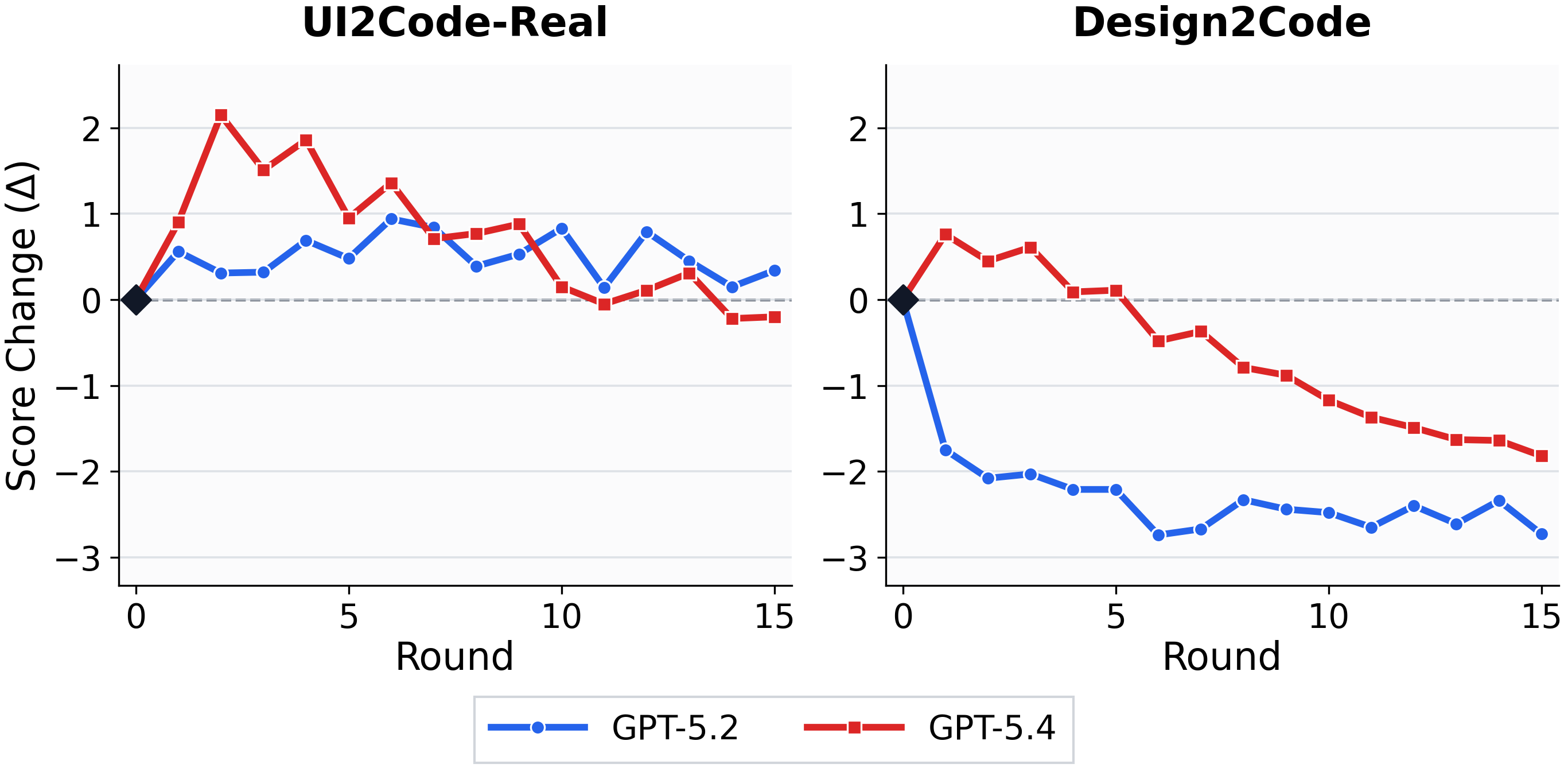}
    \vspace{-6mm}
    \caption{Score delta relative to the initial generation under naïve self-refinement. Both models show diminishing gains and later degradation; on Design2Code, refinement falls below the initial generation.}
    \vspace{-4mm}
    \label{fig:gpt-visual-repair-coupling}
\end{figure}

\begin{figure*}[!ht]
    \centering
    \includegraphics[width=1.0\linewidth]{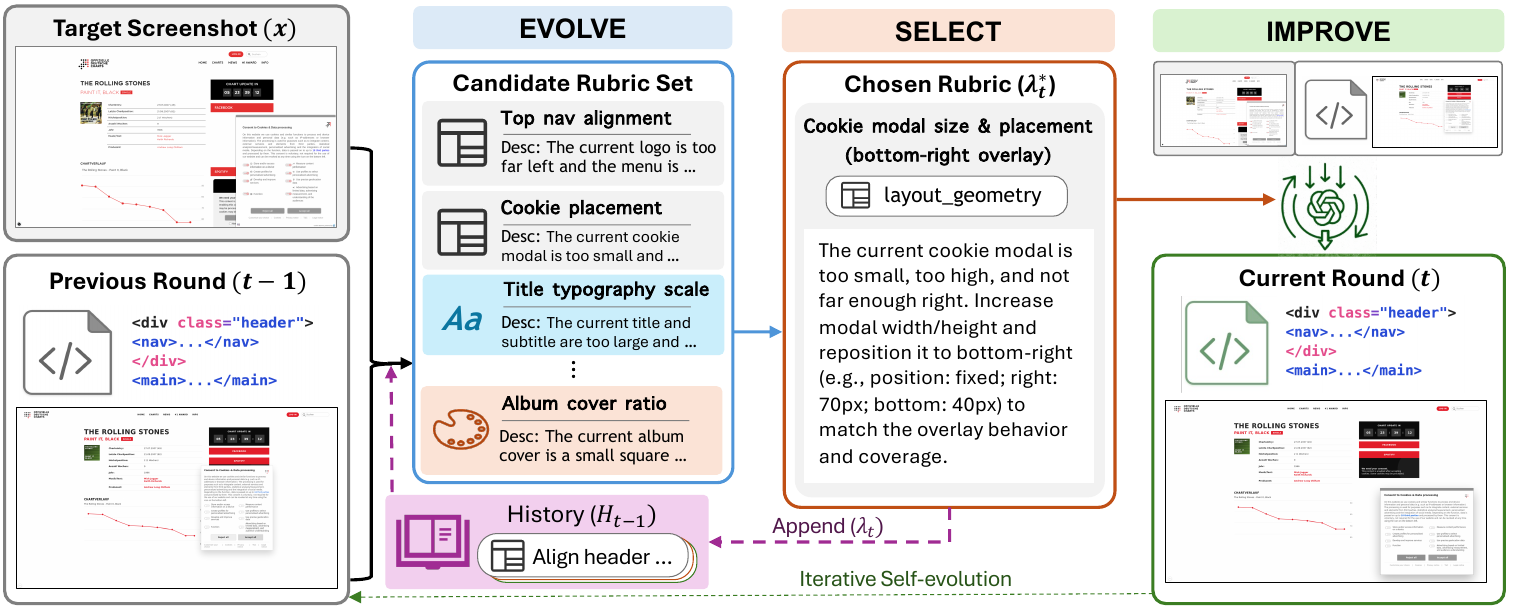}
    \vspace{-6mm}
    \caption{Self-evolution process of \ours. Given the target screenshot, the previous-round code and rendering, and the accumulated rubric history, \ours first generates a set of candidate rubrics for remaining visual mismatches. It then selects the most crucial rubric—position and scale of the cookie modal in this example—as a localized repair target, and uses it as context for VLM code revision. After each round, the selected rubric is appended to the history and reused in subsequent rounds, enabling iterative refinement while reducing repeated or uncontrolled edits.}
    \vspace{-4mm}
    \label{fig:main}
\end{figure*}

\section{Method}
\label{sec:setup}

\subsection{Task Formulation}
\label{sec:setup-formulation}

We formalize self-evolution for UI-to-code generation as sequential repair. Let $x$ denote the target UI screenshot, and let $R(\cdot)$ be a deterministic rendering function implemented by a headless browser. A VLM generator $G_\theta$ first produces an initial code draft $c_0 = G_\theta(x)$, whose rendering is $r_0 = R(c_0)$. At each refinement round $t \ge 1$, a feedback interface $\mathcal{I}$ constructs a repair context
\[
  \phi_t \;=\; \mathcal{I}\bigl(x,\, c_{t-1},\, r_{t-1},\, H_{t-1}\bigr),
\]
where $H_{t-1}$ summarizes an optional per-instance history of previous
repair steps. The generator then refines the previous code based on this context:
$c_t = G_\theta(x, c_{t-1}, r_{t-1}, \phi_t)$. The goal is a faithful
reconstruction $c_T$ such that $R(c_T)$ visually matches $x$. 
Different self-evolution methods differ in how they construct the repair context $\phi_t$; Section~\ref{sec:method-rubric} presents our rubric-based instantiation.

\subsection{Visual Repair Coupling}
\label{sec:method-coupling}

Let $\Delta(c, c')$ denote the textual difference between two code
states, and let $\Delta(R(c), R(c'))$ denote the rendered-pixel
difference between their renderings. In ordinary text generation, edits and outputs
are coextensive: $\Delta(c, c')$ \emph{is} the change. In visual code
generation, the rendering map $R$ is \emph{non-local}: editing line $i$
of $c$ can change various aspects of $R(c)$ through cascading layout,
flex/grid sizing, font metric, and asset placement effects. We call this property \emph{visual repair coupling}.

In a preliminary study, we apply GPT-5.2 and GPT-5.4 to iteratively refine their generated code to match the target screenshot for $N=15$ rounds. Results are shown in Figure~\ref{fig:gpt-visual-repair-coupling}. 
On UI2Code-Real, early gains are small or transient: GPT-5.2 reaches a modest peak improvement of $+0.94$ judge points and finishes at $+0.34$, while GPT-5.4 peaks higher at $+2.15$ by round~2 but then declines below the initial generation ($-0.20$). The failure mode is stronger on Design2Code: all $15/15$ GPT-5.2 refined rounds are below round~0, ending at $-2.73$, while GPT-5.4 shows only a short-lived $+0.76$ gain in the first round before declining to $-1.82$, with $10/15$ rounds below the initial generation.
A representative example in Figure~\ref{fig:teaser} illustrates this drift: the model attempts to fix the parent assignment of the bottom gray bar, but instead introduces additional errors in page scale, bar length, and the background color of the bottom-left container, turning a repair attempt into a broader visual degradation.

\subsection{\ours}
\label{sec:method-rubric}

The results above suggest that effective self-evolution requires a structured visual-repair context. Such context should not only identify what is wrong, but also constrain the scope of the next edit, so that a local repair does not unnecessarily perturb already faithful regions. 

In this section, we introduce \ours, which instantiates each round's repair context with a single-step visual rubric. Rather than imposing hard constraints on the generated code, \ours takes rubrics as explicit conditioning signals. Each rubric names a targeted visual mismatch, specifies the intended correction, and focuses the current refinement round on a single prioritized repair target, thereby allowing the generator to revise the code flexibly yet with localized guidance.

Formally, we define a visual-repair rubric as a structured object
\[
  \lambda \;=\; \langle\,\texttt{title},\,\texttt{type},\,\texttt{description}\,\rangle,
\]
where \texttt{description} specifies what is currently mismatched and
how it should be corrected, and \texttt{type} is drawn from a fixed
five-aspect taxonomy:
\textsc{layout\_geometry},
\textsc{spacing\_density},
\textsc{typography\_text},
\textsc{styling\_visual},
\textsc{completeness}.
The taxonomy is enforced during candidate generation so that the
candidate set spans visually distinct dimensions rather than collapsing
onto the most salient defect (e.g.\ a single bright colour mismatch).
Items with malformed JSON or invalid \texttt{type} are discarded.

We organize the feedback interface $\mathcal{I}$ around three atomic operations on rubrics to construct the repair context $\phi_t$, where the first two are implemented as separate VLM calls:

\begin{description}[leftmargin=1.4em,itemsep=3pt,topsep=2pt]
\item[\textsc{Evolve}\,$(x, r_{t-1}, c_{t-1}, H_{t-1}) \to \Lambda_t$.]
  Produces a typed candidate set of $K$ rubrics for the current state.
  When the cross-round history $H_{t-1}$ is non-empty, it is passed as
  an explicit avoid list, and \textsc{Evolve} is instructed to return
  only rubrics whose issues are \emph{not} already covered.

\item[\textsc{Select}\,$(x, r_{t-1}, c_{t-1}, \Lambda_t) \to \lambda_t^{\star}$.]
  Selects a single rubric from $\Lambda_t$ as $\lambda_t^{\star}$, choosing the one expected to maximize visual improvement under a targeted edit. The selected rubric defines the current repair context,
  i.e., $\phi_t = \lambda_t^{\star}$.

\item[\textsc{History}\,$(H_{t-1}, \lambda_t^{\star}) \to H_t = H_{t-1} \cup \{\lambda_t^{\star}\}$.]
  Appends the selected rubric to a per-instance, append-only history,
  consumed by the next round's \textsc{Evolve} as the avoid list.
\end{description}

\input{latex/sections/tables/main_result}

Given this scoped visual-repair context, the generator emits the next HTML/CSS code conditioned on the target screenshot, the previous state $(c_{t-1}, r_{t-1})$, and the selected rubric:
\[
  c_t = G_\theta(x, c_{t-1}, r_{t-1}, \phi_t),
  \quad \text{where } \phi_t = \lambda_t^{\star}.
\]

As illustrated in Figure~\ref{fig:main}, each round first obtains a selected rubric through $\textsc{Evolve} \rightarrow \textsc{Select}$, then uses it to refine the code and appends it to the history via \textsc{History}. Detailed evolution prompts are provided in Appendix~\ref{appendix:prompt-for-rubse}.
Keeping \textsc{Evolve} and \textsc{Select} as distinct
steps is intentional: the former externalizes a typed set of candidate
repair targets, while the latter prioritizes one of them under a
constrained edit budget, so that prioritization is not entangled with
candidate generation in a single pass. This separation design is validated in Appendix~\ref{appendix:computational-cost}.

%% file: latex/sections/tables/main_result.tex
\definecolor{table-highlight}{HTML}{DDEEFF}  

\begin{table*}[!ht]
  \centering
  \scalebox{0.94}{
  \setlength{\tabcolsep}{2.5pt}
  \renewcommand{\arraystretch}{0.98}
  \begin{tabular}{ll|cc|cc|cc}
    \hline
    \multirow{2}{*}{\textbf{Model}} & \multirow{2}{*}{\textbf{Method}} & \multicolumn{2}{c|}{\textbf{UI2Code-Real}} & \multicolumn{2}{c|}{\textbf{Design2Code}} & \multicolumn{2}{c}{\textbf{Design2Code-HARD}} \\
    & & Overall $\uparrow$ & Aspect $\uparrow$ & Overall $\uparrow$ & Aspect $\uparrow$ & Overall $\uparrow$ & Aspect $\uparrow$ \\
    \hline
    \multirow{5}{*}{GPT-5.4} & Direct & 85.4 & 5.32 & 86.9 & 5.63 & 79.8 & 4.72 \\
                      & + Self-Evolve ($r{=}10$) & 85.6 & 5.43 & 85.8 & 5.50 & 78.0 & 4.61 \\
                      & + Self-Evolve (best) & \hphantom{\textsubscript{r2}}87.6\textsubscript{r2} & 5.61 & \hphantom{\textsubscript{r1}}87.7\textsubscript{r1} & 5.68 & \hphantom{\textsubscript{r1}}80.0\textsubscript{r1} & \underline{4.77} \\
                      \rowcolor{white} \cellcolor{white} & + \ours ($r{=}10$) & \underline{87.9} & \textbf{5.66} & \underline{87.8} & \underline{5.76} & \underline{80.3} & 4.74 \\
                      \rowcolor{white} \cellcolor{white}  & + \ours (best) & \hphantom{\textsubscript{r8}}\textbf{88.1}\textsubscript{r8} & \underline{5.63} & \hphantom{\textsubscript{r3}}\textbf{88.2}\textsubscript{r3} & \textbf{5.79} & \hphantom{\textsubscript{r4}}\textbf{80.8}\textsubscript{r4} & \textbf{4.79} \\
    \hline
    \multirow{5}{*}{GPT-5.2} & Direct & 81.6 & 5.03 & 86.4 & \underline{5.59} & 82.7 & 5.03 \\
                      & + Self-Evolve ($r{=}10$) & 82.5 & 5.06 & 83.9 & 5.12 & 82.3 & 4.92 \\
                      & + Self-Evolve (best) & \hphantom{\textsubscript{r6}}82.6\textsubscript{r6} & 5.06 & \hphantom{\textsubscript{r1}}84.6\textsubscript{r1} & 5.20 & \hphantom{\textsubscript{r2}}82.8\textsubscript{r2} & 4.97 \\
                     \rowcolor{white} \cellcolor{white}  & + \ours ($r{=}10$) & \underline{85.0} & \underline{5.29} & \underline{87.2} & 5.43 & \underline{84.4} & \underline{5.13} \\
                     \rowcolor{white} \cellcolor{white}  & + \ours (best) & \hphantom{\textsubscript{r7}}\textbf{85.1}\textsubscript{r7} & \textbf{5.35} & \hphantom{\textsubscript{r2}}\textbf{88.1}\textsubscript{r2} & \textbf{5.71} & \hphantom{\textsubscript{r7}}\textbf{84.8}\textsubscript{r7} & \textbf{5.15} \\
    \hline
    \multirow{5}{*}{Claude-Sonnet-4.5} & Direct & 73.3 & 4.45 & 83.0 & 4.98 & 79.8 & 4.80 \\
                      & + Self-Evolve ($r{=}10$) & 75.5 & 4.70 & 83.3 & 5.04 & 80.1 & \textbf{4.87} \\
                      & + Self-Evolve (best) & \hphantom{\textsubscript{r3}}75.8\textsubscript{r3} & 4.69 & \hphantom{\textsubscript{r3}}\textbf{83.5}\textsubscript{r3} & 5.04 & \hphantom{\textsubscript{r8}}\textbf{81.2}\textsubscript{r8} & \underline{4.84} \\
                     \rowcolor{white} \cellcolor{white}  & + \ours ($r{=}10$) & \underline{77.6} & \textbf{4.78} & \textbf{83.5} & \textbf{5.05} & 80.4 & \underline{4.84} \\
                     \rowcolor{white} \cellcolor{white}  & + \ours (best) & \hphantom{\textsubscript{r7}}\textbf{79.1}\textsubscript{r7} & \underline{4.74} & \hphantom{\textsubscript{r10}}\textbf{83.5}\textsubscript{r10} & \textbf{5.05} & \hphantom{\textsubscript{r3}}\textbf{81.2}\textsubscript{r3} & 4.82 \\
    \hline
    \multirow{5}{*}{Qwen3-VL-32B} & Direct & 66.2 & 4.10 & 79.6 & 4.84 & 78.0 & 4.74 \\
                      & + Self-Evolve ($r{=}10$) & \textbf{68.6} & \textbf{4.23} & 81.3 & 4.95 & \textbf{84.4} & \underline{4.99} \\
                      & + Self-Evolve (best) & \hphantom{\textsubscript{r10}}\textbf{68.6}\textsubscript{r10} & \textbf{4.23} & \hphantom{\textsubscript{r7}}\underline{81.4}\textsubscript{r7} & 4.94 & \hphantom{\textsubscript{r10}}\textbf{84.4}\textsubscript{r10} & \underline{4.99} \\
                     \rowcolor{white} \cellcolor{white}  & + \ours ($r{=}10$) & 68.3 & 4.18 & \underline{81.4} & \underline{4.96} & 82.0 & \textbf{5.04} \\
                     \rowcolor{white} \cellcolor{white}  & + \ours (best) & \hphantom{\textsubscript{r5}}68.5\textsubscript{r5} & 4.19 & \hphantom{\textsubscript{r5}}\textbf{82.0}\textsubscript{r5} & \textbf{4.97} & \hphantom{\textsubscript{r1}}82.8\textsubscript{r1} & \underline{4.99} \\
    \hline
    \multirow{5}{*}{Qwen3.5-9B} & Direct & 67.9 & 4.18 & 70.3 & 4.27 & 72.0 & 4.36 \\
                      & + Self-Evolve ($r{=}10$) & 69.8 & 4.28 & 73.6 & 4.47 & 76.0 & 4.53 \\
                      & + Self-Evolve (best) & \hphantom{\textsubscript{r10}}69.8\textsubscript{r10} & 4.28 & \hphantom{\textsubscript{r7}}74.3\textsubscript{r7} & 4.47 & \hphantom{\textsubscript{r10}}76.0\textsubscript{r10} & 4.53 \\
                    \rowcolor{white} \cellcolor{white}   & + \ours ($r{=}10$) & \underline{70.5} & \underline{4.34} & \underline{75.0} & \textbf{4.56} & \textbf{77.6} & \textbf{4.64} \\
                    \rowcolor{white} \cellcolor{white}   & + \ours (best) & \hphantom{\textsubscript{r9}}\textbf{70.9}\textsubscript{r9} & \textbf{4.35} & \hphantom{\textsubscript{r7}}\textbf{75.6}\textsubscript{r7} & \underline{4.55} & \hphantom{\textsubscript{r10}}\textbf{77.6}\textsubscript{r10} & \textbf{4.64} \\
    \hline
    \multirow{5}{*}{Qwen3.6-35B-A3B} & Direct & 73.3 & 4.52 & 80.6 & 4.95 & 81.3 & 4.93 \\
                      & + Self-Evolve ($r{=}10$) & 75.1 & 4.61 & 81.6 & 5.04 & 82.5 & 5.01 \\
                      & + Self-Evolve (best) & \hphantom{\textsubscript{r9}}75.5\textsubscript{r9} & 4.58 & \hphantom{\textsubscript{r8}}81.8\textsubscript{r8} & 5.04 & \hphantom{\textsubscript{r10}}82.5\textsubscript{r10} & 5.01 \\
                     \rowcolor{white} \cellcolor{white}   & + \ours ($r{=}10$) & \textbf{77.0} & \textbf{4.80} & \textbf{82.0} & \textbf{5.10} & \underline{83.6} & \underline{5.05} \\
                     \rowcolor{white} \cellcolor{white}   & + \ours (best) & \hphantom{\textsubscript{r10}}\textbf{77.0}\textsubscript{r10} & \textbf{4.80} & \hphantom{\textsubscript{r10}}\textbf{82.0}\textsubscript{r10} & \textbf{5.10} & \hphantom{\textsubscript{r7}}\textbf{85.2}\textsubscript{r7} & \textbf{5.14} \\
    \hline
  \end{tabular}
  \renewcommand{\arraystretch}{1.0}
  }
  \vspace{-1.5mm}
  \caption{Self-evolution results across models and  benchmarks. We report the overall judge score (0--100) and the mean aspect score (1--7 Likert). ``Direct'' denotes the initial code generation round. ``Self-Evolve'' denotes naïve self-evolution. Numbers are reported for the final round-10 result and the best intermediate round selected by the ``Overall'' score; the subscript indicates the selected round. For each model, the best result in each column is shown in \textbf{bold}, and the second-best is \underline{underlined}.
}
\vspace{-4mm}
\label{tab:main-results}
\end{table*}

%% file: latex/sections/4_experiments.tex
\section{Experiments}
\label{sec:experiments}

\subsection{Settings}
\label{sec:experiment-settings}

\paragraph{Implementation Details.} We evaluate \ours on two groups of VLMs: 
1) for frontier VLMs, we use GPT-5.4~\citep{openai2026gpt54}, GPT-5.2~\citep{openai2025gpt52}, and Claude-Sonnet-4.5~\citep{anthropic2025claudesonnet45}; 
2) for open-source models, we evaluate three leading Qwen~\cite{bai2025qwen3} variants: Qwen3-VL-32B-Instruct, Qwen3.5-9B, and Qwen-3.6-35B-A3B. 
Unless otherwise specified, all self-evolution methods are run for 10 refinement rounds.
For all models, we follow their default inference hyperparameters and set the maximum generation tokens to 32,768. Open-source-model experiments use vLLM v0.19.0 and four NVIDIA H200 GPUs.

\paragraph{Benchmarks.}
Design2Code~\citep{si2025design2code} is a widely adopted UI-to-code benchmark containing webpages with diverse structures and layouts, where external images are replaced with placeholders to focus evaluation on reproducible HTML/CSS elements. We evaluate on both the normal (484 samples) and hard (80 samples) subsets. UI2Code-Real~\citep{yang2025ui2code} contains 115 real-world webpages collected from in-the-wild sources, allowing us to assess performance under more challenging and realistic UI complexity.

\paragraph{Evaluation Protocol.} Following~\citet{yang2025ui2code}, we evaluate visual fidelity by using a frontier VLM to compare each rendered webpage image against the target screenshot. We report two metrics: 1) an overall score on a 0--100 scale, and 2) an aspect-level score by averaging 1--7 Likert-scale ratings over five predefined visual aspects. In experiment, we apply GPT-5.2 as the judge model and, for each generated rendering and metric, average over three independent judge runs. Additional protocol details are provided in Appendix~\ref{appendix:evaluation-protocol}, and judge variability
is analyzed in Appendix~\ref{appendix:judge_variability}. %

\subsection{Main Results}
\label{subsec:main-results-findings}

We compare \ours with direct initial code generation and naïve self-evolution. For iterative methods, we report both the final-round performance and the best round selected by the \textit{overall} score, with the aspect score taken from the same round, as an oracle-performance reference. Results are summarized in Table~\ref{tab:main-results}.

\paragraph{\ours stabilizes self-evolution and improves code generation fidelity.} At round 10, \ours improves over naïve self-evolution by an average of $+1.20$ overall score and $+0.11$ aspect-mean score across the 18 model--benchmark settings. The effect is clearest on frontier executors ($+1.90$ judge, $+0.16$ aspect), where unconstrained revision often loses early gains. Open-source executors also benefit: averaged over the 9 Qwen settings, \ours remains $+0.50$ overall score and $+0.06$ aspect score higher. At the selected best rounds, \ours outperforms naïve self-evolution on both metrics in $14/18$ settings. This indicates that \ours improves fixed-budget stability while also providing a higher upper bound for iterative self-evolution across different models and benchmarks. A pairwise human evaluation on 60 UI2Code-Real samples provides complementary evidence that these gains are perceptible: the round-10 outputs of \ours{} are preferred over those of naïve self-evolution for both GPT-5.2 and GPT-5.4 ($p<0.01$; Appendix~\ref{appendix:human_eval}).

\paragraph{\ours encourages long-tail iterative improvements on frontier models.}
On frontier VLMs, \ours reaches its best checkpoint substantially later than naïve self-evolution on average ($r{=}5.7$ vs.\ $r{=}3.0$), while also yielding a larger gain over direct generation ($+2.22$ vs.\ $+0.77$ overall points). This pattern supports the intended role of rubric history: each round can target a remaining long-tail visual mismatch while preserving previously repaired regions, allowing frontier models to continue exploring long-tail and localized refinements after free-form self-evolution has plateaued or drifted.

\paragraph{Open-source models show larger headroom but heterogeneous dynamics.} \ours improves over direct generation in all 9 Qwen settings, with an average best-round overall gain of $+3.60$ points versus $+2.79$ for naïve self-evolution. For Qwen3.5-9B and Qwen3.6-35B-A3B, \ours consistently excels in both final-round and best-round results across metrics and benchmarks. Qwen3-VL-32B-Instruct exhibits mixed results: naïve self-evolution achieves higher overall scores on UI2Code-Real and Design2Code-HARD, while \ours wins on Design2Code. As discussed in Appendix~\ref{appendix:qwen3vl-exception}, this behavior stems from a model-specific imbalance in rubric generation and selection: Qwen3-VL-32B-Instruct overemphasizes localized spacing-density adjustments while underweighting completeness issues. This suggests that typed rubrics are broadly useful for open-source executors, although their effectiveness is model-dependent.

\paragraph{Computational Cost.}
Compared with naïve self-evolution, \ours{} incurs 1.60$\times$ the API
cost with GPT-5.4 and 2.5\% higher inference latency with Qwen3.5-9B
under standard vLLM serving. A detailed cost breakdown is provided in
Appendix~\ref{appendix:computational-cost}.

\begin{figure}[!t]
    \centering
    \includegraphics[width=0.999\linewidth]{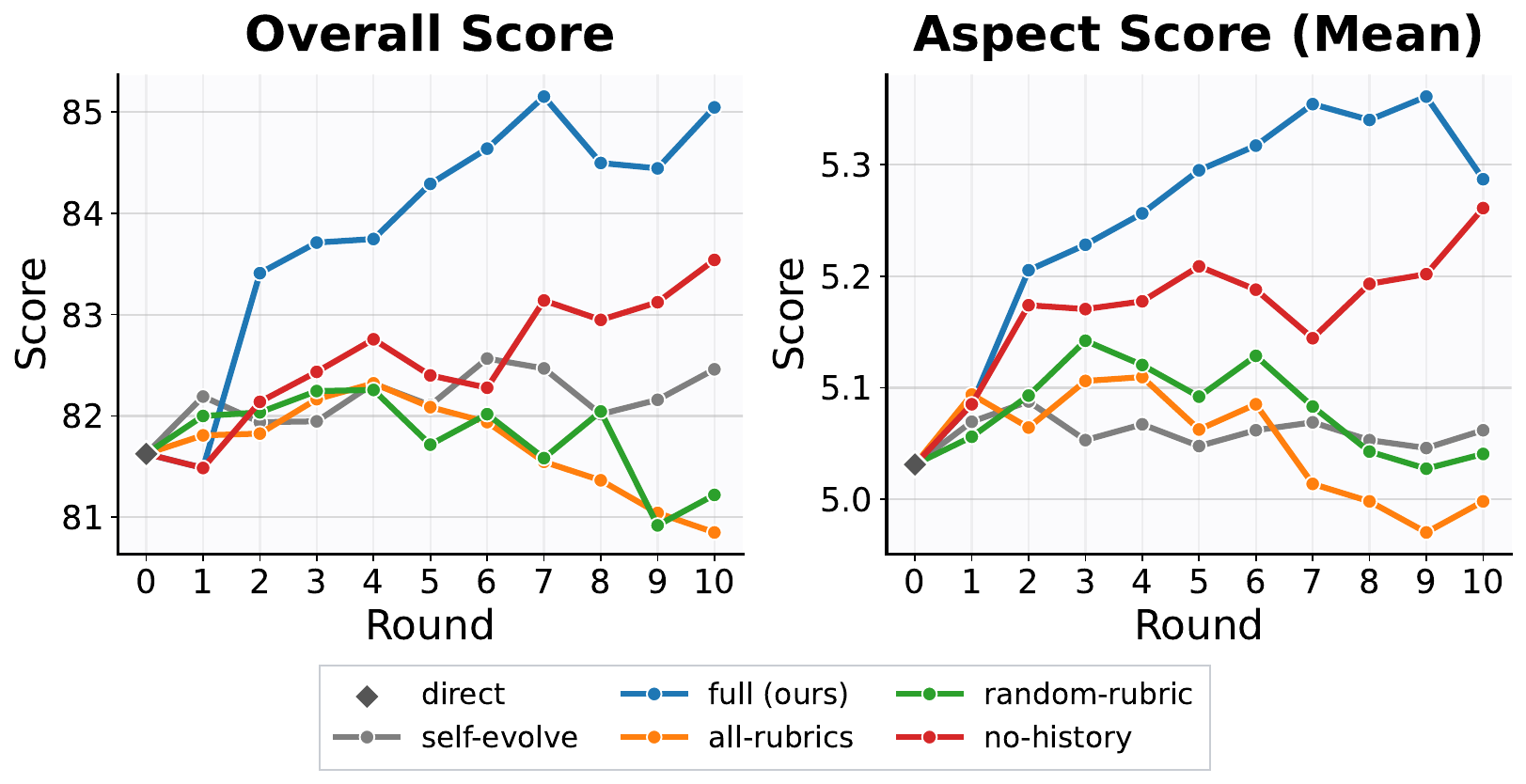}
    \vspace{-5.5mm}
    \caption{Ablation study on UI2Code-Real with GPT-5.2 as the self-evolution backbone. The full \ours achieves the overall strongest and most consistent gains across both evaluation metrics.}
    \vspace{-5mm}
    \label{fig:gpt52-ablation}
\end{figure}

%% file: latex/sections/5_analysis.tex
\begin{figure*}[!ht]
    \centering
    \includegraphics[width=0.999\linewidth]{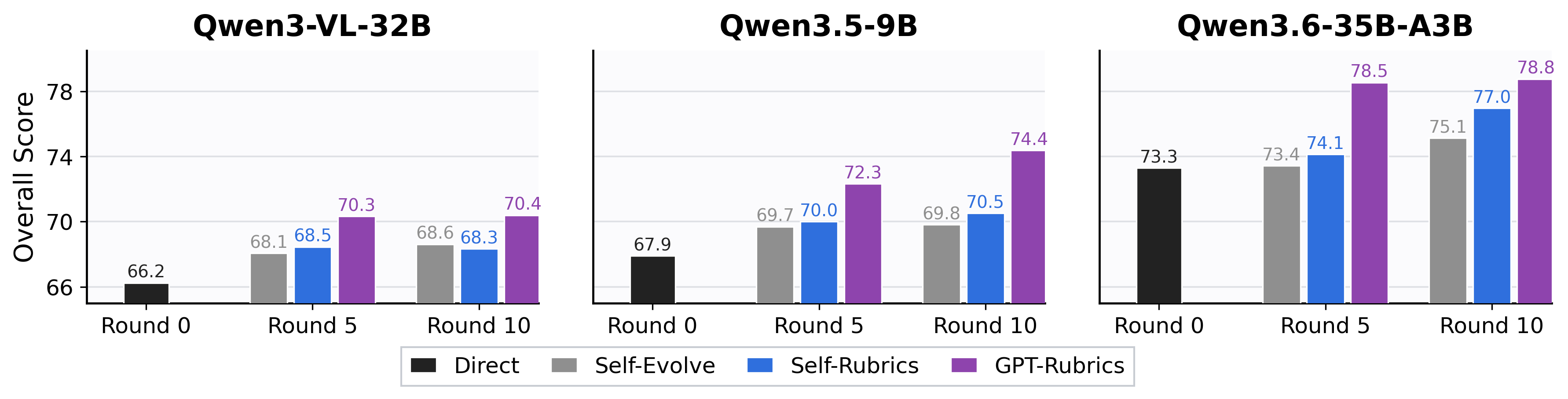}
    \vspace{-7.5mm}
    \caption{Results of applying GPT-generated rubrics for open-source VLM self-evolution. Across all three Qwen models, rubrics generated and selected by GPT-5.4 lead to stronger and more consistent gains than self-generated rubrics, particularly in early refinement rounds.}
    \label{fig:gpt-rubrics-ui2code}
    \vspace{-3.5mm}
\end{figure*}

\section{Analysis}
\label{sec:analysis}

\subsection{Ablation Study}
\label{sec:ablation}

We conduct an ablation study on UI2Code-Real using GPT-5.2 as the VLM for self-evolution. As shown in Figure~\ref{fig:gpt52-ablation}, the full method achieves the strongest and most consistent improvements across both overall and aspect-level metrics.
Using all generated rubrics or a randomly selected one yields modest gains over naïve self-evolution in early rounds, but the gains are unstable and may drop in later iterations, confirming that selection helps control the edit scope toward the most crucial yet localized visual issue. Removing rubric history leads to consistent but smaller improvements, indicating that past rubrics reduce repeated local repairs and encourage exploration across visual dimensions.

\begin{table}[t]
\centering
\small
\setlength{\tabcolsep}{2pt}
\vspace{-2mm}
\begin{tabular}{l|cc|cc}
\toprule
\renewcommand{\arraystretch}{0.95}
\multirow{2}{*}{Model} & \multicolumn{2}{c|}{Self-Evolve} & \multicolumn{2}{c}{\ours (ours)} \\
\cline{2-3}\cline{4-5}
 & Collapse $\downarrow$ & Recover $\uparrow$ & Collapse $\downarrow$ & Recover $\uparrow$ \\
\midrule
GPT-5.2 & 20.8\% & 26.2\% & 13.5\% & 24.2\% \\
GPT-5.4 & 17.9\% & 18.5\% & 11.2\% & 50.0\% \\
Claude-4.5 & 18.0\% & 17.3\% & 13.8\% & 24.1\% \\
\hline
Qwen3-VL & 11.5\% & 23.1\% & 15.3\% & 22.6\% \\
Qwen3.5 & 17.9\% & 21.2\% & 19.4\% & 23.5\% \\
Qwen3.6 & 14.9\% & 26.2\% & 19.6\% & 44.4\% \\
\bottomrule
\end{tabular}
\vspace{-1.5mm}
\caption{Trajectory collapse and recovery rates. Results are aggregated over all three benchmarks.}
\vspace{-5mm}

\label{tab:trajectory-any-collapse-recovery-by-model}
\end{table}

\begin{figure}[!ht]
    \centering

    \includegraphics[width=0.96\linewidth]{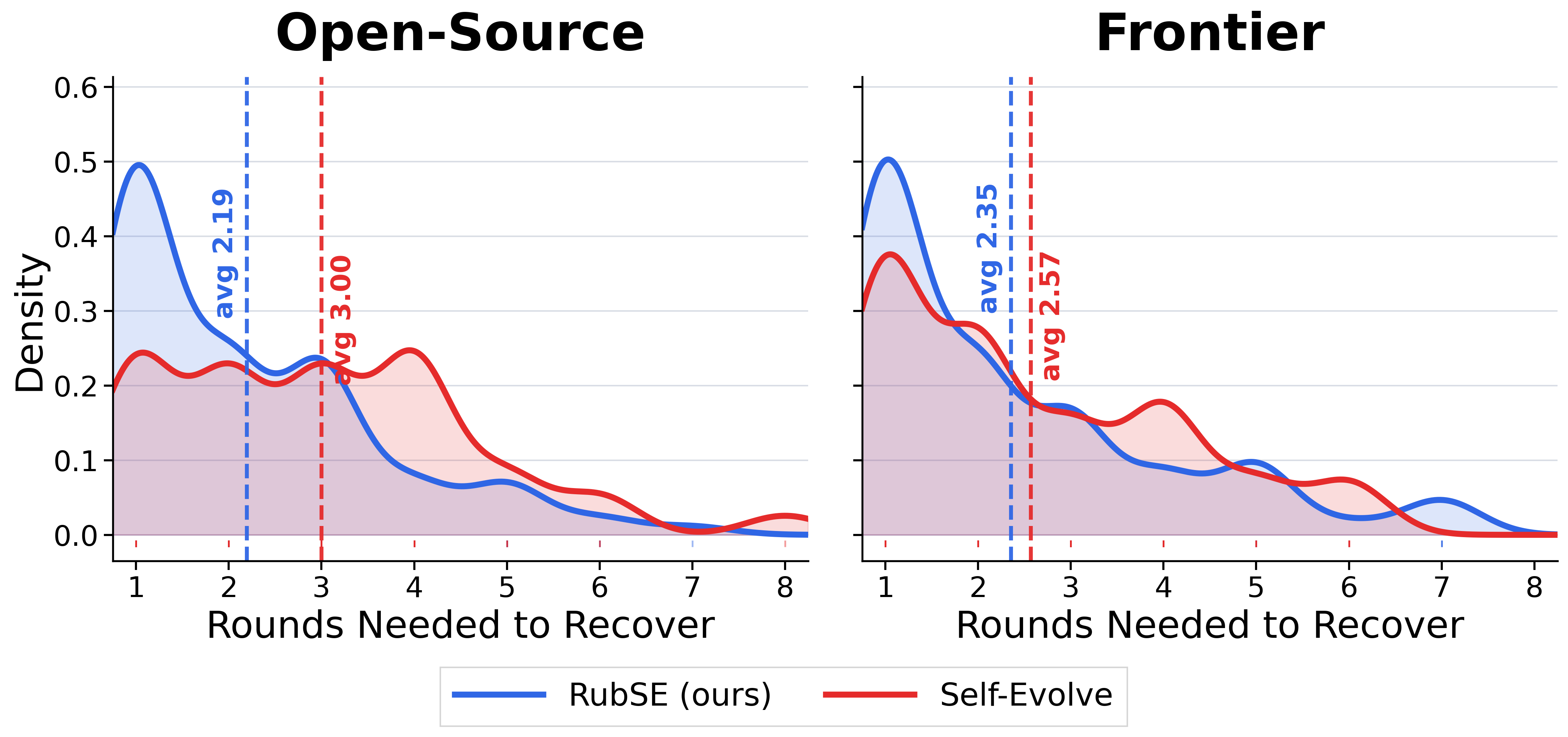}
    \vspace{-1.5mm}
    \caption{Density of rounds needed for recovery. Results are reported over all recovered trajectories. Dotted lines denote the average number of rounds required to recover from a collapse for each method.}
    \vspace{-5.5mm}
    \label{fig:collapse-recovery-rounds}
\end{figure}

\subsection{How \ours Mitigates Trajectory Collapse?}

We define a \emph{visual-repair collapse} as a consecutive-round regression where the overall score drops by at least 8 points and the aspect-mean score drops by at least 0.35. A recovery occurs when a later round returns close to the pre-collapse state, within 2 overall points and 0.10 aspect-mean. A trajectory is considered collapsed if it contains at least one collapse event. The recovery rate is computed over collapsed trajectories as the fraction that recover from all collapse events by the final round.

Results are reported in Table~\ref{tab:trajectory-any-collapse-recovery-by-model}. For frontier VLMs, 
\ours reduces collapse for all three models (18.9\% to 12.8\% on average) and improves recovery for two (20.7\% to 32.8\% on average), suggesting that rubrics help strong code executors avoid over-broad repairs and recover from severe errors. For Qwen executors, the effect is mixed: average collapse increases from 14.8\% to 18.1\%, while average recovery improves from 23.5\% to 30.2\%, with only Qwen3-VL-32B showing a slight decline.
Figure~\ref{fig:collapse-recovery-rounds} further shows the distribution of rounds needed to recover from each collapse: for both open-source and frontier VLMs, \ours concentrates in earlier rounds, requiring fewer recovery rounds on average.
Overall, rubrics provide a stabilizing visual-repair context: they broadly improve recovery rate and speed, and substantially reduce collapse when paired with frontier executors.

\subsection{Do Stronger Rubrics Transfer to Weaker Code Improvers?}
\label{sec:analysis-gpt-qwen-rubrics.}

To examine how rubric quality affects iterative self-evolution, we keep each Qwen model as the code improver but replace its self-generated rubrics with those generated and selected by GPT-5.4, the strongest VLM in our experiments. As shown in Figure~\ref{fig:gpt-rubrics-ui2code}, GPT-generated rubrics consistently outperform self-generated rubrics on UI2Code-Real across all three Qwen models, yielding stronger and more stable iterative gains. This suggests that high-quality rubrics can serve as model-agnostic visual-repair context: even when the code improver itself is weaker, better external guidance can effectively steer refinement toward more localized and reliable visual corrections.

\paragraph{GPT rubrics operate at a larger and more actionable visual-repair scale.}
We analyze self-generated and GPT-generated rubrics across three Qwen models and three benchmarks, with numerical results reported in Table~\ref{tab:rubric-content-gpt-vs-self} and qualitative examples shown in Table~\ref{tab:appendix-case-qwen-gpt-rubrics-examples}. Self-rubrics produced by Qwen models are typically more low-level and narrowly scoped: they often prescribe concrete patches, naming CSS properties, pixel values, colors, or individual widgets, as reflected by their higher use of CSS/HTML cues on Design2Code (70\% vs. 53\%) and numeric or color constants (35--46\% vs. 7\%). 
In contrast, GPT-generated rubrics abstract away from single CSS fixes and instead focus on target perceptual relations, such as layout, scale, hierarchy, and spatial relationships between regions. Quantitatively, they contain more global-structure cues across all datasets (89--91\% vs. 68--71\%) and more action-oriented revision language (77--82\% vs. 50--65\%). This difference in \emph{feedback granularity} helps explain Figure~\ref{fig:gpt-rubrics-ui2code}: GPT-generated rubrics transfer effectively to weaker Qwen code improvers, producing larger early gains by steering them beyond local CSS-level repair loops toward higher-level structural and actionable edits. Self-generated rubrics are still useful for Qwen models, but mainly through low-level edits that yield small yet consistent stepwise improvements.

\begin{table}[!t]
\centering
\small
\setlength{\tabcolsep}{1.5pt}
\
\scalebox{0.9}{%
\begin{tabular}{llrrrrrrr}
\toprule
Dataset & Rubrics & Tok. & Global & CSS & Numeric & Space & Style & Action \\
\midrule
\multirow{2}{*}{UI2C}
& Qwen & 56.5 & 71\% & 50\% & 35\% & 20\% & 40\% & 50\% \\
& GPT  & 57.3 & 89\% & 51\% & 7\%  & 30\% & 41\% & 77\% \\
\midrule
\multirow{2}{*}{D2C}
& Qwen & 56.9 & 68\% & 70\% & 43\% & 34\% & 42\% & 65\% \\
& GPT  & 54.8 & 91\% & 53\% & 7\%  & 34\% & 51\% & 80\% \\
\midrule
\multirow{2}{*}{D2C-H}
& Qwen & 53.4 & 70\% & 65\% & 46\% & 26\% & 47\% & 58\% \\
& GPT  & 54.6 & 90\% & 54\% & 7\%  & 35\% & 49\% & 82\% \\
\bottomrule
\end{tabular}%
}
\vspace{-1mm}
\caption{Content analysis of GPT- and Qwen-generated rubrics. ``Tok.'' denotes the average number of words per rubric. Values are the percentage of rubrics containing the corresponding type of visual repair.}
\vspace{-3mm}
\label{tab:rubric-content-gpt-vs-self}
\end{table}

%% file: latex/sections/6_conclusion.tex
\section{Conclusion}

In this paper, we introduced \ours, a rubric-guided self-evolution framework for UI-to-code generation that instantiates visual-repair context as structured rubrics. Each rubric defines a focused repair target, identifies the corresponding visual mismatch, and specifies a possible correction direction. To construct visual-repair context at each iteration round, \ours uses an \textsc{Evolve}--\textsc{Select}--\textsc{History} loop over rubrics: it explores candidate repair directions, selects one prioritized target, and carries previous rubrics forward to discourage repeated or over-broad revisions. This design preserves the flexibility of open-ended code revision while making iterative refinement more controlled and less prone to repeated repairs.
Experiments across frontier and open-source vision-language models show that \ours improves both final-round stability and best-round performance over naïve self-evolution. Further analyses show that rubric-guided repair helps recover from severe visual regressions, and that high-quality rubrics can transfer from stronger rubric generators to weaker code improvers. Overall, \ours provides a simple and practical mechanism to make test-time UI-to-code self-evolution more structured and reliable, paving the way for self-improving systems for complex generation tasks with coupled final artifacts.

%% file: latex/sections/7_limitation.tex
\section*{Limitations}

First, our experiments primarily evaluate three leading frontier models and three representative open-source variants on HTML/CSS webpage generation, leaving broader model families and UI languages underexplored. Second, \ours does not explicitly guarantee code correctness during self-evolution. As a result, some refinement steps may introduce syntax or runtime errors, leading to rendering failures. Incorporating an explicit debugging loop into self-evolution would be a promising direction. Third, although VLM-based evaluation has improved, its judgments are not perfectly stable and may diverge from human evaluations in a minority of cases, particularly when assessing subtle visual differences in later refinement rounds. Our human study (Appendix B.1) supports the main comparison on 60 UI2Code-Real samples for two models. Developing more robust visual graders remains an important direction for future work.

%% file: latex/sections/x_appendix.tex
\clearpage

\label{sec:appendix}

\input{latex/sections/x_evaluation_protocol}

\input{latex/sections/x_additional_experiments}

\input{latex/sections/x_additional_case}

\input{latex/sections/x_prompt_for_generation}
\input{latex/sections/x_additional_discussion}

%% file: latex/sections/x_evaluation_protocol.tex
\section{Evaluation Protocol}
\label{appendix:evaluation-protocol}

In experiments, we evaluate generated code quality by comparing the rendered image against the target image under the VLM-as-a-Judge paradigm. We use VLM judges in two complementary ways:

\begin{itemize}
    \item \textbf{Overall Rating}: We adapt the overall-rating judge from~\citet{yang2025ui2code}, in which a VLM evaluator assigns a single overall score on a 100-point scale in one pass. Their study validates this judge by demonstrating strong agreement with human judgments. The evaluation prompt is provided in Table~\ref{tab:appendix-evaluation-prompt-overall}.

    \item \textbf{Aspect-level Rating}: As shown in Table~\ref{tab:appendix-prompt-evaluation-aspect}, we define five aspects of visual fidelity. For each aspect, the VLM judge independently assigns a Likert-scale score from 1 to 7, and we report the mean score across aspects as the final aspect-level metric. This aspect-level evaluation provides a more fine-grained measure of localized visual changes during iterative refinement. As a sanity check, we manually inspected 60 rendered-image pairs from consecutive iterative rounds. The direction of model-based aspect-level score changes matched manual judgments in 47 cases, suggesting a useful approximate signal for tracking localized refinement trends.
\end{itemize}

Note that intermediate Likert-scale anchors are used only for the aspect-level ratings, not for the overall rating. The overall rating provides a holistic assessment across layout, spacing, typography, styling, and completeness. Because overall quality may reflect different combinations of errors, intermediate anchors would require prescribing potentially arbitrary trade-offs across these heterogeneous aspects. In contrast, each aspect-level rating isolates a single visual dimension, allowing intermediate anchor descriptions to be more clearly defined and helping to improve rating consistency. As a sanity check, the overall score correlates well with the anchored aspect mean (Pearson's $r=0.67$; Spearman's $\rho=0.74$).

\input{latex/sections/tables/evaluation_prompts}

\paragraph{Why not CLIP evaluation.}
We do not include the CLIP-based evaluation used in~\citet{si2025design2code}. CLIP similarities are useful for capturing coarse overall quality differences across generation models, but are often insensitive to the smaller, localized changes produced by iterative self-evolution. To verify this, we apply \verb|google/siglip2-large-patch16-512| to compute image similarity and manually inspect 40 pairs of consecutive iterative rounds with clear aspect-level performance changes. In 17 cases, the CLIP score either remained nearly unchanged or showed a trend opposite to the human-observed visual improvement or degradation, suggesting that CLIP-based metrics are not sufficiently informative for analyzing fine-grained refinement behavior.

%% file: latex/sections/tables/evaluation_prompts.tex
\begin{table}[!ht]
\begin{tcolorbox}[colback=gray!5!white,colframe=gray!75!black,title=Evaluation Prompt (Overall)]

\small

I will provide you with two UI images. The first is the reference UI screenshot; 
the second is a generated image rendered from VLM-produced code. 
Evaluate the visual fidelity of the generated image with respect to the reference, 
scoring from 0 to 100 (0 = not faithful at all, 100 = perfectly faithful).

\vspace{0.2em}

Consider the following aspects:

\vspace{0.15em}

\begin{itemize}[leftmargin=*,topsep=2pt,itemsep=-1pt]
    \item Layout \& geometry: structure, alignment, relative positions, and element sizes.
    \item Spacing: padding, margins, gaps, line-heights, and overall visual density.
    \item Typography: font sizes/weights, hierarchy, text placement, and line breaks.
    \item Styling: colors, backgrounds, borders, corner radius, shadows, and button styles.
    \item Completeness: all visible UI components are present with no omissions or hallucinations.
\end{itemize}

\vspace{0.2em}

First provide your reasoning, then output the final score inside LaTeX \verb|\boxed{}|.

\end{tcolorbox}
\vspace{-4mm}
\caption{Evaluation prompt for overall score.}
\vspace{-4mm}
\label{tab:appendix-evaluation-prompt-overall}
\end{table}

\begin{table}[!h]
\begin{tcolorbox}[colback=gray!5!white,colframe=gray!75!black,title=Evaluate Prompt (Aspect)]

\footnotesize

You will be shown two UI images. The first is the reference UI screenshot; 
the second is a generated image rendered from VLM-produced code.
Evaluate the visual fidelity of the generated image with respect to the reference, for the following aspect ONLY.

\vspace{0.2em}

\textbf{Aspect:} \texttt{\{aspect\_name\}}

\textbf{Definition:} \texttt{\{aspect\_description\}}

\vspace{0.2em}

Scoring: Use a 1--7 Likert scale 
(1 = not faithful at all, 7 = perfectly faithful).

\vspace{0.2em}

\textbf{Score standards (Likert 1--7):}
\begin{itemize}[leftmargin=*,topsep=2pt,itemsep=-3pt]
    \item 7: Perfect match for this aspect; no noticeable differences.
    \item 6: Very close match; only tiny, hard-to-notice differences.
    \item 5: Mostly matches; small differences are noticeable but do not change the overall impression.
    \item 4: Mixed quality; clear differences that moderately affect the aspect.
    \item 3: Weak match; major differences are obvious and significantly affect the aspect.
    \item 2: Very poor match; severe differences dominate this aspect.
    \item 1: Completely unfaithful for this aspect.
\end{itemize}

\vspace{0.2em}

\textbf{Important notes:}
\begin{itemize}[leftmargin=*,topsep=2pt,itemsep=-1.5pt]
    \item Evaluate only the specified aspect; do NOT consider other aspects.
    \item Ignore differences in external photographic content that cannot be faithfully reproduced using HTML/CSS.
\end{itemize}

\vspace{0.2em}

\textbf{Output format (strict):} Return a single JSON object ONLY, with keys:
\begin{itemize}[leftmargin=*,topsep=2pt,itemsep=-1.5pt]
    \item \texttt{aspect}: string, must equal the Aspect name above.
    \item \texttt{rationale}: justification focusing only on the specified aspect.
    \item \texttt{score}: integer in [1, 7].
\end{itemize}

\vspace{0em}
\noindent\rule{\linewidth}{0.4pt}
\vspace{0em}

\textbf{Aspect List:}
\begin{itemize}[leftmargin=*,topsep=2pt,itemsep=-1.5pt]
    \item \texttt{layout\_geometry}: Layout \& geometry: correctness of the overall page structure, including major regions (e.g., header, sidebar, main content), element alignment, relative positioning, and approximate element sizes.

    \item \texttt{spacing\_density}: Spacing \& density: consistency of padding, margins, gaps, and line-heights, as well as whether the overall visual density matches the reference UI.

    \item \texttt{typography\_text}: Typography \& text hierarchy: faithfulness of font sizes, font weights, hierarchical relationships between titles and body text, text placement, and line breaks.

    \item \texttt{styling\_visual}: Visual styling: accuracy of colors, backgrounds, borders, corner radius, shadows, dividers, and button or card styling.

    \item \texttt{completeness}: Content completeness: whether all visible UI elements in the reference are present in the generated UI, with no missing components and no hallucinated or extraneous elements.
\end{itemize}

\end{tcolorbox}
\vspace{-4mm}
\caption{Evaluation prompt for aspect-level scoring.}
\label{tab:appendix-prompt-evaluation-aspect}
\end{table}

%% file: latex/sections/x_additional_experiments.tex
\section{Additional Experiments and Analyses}
\label{appendix:additional_experiments}

\subsection{Human Evaluation}
\label{appendix:human_eval}

\begin{table}[t]
\centering
\small
\setlength{\tabcolsep}{2.3pt}
\begin{tabular}{l|ccc|ccc}
\toprule
Model &
\shortstack{\#\ours} &
\#Tie &
\shortstack{\#Naïve} &
\shortstack{Win} &
\shortstack{Win+Tie} &
$p$-value \\
\midrule
GPT-5.2 & 39 & 4 & 17 & 65.0\% & 71.7\% & 0.0046 \\
GPT-5.4 & 35 & 10 & 15 & 58.3\% & 75.0\% & 0.0066 \\
\bottomrule
\end{tabular}
\vspace{-2mm}
\caption{Human pairwise evaluation of the round-10 outputs produced by \ours and naïve self-evolution on UI2Code-Real. Win rates and win-plus-tie rates are computed over all 60 comparisons for each model. The reported $p$-values are obtained using two-sided exact sign tests that exclude ties.}
\label{tab:append-human-evaluation}
\vspace{-4mm}
\end{table}

We conduct an additional human evaluation on 60 samples from UI2Code-Real, comparing the round-10 outputs of \ours and naïve self-evolution generated by GPT-5.2 and GPT-5.4. Following the same overall-quality evaluation criteria as those specified in Table~\ref{tab:appendix-evaluation-prompt-overall}, two authors first evaluate each pair independently. They then resolve disagreements through discussion to assign each pair one of three consensus labels: a \ours win, a tie, or a naïve self-evolution win.

As shown in Table~\ref{tab:append-human-evaluation}, \ours is clearly preferred over naïve self-evolution for both GPT-5.2 and GPT-5.4. Two-sided exact sign tests on the non-tied comparisons confirm that this preference is statistically significant for both generation models. This provides complementary evidence that the improvements measured by the automated evaluators correspond to perceptible gains in output quality.

\subsection{Judge-Score Variability}
\label{appendix:judge_variability}

To assess evaluation variability, we re-evaluate the initial and round-10
outputs of all models using three additional independent judge runs,
separate from those used for the main results. Table~\ref{tab:judge_confidence} reports the mean~$\pm$~standard deviation of the round-10 improvement over the corresponding initial generation across these three runs. 

The relative ordering of \ours{} and naïve self-evolution in these additional runs matches the main results (Table~\ref{tab:main-results}) in every setting: \ours{} achieves a larger mean improvement in all nine frontier-model settings and in seven of the nine open-source-model settings, with both exceptions occurring on Qwen3-VL-32B-Instruct, whose behavior is analyzed in Appendix~\ref{appendix:qwen3vl-exception}.
Standard deviations range from $0.09$ to $0.82$ points and are generally smaller for frontier models. Overall, these additional runs reproduce the comparative pattern in the main results.

\begin{table*}[t]
\centering
\small
\scalebox{1.12}{
\begin{tabular}{llccc}
\toprule
\textbf{Model} & \textbf{Method} & \textbf{UI2Code-Real} & \textbf{Design2Code} & \textbf{Design2Code-HARD} \\
\midrule
\multirow{2}{*}{GPT-5.4}
 & Self-Evolve & $+$0.39{\scriptsize$\pm$0.13} & $-$1.28{\scriptsize$\pm$0.09} & $-$1.98{\scriptsize$\pm$0.59} \\
 & \ours{}     & $+$2.87{\scriptsize$\pm$0.29} & $+$0.98{\scriptsize$\pm$0.26} & $+$0.70{\scriptsize$\pm$0.35} \\
\midrule
\multirow{2}{*}{GPT-5.2}
 & Self-Evolve & $+$0.65{\scriptsize$\pm$0.22} & $-$2.54{\scriptsize$\pm$0.22} & $-$0.29{\scriptsize$\pm$0.21} \\
 & \ours{}     & $+$3.32{\scriptsize$\pm$0.28} & $+$0.72{\scriptsize$\pm$0.10} & $+$1.47{\scriptsize$\pm$0.51} \\
\midrule
\multirow{2}{*}{Claude-Sonnet-4.5}
 & Self-Evolve & $+$2.25{\scriptsize$\pm$0.18} & $+$0.44{\scriptsize$\pm$0.58} & $+$0.37{\scriptsize$\pm$0.34} \\
 & \ours{}     & $+$4.66{\scriptsize$\pm$0.31} & $+$0.57{\scriptsize$\pm$0.39} & $+$0.76{\scriptsize$\pm$0.30} \\
\midrule
\multirow{2}{*}{Qwen3-VL-32B}
 & Self-Evolve & $+$2.15{\scriptsize$\pm$0.60} & $+$1.59{\scriptsize$\pm$0.14} & $+$5.68{\scriptsize$\pm$0.63} \\
 & \ours{}     & $+$2.01{\scriptsize$\pm$0.61} & $+$1.76{\scriptsize$\pm$0.12} & $+$3.67{\scriptsize$\pm$0.74} \\
\midrule
\multirow{2}{*}{Qwen3.5-9B}
 & Self-Evolve & $+$1.74{\scriptsize$\pm$0.12} & $+$3.28{\scriptsize$\pm$0.23} & $+$4.47{\scriptsize$\pm$0.70} \\
 & \ours{}     & $+$2.53{\scriptsize$\pm$0.12} & $+$4.82{\scriptsize$\pm$0.09} & $+$5.29{\scriptsize$\pm$0.64} \\
\midrule
\multirow{2}{*}{Qwen3.6-35B-A3B}
 & Self-Evolve & $+$1.98{\scriptsize$\pm$0.23} & $+$1.08{\scriptsize$\pm$0.26} & $+$0.66{\scriptsize$\pm$0.67} \\
 & \ours{}     & $+$4.27{\scriptsize$\pm$0.82} & $+$1.67{\scriptsize$\pm$0.31} & $+$1.91{\scriptsize$\pm$0.59} \\
\bottomrule
\end{tabular}
}
\vspace{-1mm}
\caption{Round-10 improvement in overall score over the initial
generation, reported as mean~$\pm$~std over three additional
independent GPT-5.2 judge runs. Across all models and benchmarks, the relative ordering of \ours{} and naïve self-evolution in these additional runs is consistent with that reported in the main results (Table~\ref{tab:main-results}).}
\vspace{-1mm}
\label{tab:judge_confidence}
\end{table*}


\subsection{Computational Cost}
\label{appendix:computational-cost}

We analyze the computational cost of \ours for both API-based and open-source models using 50 randomly selected UI2Code-Real samples over 10 refinement rounds (500 rounds in total). All reported values are averaged per refinement round.

\paragraph{API-based cost.}
We report the average per-round input and output token counts and API cost of GPT-5.4 based on OpenAI's standard rates of \$2.50 per million input tokens and \$15 per million output tokens. As shown in Table~\ref{tab:appendix-api-cost}, although \ours makes three model calls per refinement round, EVOLVE and SELECT generate relatively short outputs. Compared with naïve self-evolution, \ours uses $3.16\times$ as many input tokens but only $1.21\times$ as many output tokens, resulting in $2.38\times$ as many total tokens and $1.60\times$ the API cost.

\begin{table}[t]
\centering
\small
\setlength{\tabcolsep}{2.5pt}
\begin{tabular}{lrrrr}
\toprule
Method
& \# Input
& \# Output
& \# Total
& Cost (\textcent) \\
\midrule
Naïve Self-Evolve
& 5,912.0
& 3,971.0
& 9,883.0
& 7.4 \\
\midrule
\ours{} EVOLVE
& 6,103.8
& 605.3
& 6,709.1
& 2.4 \\
\ours{} SELECT
& 6,517.1
& 85.4
& 6,602.5
& 1.8 \\
\ours{} IMPROVE
& 6,059.2
& 4,131.5
& 10,190.7
& 7.7 \\
\ours{} total
& 18,680.1
& 4,822.2
& 23,502.3
& 11.9 \\
\bottomrule
\end{tabular}
\vspace{-1mm}
\caption{Average per-round token usage and API cost on 50 UI2Code-Real samples over 10 refinement rounds using GPT-5.4. ``\#'' denotes the number of tokens, and cost is reported in cents (\textcent).}
\label{tab:appendix-api-cost}
\end{table}

\paragraph{Open-source inference latency.}
For the open-source setting, we measure average per-round inference time using Qwen3.5-9B served with \texttt{vLLM} on a single H200 GPU with a batch size of 1. The first two samples are excluded to allow the inference engine to warm up. As shown in Table~\ref{tab:appendix-open-source-cost}, \ours takes 105.3 seconds per refinement round, compared with 102.7 seconds for naïve self-evolution, corresponding to only 2.5\% additional latency. In this setting, inference time is dominated by autoregressive decoding of the long HTML output rather than prompt prefilling; consequently, the short-output EVOLVE and SELECT calls introduce relatively little additional latency.

\begin{table}[t]
\centering
\small
\setlength{\tabcolsep}{2.5pt}
\begin{tabular}{lrrrr}
\toprule
Method
& \# Input
& \# Output
& \# Total
& Time (s) \\
\midrule
Naïve Self-Evolve
& 9,025.4
& 3,827.7
& 12,853.1
& 102.7 \\
\midrule
\ours{} EVOLVE
& 9,215.4
& 487.5
& 9,702.9
& 13.2 \\
\ours{} SELECT
& 9,510.5
& 81.3
& 9,591.8
& 2.4 \\
\ours{} IMPROVE
& 9,113.8
& 3,423.9
& 12,537.7
& 89.7 \\
\ours{} total
& 27,839.7
& 3,992.7
& 31,832.4
& 105.3 \\
\bottomrule
\end{tabular}
\caption{Average per-round token usage and inference time for Qwen3.5-9B. Results exclude the first two samples, which are used to warm up the vLLM inference engine. ``\#'' denotes the number of tokens.}
\vspace{-4mm}
\label{tab:appendix-open-source-cost}
\end{table}

\vspace{3mm}
Considering both effectiveness and efficiency, \ours increases monetary cost to $1.60\times$ with GPT-5.4 while mitigating the late-round degradation associated with visual repair coupling. With Qwen3.5-9B, \ours incurs only 2.5\% additional latency while achieving $1.4\times$ the average round-10 performance gain on the selected 50-sample subset. This moderate additional cost is spent on structuring the repair context rather than on generating more code, and is accompanied by more stable and effective self-evolution.

\paragraph{Separation of \textsc{Evolve} and \textsc{Select}.}
As discussed in Section~\ref{sec:method-rubric}, issuing \textsc{Evolve}
and \textsc{Select} as separate calls keeps candidate generation and
prioritization as distinct steps. We additionally evaluate a variant that merges EVOLVE and SELECT into a single request on the same 50-sample subset. The merged call uses an average of 6,168.8 input tokens and 641.7 output tokens per round, costing 2.5 cents. By comparison, the separate EVOLVE and SELECT calls cost 2.4 and 1.8 cents, respectively, as reported in Table~\ref{tab:appendix-api-cost}. Merging the two modules reduces the total per-round cost by 14.2\%, but decreases the average round-10 improvement in overall score by 0.5 points.


\subsection{Understanding the Exception of Qwen3-VL-32B-Instruct }
\label{appendix:qwen3vl-exception}

\begin{table}[t]
\centering
\small
\setlength{\tabcolsep}{3pt}
\begin{tabular}{@{}llcc@{}}
\toprule
Model & Rubric-gen.
& \shortstack{Spacing density}
& \shortstack{Completeness} \\
\midrule
Qwen3-VL
    & Self
    & 27.1\% $\rightarrow$ 13.6\%
    & 2.3\% $\rightarrow$ 4.5\% \\

Qwen3.5
    & Self
    & 12.8\% $\rightarrow$ 4.8\%
    & 8.1\% $\rightarrow$ 10.0\% \\

Qwen3.6
    & Self
    & 8.8\% $\rightarrow$ 3.6\%
    & 9.2\% $\rightarrow$ 8.9\% \\
\midrule
Qwen3-VL
    & GPT-5.4
    & 16.6\% $\rightarrow$ 8.7\%
    & 7.7\% $\rightarrow$ 8.0\% \\
\bottomrule
\end{tabular}
\vspace{-0.5mm}
\caption{Rubric-type distributions produced by EVOLVE and retained by SELECT. ``Model'' denotes the model used for visual-code refinement, while ``Rubric-gen.'' denotes the model used to generate and select rubrics. Each cell reports EVOLVE $\rightarrow$ SELECT.}
\vspace{-3mm}
\label{tab:appendix-qwen3vl-rubric-distribution}
\end{table}

To better understand why Qwen3-VL-32B-Instruct does not consistently benefit from \ours compared with naïve self-evolution, we analyze the distribution of rubric types across Qwen models during the EVOLVE $\rightarrow$ SELECT procedure. As reported in Table~\ref{tab:appendix-qwen3vl-rubric-distribution}, Qwen3-VL-32B-Instruct generates substantially more spacing-density rubrics and fewer completeness rubrics than Qwen3.5-9B and Qwen3.6-35B-A3B, and SELECT only partially corrects this imbalance. Consequently, its selected targets remain overly focused on local adjustments, such as margins and line height, while underemphasizing missing or extraneous components, which can trap refinement in repeated local edits. Using GPT-5.4-generated rubrics substantially reduces this imbalance and clearly outperforms naïve self-evolution (Figure~\ref{fig:gpt-rubrics-ui2code}). These results suggest that the performance gap primarily originates from biased rubric generation that SELECT does not fully correct, rather than from the model's ability to perform rubric-guided improvement.

%% file: latex/sections/x_additional_case.tex
\section{Case Study}
\label{appendix:case_study}
\begin{itemize}[leftmargin=*]
    \item In Table~\ref{tab:appendix-case-qualitative-rounds}, we provide a trajectory-level comparison between \ours and naïve self-evolution on the same target screenshot. \ours yields more controlled and consistent improvements, whereas naïve self-evolution oscillates across rounds, repeatedly altering already-correct regions and exhibiting stronger visual repair coupling.

    \item In Table~\ref{tab:appendix-case-qwen-gpt-rubrics-examples}, we present three selected examples contrasting GPT-generated rubrics with self-generated rubrics for the same previous-round Qwen outputs. These examples illustrate the granularity difference behind the quantitative trends in Table~\ref{tab:rubric-content-gpt-vs-self}: stronger external rubrics provide broader and more structural guidance.
\end{itemize}

\input{latex/sections/tables/sample_trajectory}

\input{latex/sections/tables/gpt_rubrics_qwen_example}

%% file: latex/sections/tables/sample_trajectory.tex
\begin{table*}[t]
\centering
\small
\begingroup
\setlength{\tabcolsep}{3pt}
\renewcommand{\arraystretch}{0.9}
\setlength{\fboxsep}{0pt}
\setlength{\fboxrule}{0.5pt}

\begin{tabular}{@{}R{0.04\linewidth} Z{0.24\linewidth} | Y{0.42\linewidth} Z{0.24\linewidth}@{}}
\toprule
Round & Self-Evolve & Selected Rubric & \ours \\
\midrule

GT &
\imgbox{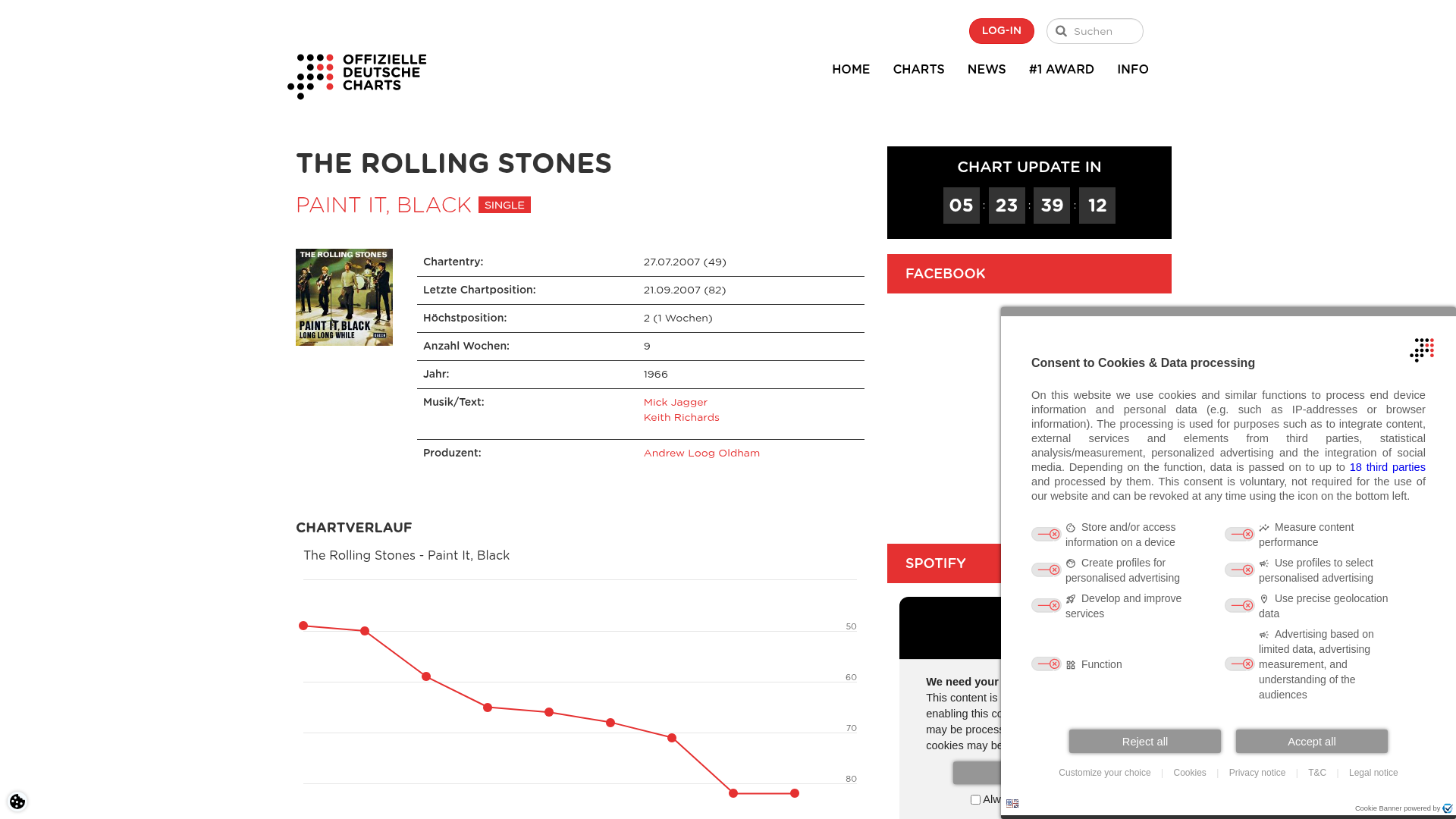} &
-- &
\imgbox{latex/figures/ours_trajectory/ui2code_gpt52_sample057/gt.png} \\

0 &
\imgbox{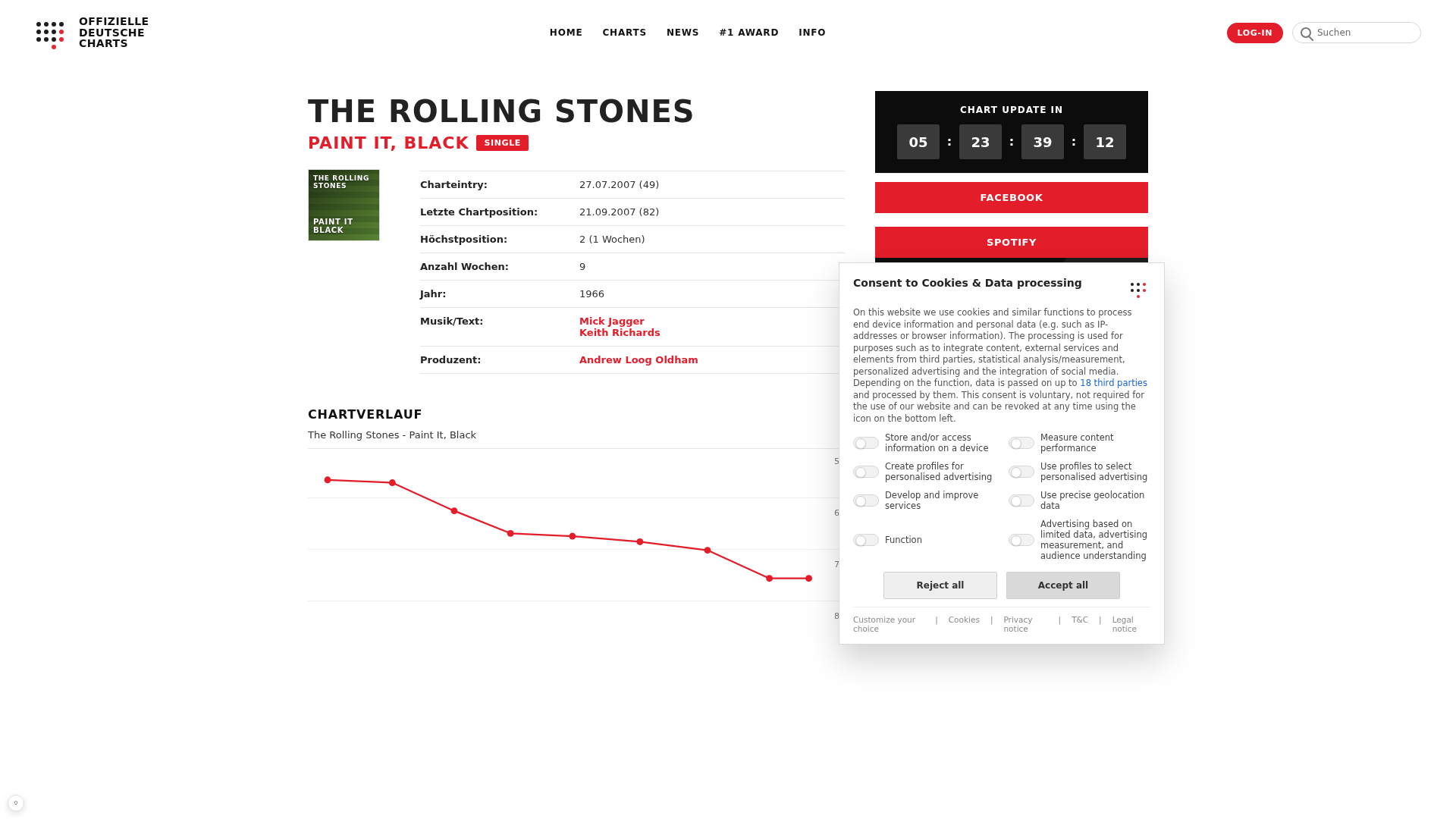} &
-- &
\imgbox{latex/figures/ours_trajectory/ui2code_gpt52_sample057/direct.png} \\

1 &
\imgbox{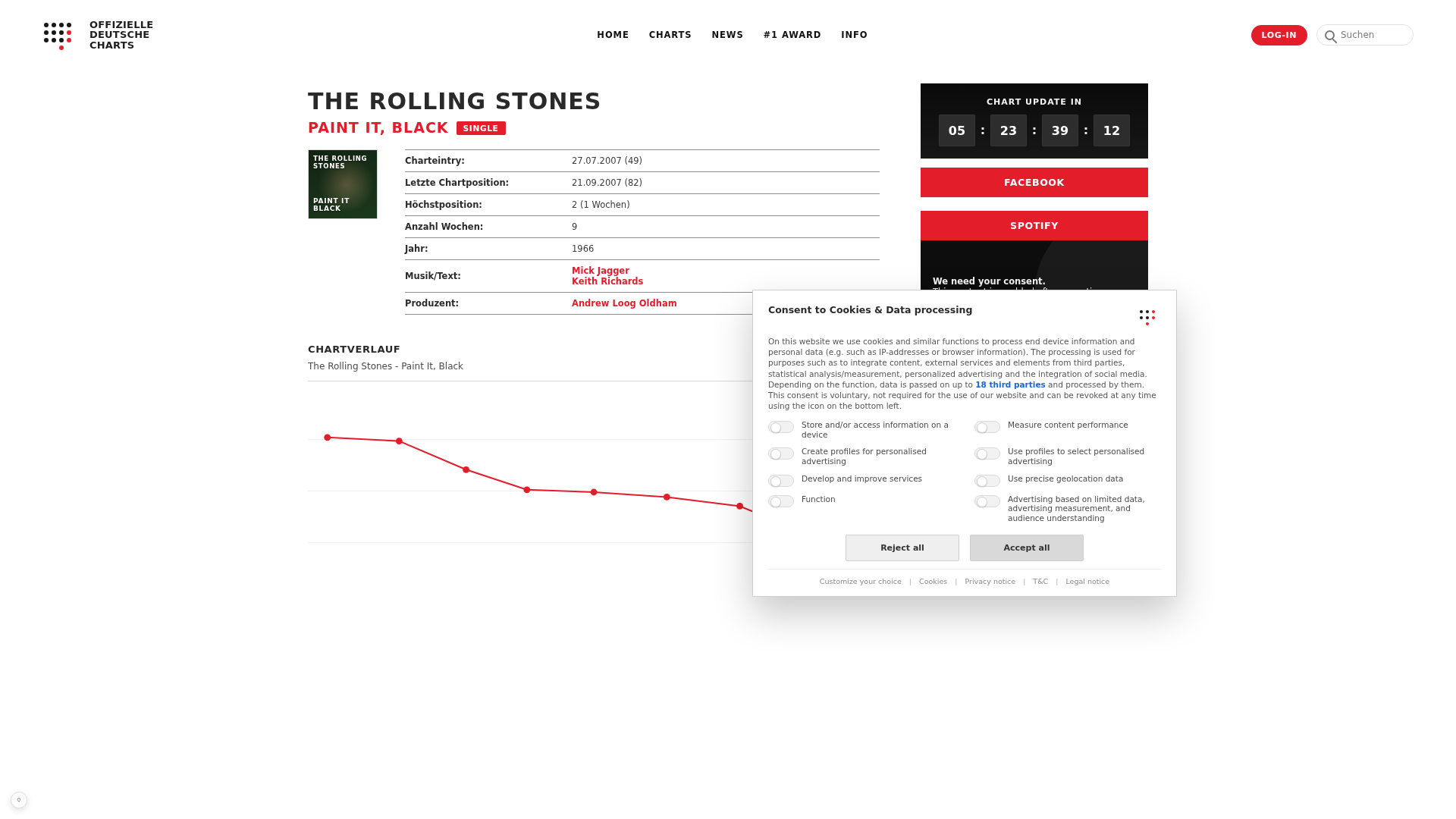} &
\rubriccell{Cookie modal size and placement}{layout\_geometry}{%
The reference cookie dialog is larger and anchored lower/right, covering a big portion of the sidebar and chart area; the current modal is smaller and placed too high (top:260px) and not as far right. Increase modal width/height and reposition it to bottom-right (e.g., position:fixed; right:70px; bottom:40px) to match the overlay behavior and coverage seen in the target.%
} &
\imgbox{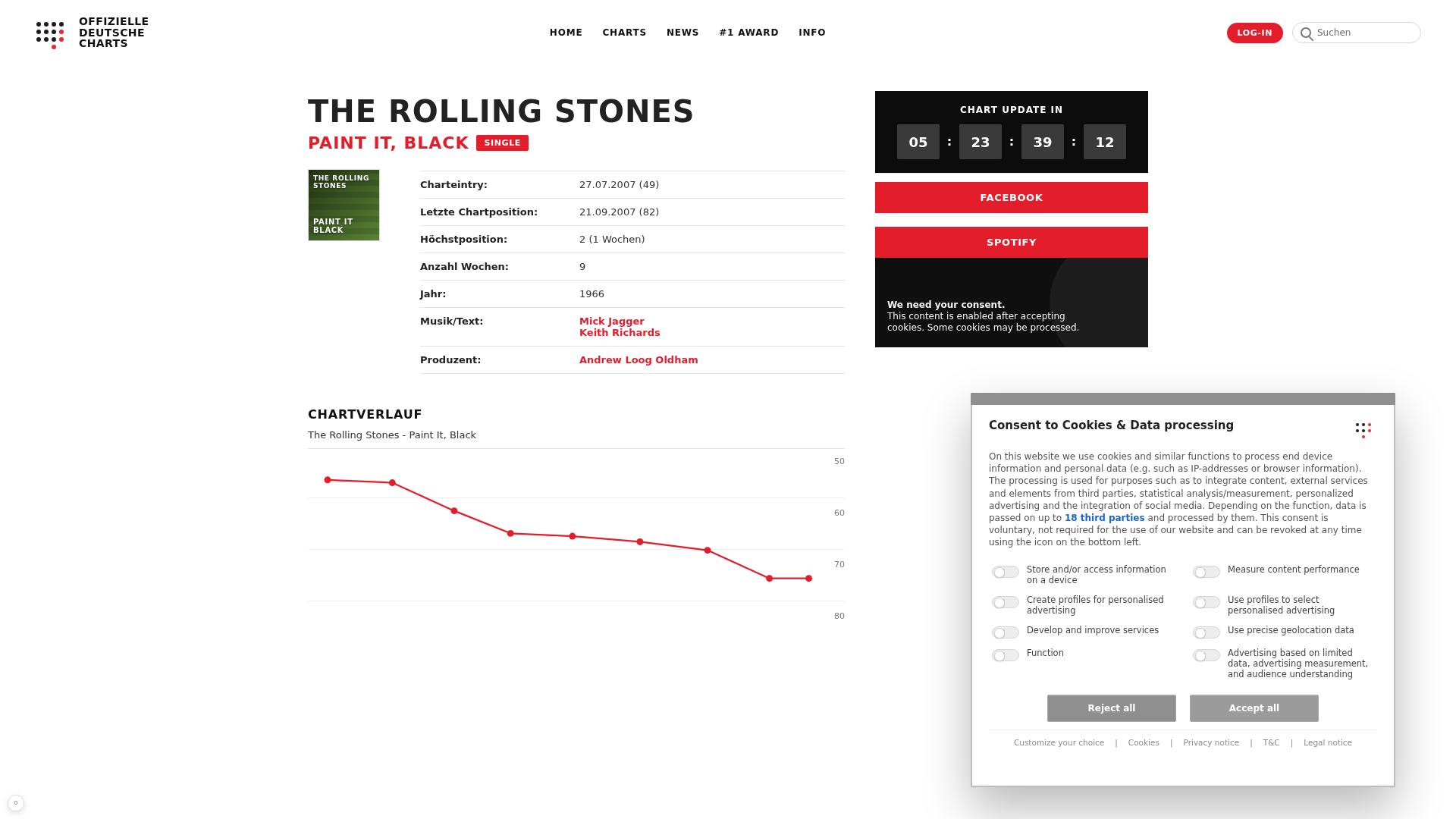} \\

2 &
\imgbox{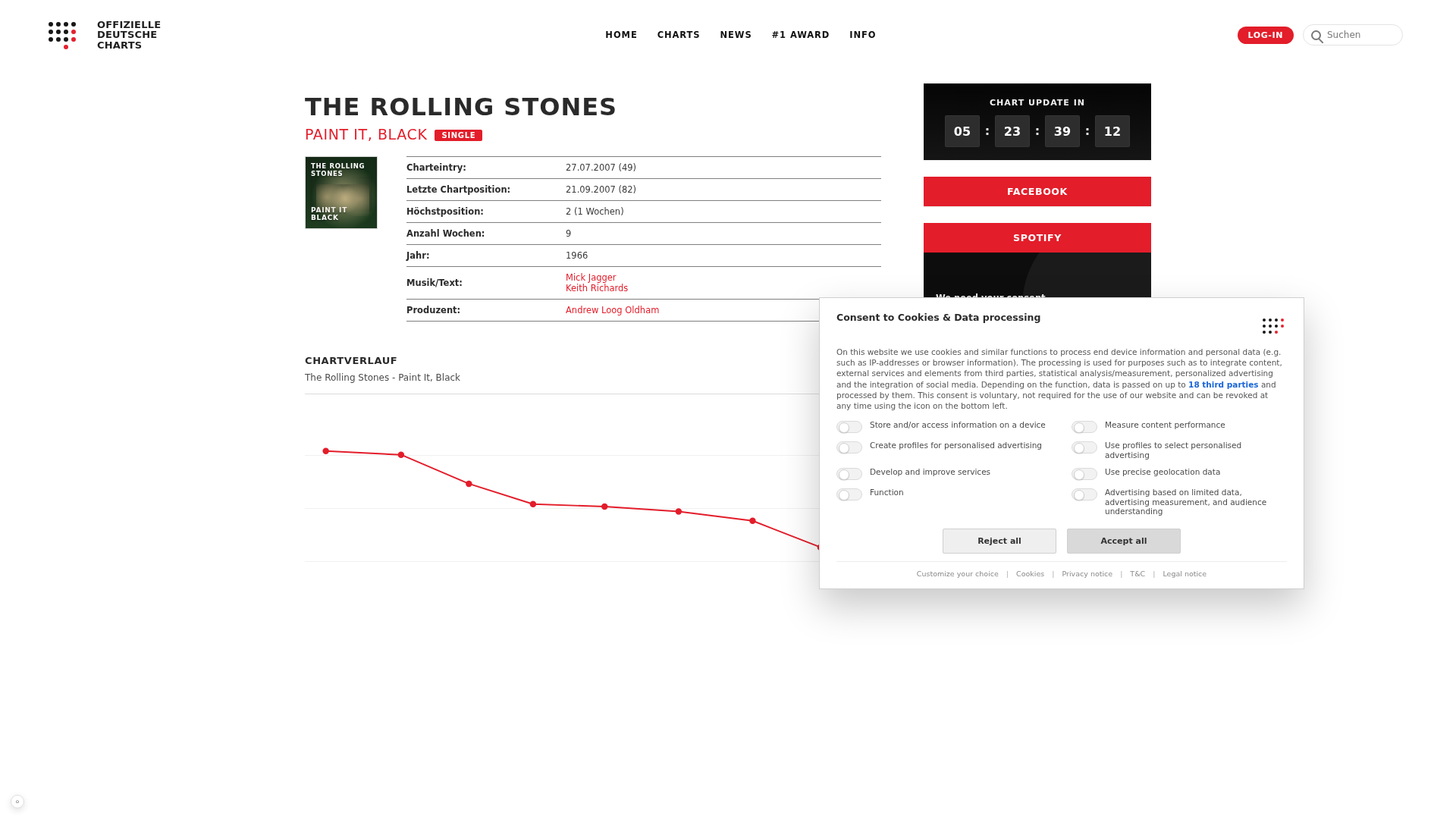} &
\rubriccell{Top navigation horizontal alignment}{layout\_geometry}{%
In the reference, the main nav (HOME, CHARTS, NEWS, \#1 AWARD, INFO) sits right-of-center near the top, while the LOG-IN + search are at the far top-right; in the current render, the nav is centered and the right controls are pushed down the line. Adjust the topbar layout, e.g., three-column grid with fixed left brand, flexible middle nav aligned right, and right controls pinned to the edge, so the nav is not centered and matches the reference positioning.%
} &
\imgbox{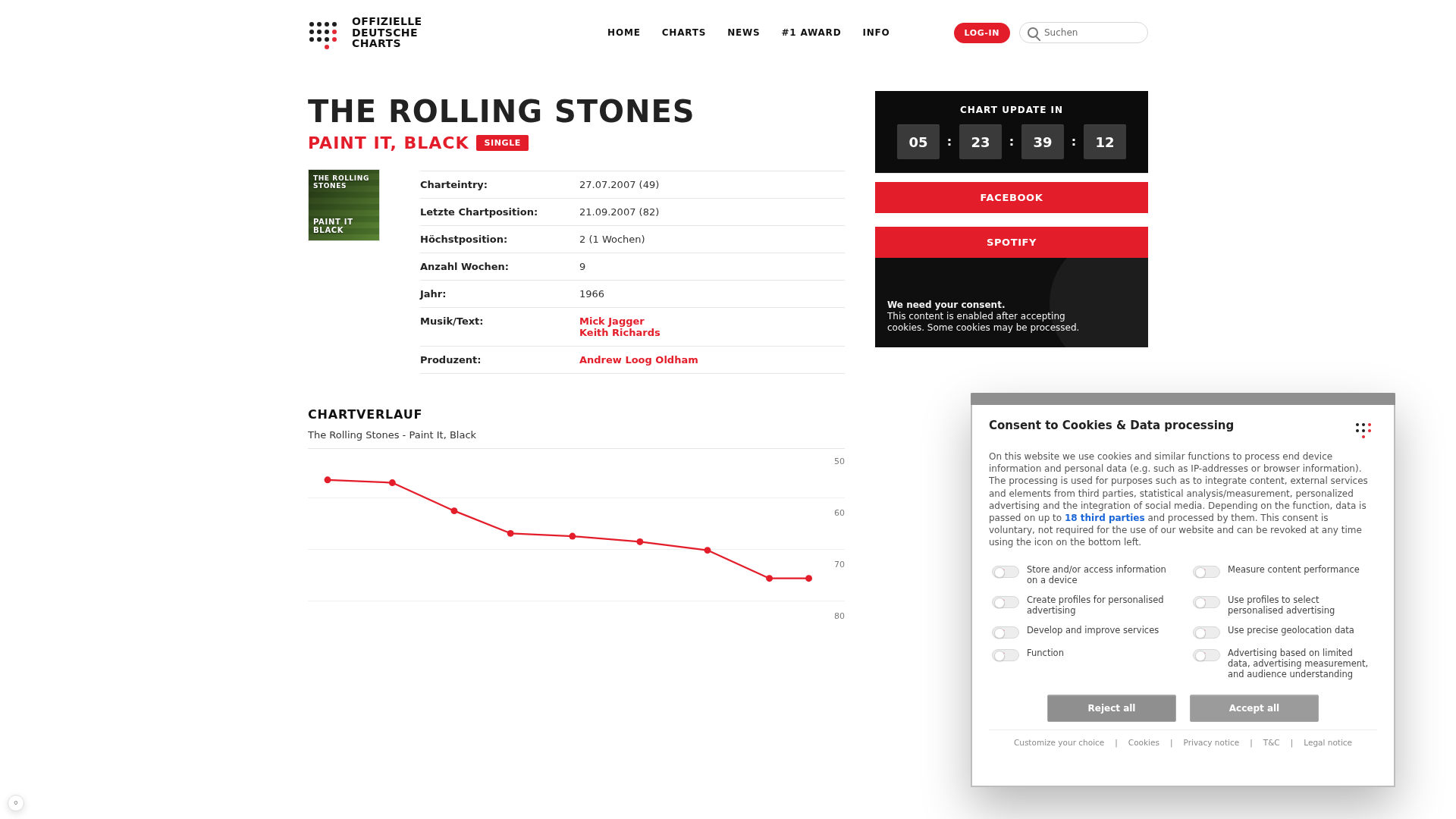} \\

3 &
\imgbox{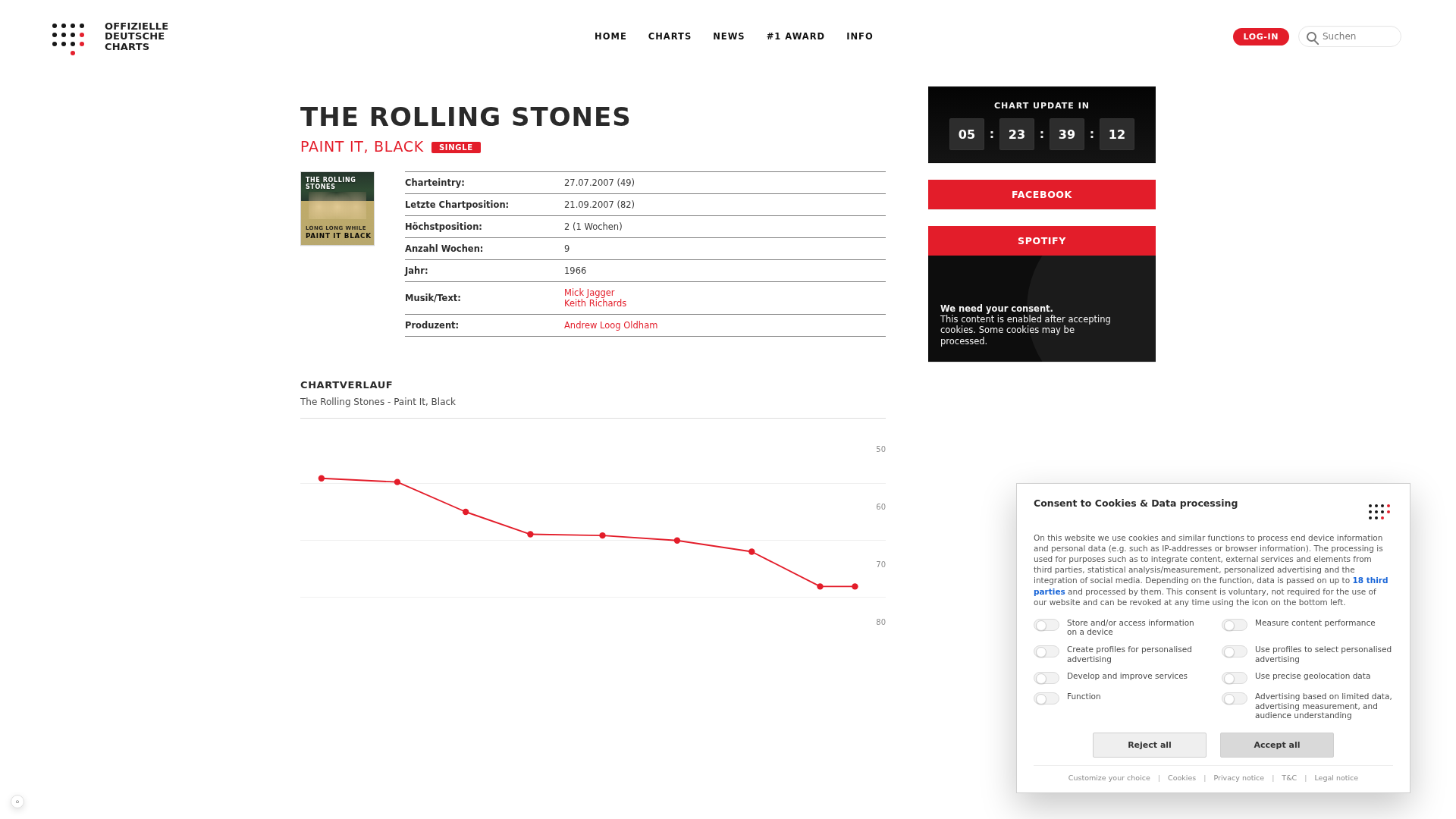} &
\rubriccell{Album cover artwork: use an image instead of gradient placeholder}{completeness}{%
The reference shows the actual ``Paint It, Black'' cover image with detailed artwork. The current render uses a green striped placeholder; replace the cover background with an image or a background-image and size it to match the square cover presentation.%
} &
\imgbox{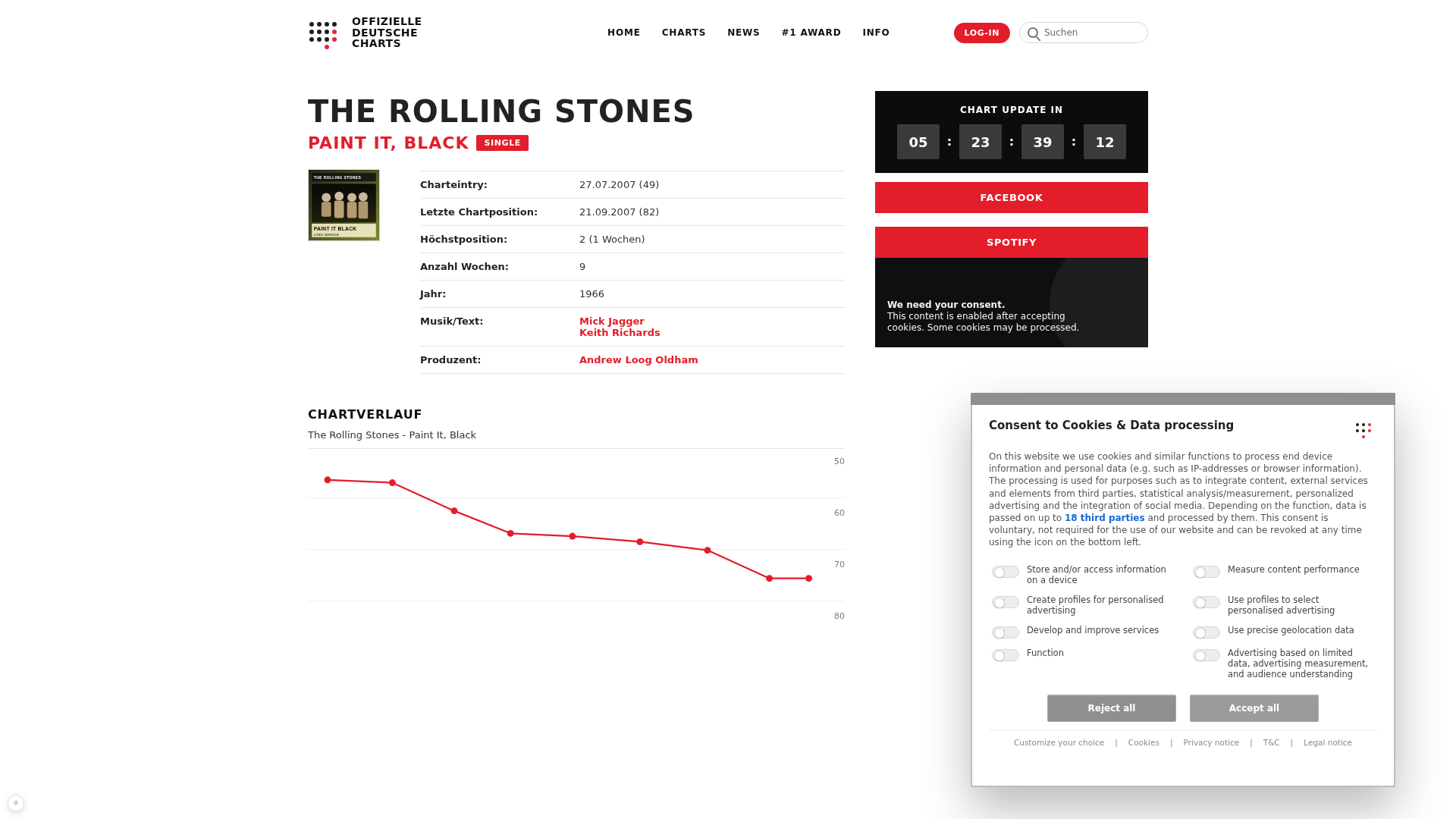} \\

4 &
\imgbox{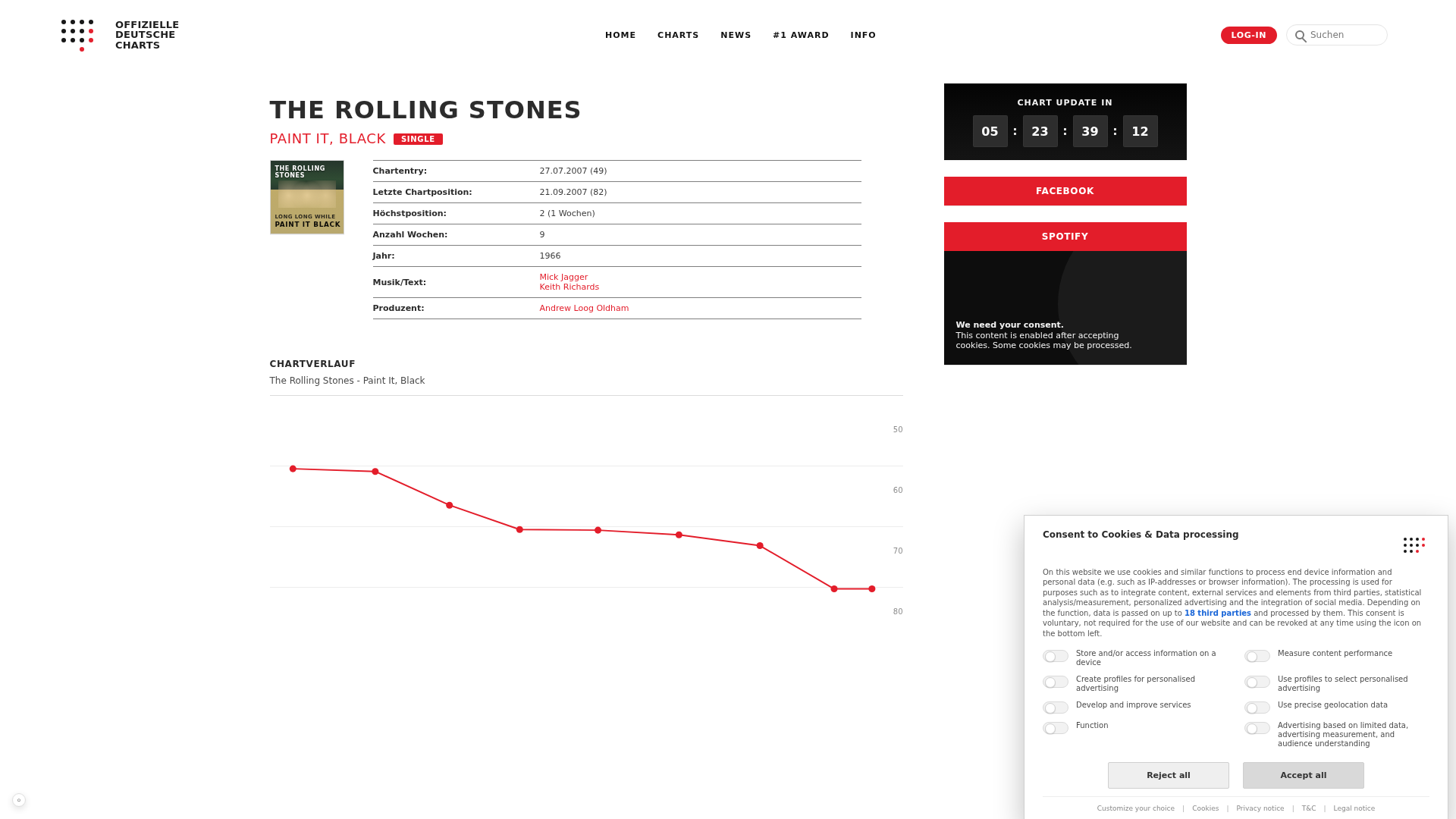} &
\rubriccell{Page width and left content scaling}{layout\_geometry}{%
The reference has a wider overall canvas with more whitespace and a slightly more left-shifted main content block; the current layout feels more compressed due to a fixed 1200px wrap and heavy inner padding. Increase the max width or switch to a larger responsive max-width and adjust left/right padding so the main column proportions match the reference.%
} &
\imgbox{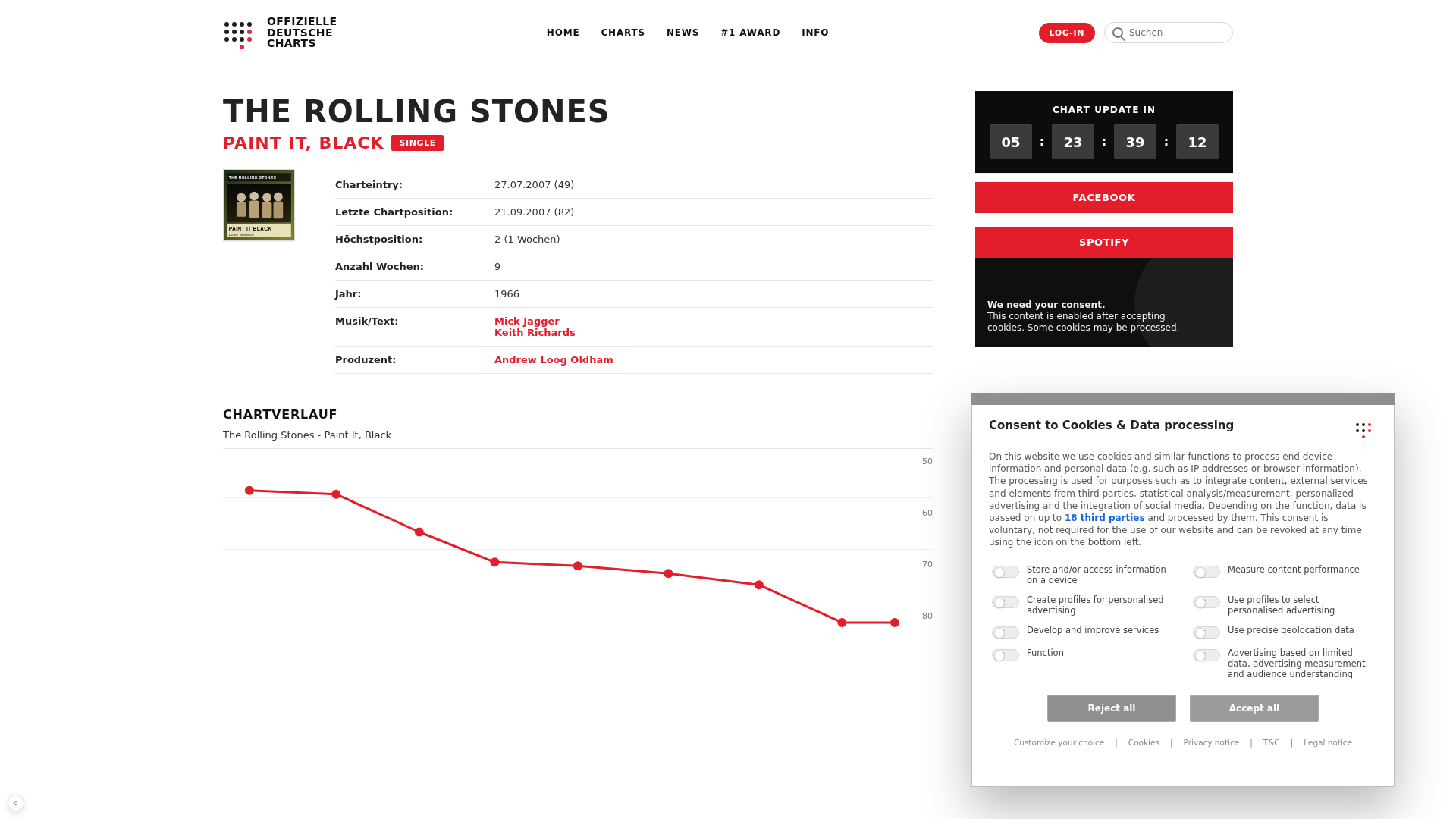} \\

5 &
\imgbox{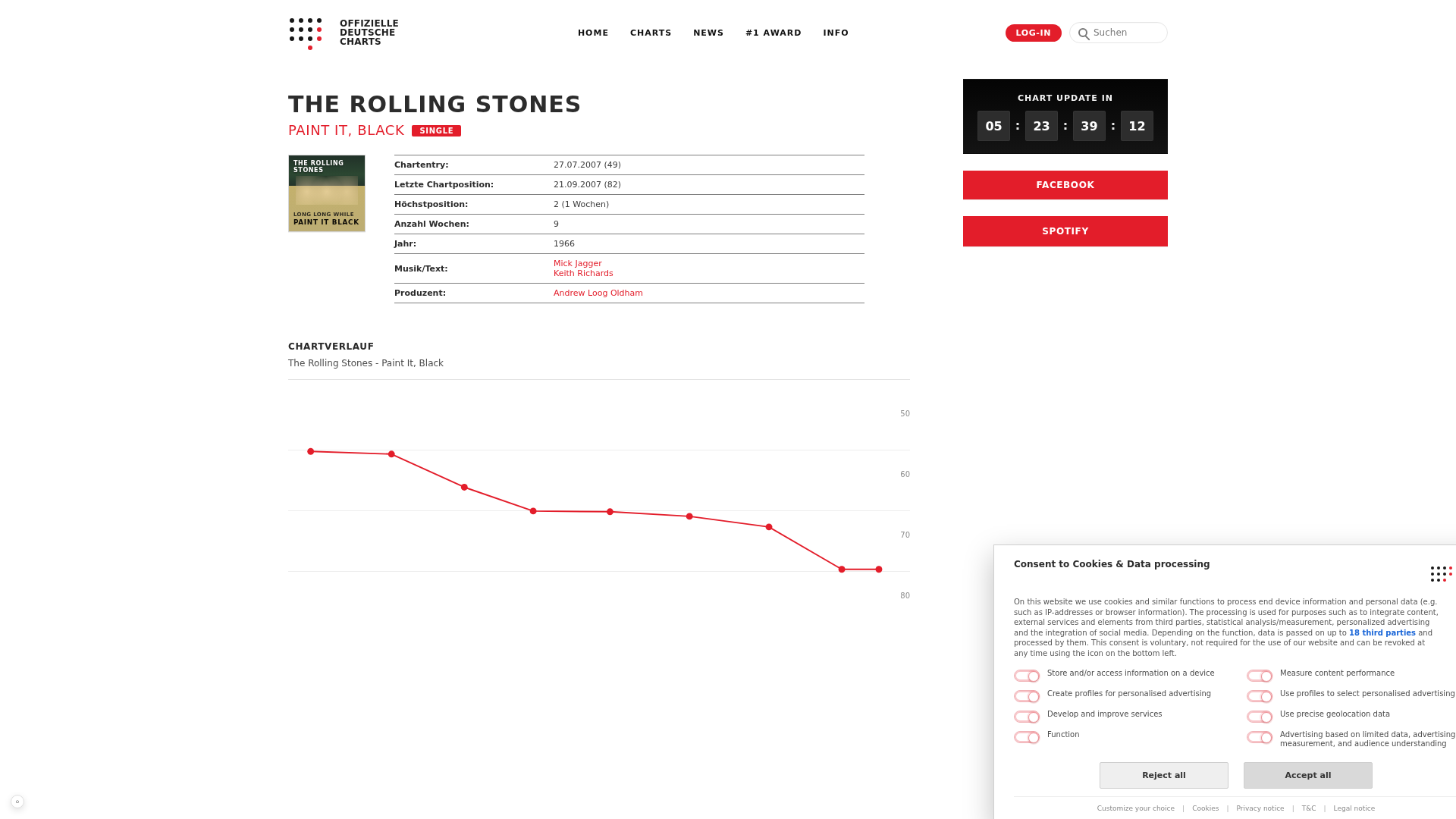} &
\rubriccell{Right sidebar card widths and stacking}{layout\_geometry}{%
In the reference, the black countdown card, red FACEBOOK bar, and red SPOTIFY bar are narrower and aligned as a single column with more whitespace to the right; in the current render they span the full 340px sidebar and look too wide. Reduce the sidebar column width and/or set a fixed card width with left alignment so the cards match the narrower block seen in the target.%
} &
\imgbox{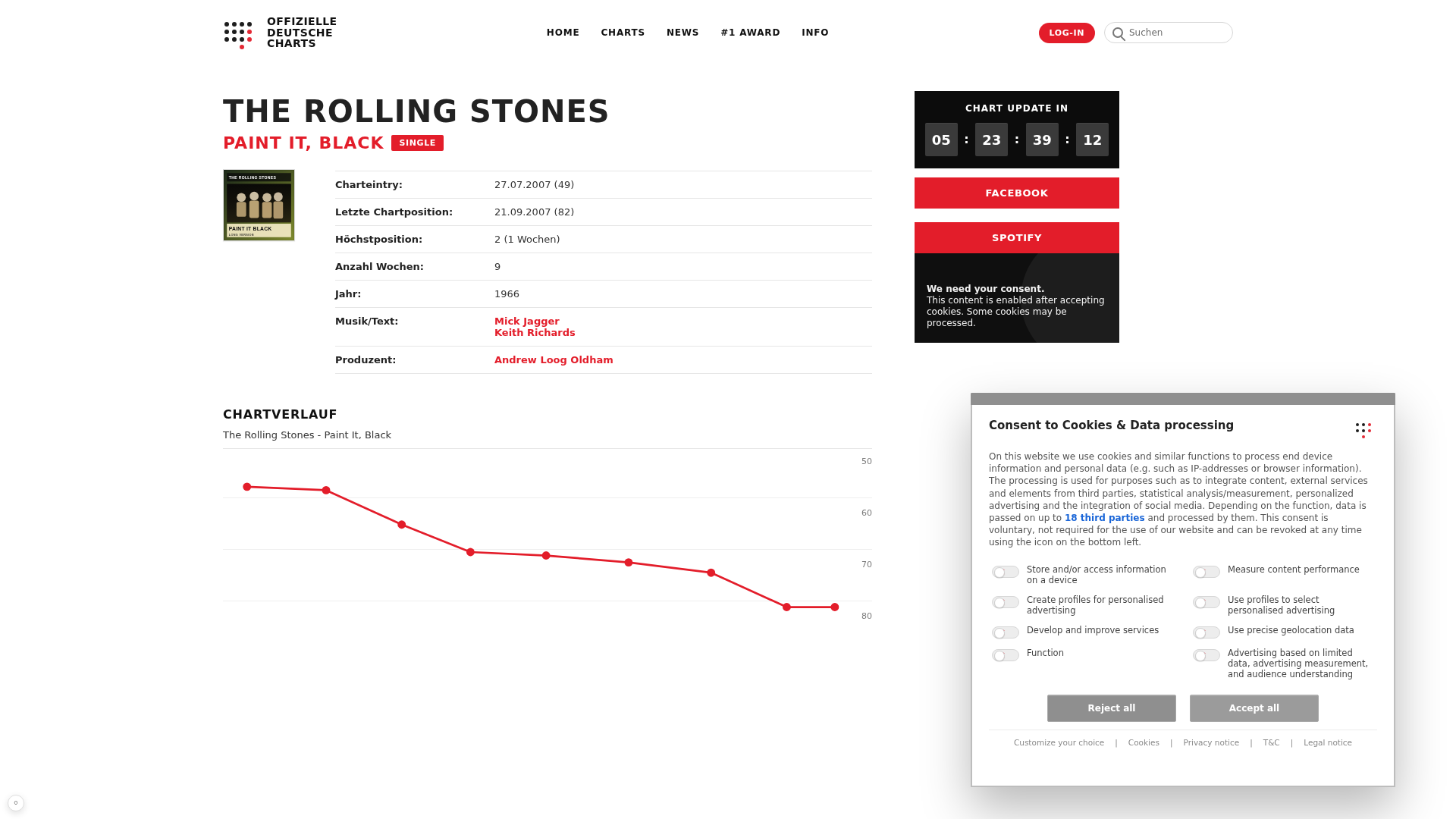} \\

6 &
\imgbox{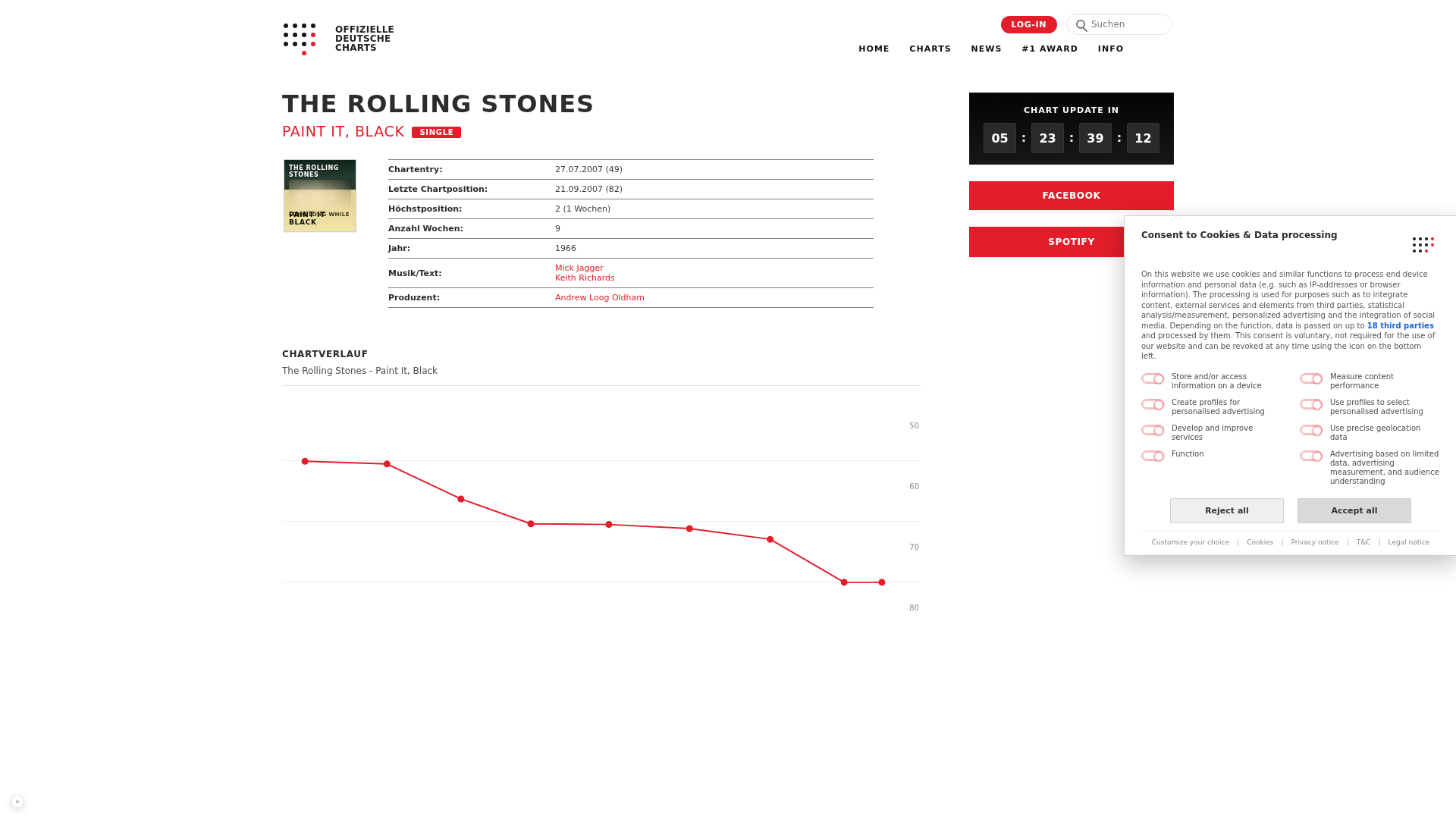} &
\rubriccell{Cookie toggle controls should be radio/slider style with red accent on right}{styling\_visual}{%
The reference shows compact toggle/radio-like controls with a red accent/knob on the right side of each control, while the current switches are gray with a white knob on the left and an `$\times$' overlay. Redesign the toggle to place the knob/indicator on the right and use the red accent to match the reference control styling.%
} &
\imgbox{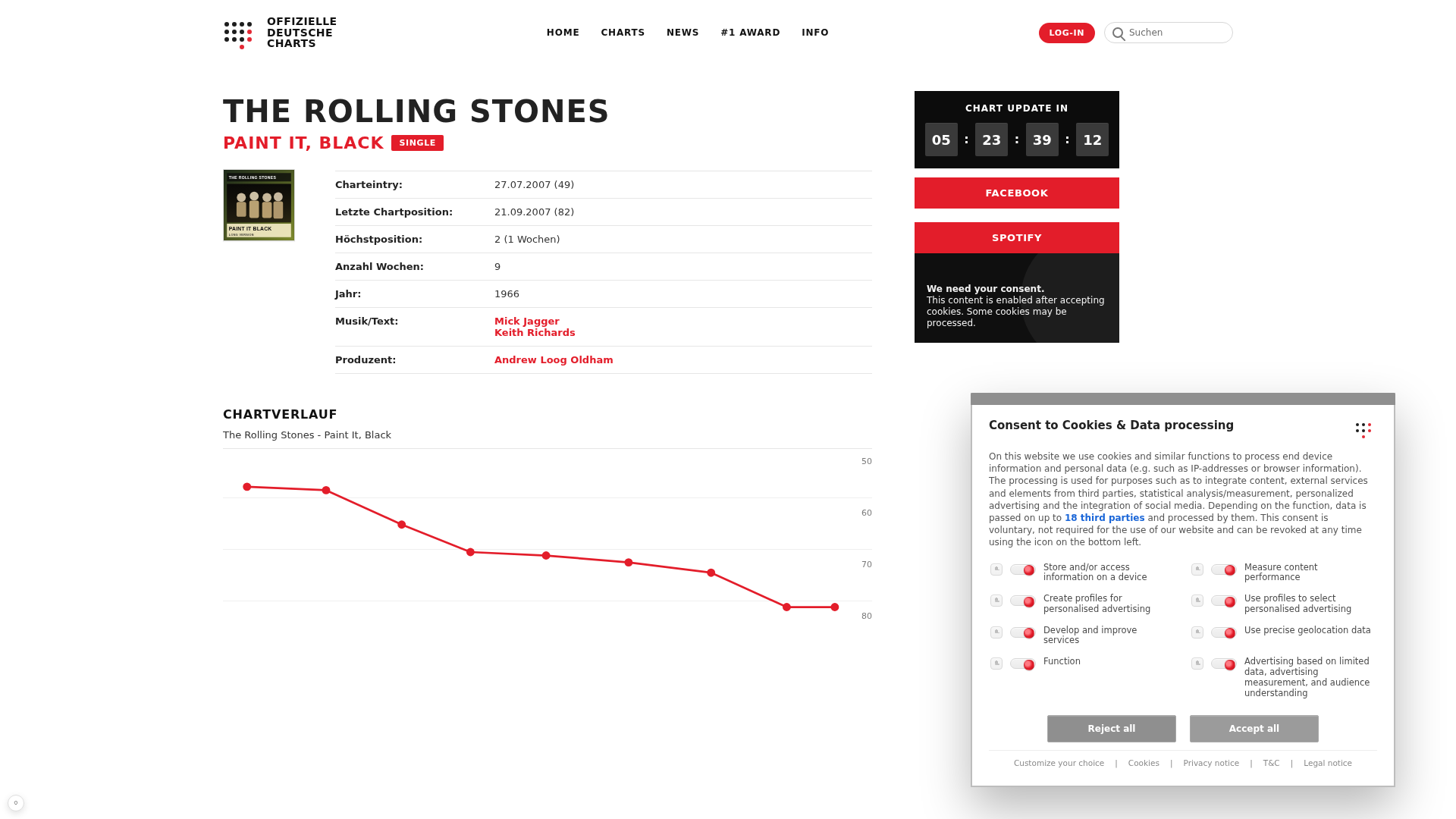} \\

7 &
\imgbox{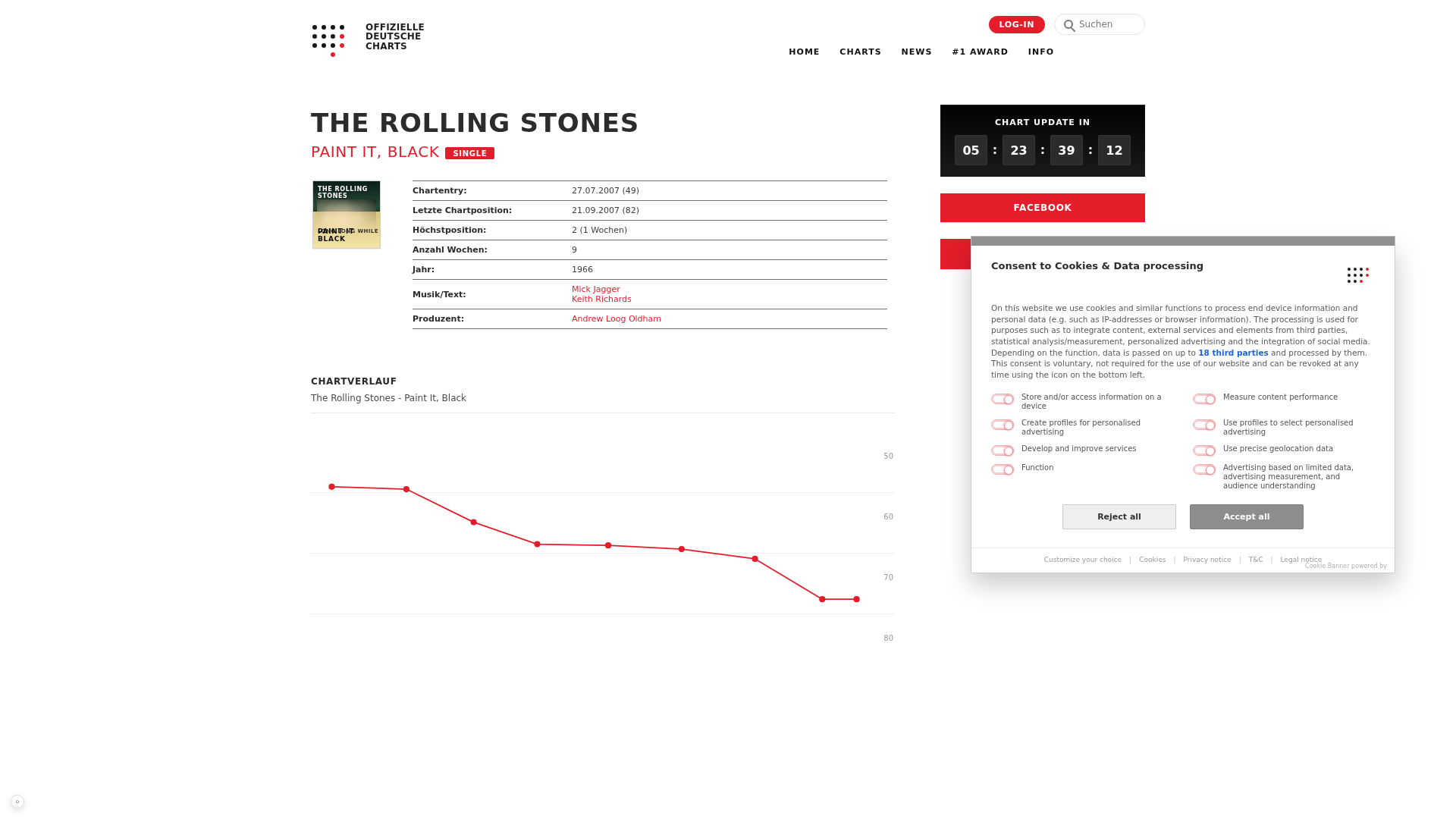} &
\rubriccell{FACEBOOK bar text alignment}{typography\_text}{%
In the reference, the FACEBOOK label is left-aligned within the red bar with noticeable left padding; the current implementation centers the label. Change the FACEBOOK bar to left-align text and add left padding similar to the reference, roughly 18--24px.%
} &
\imgbox{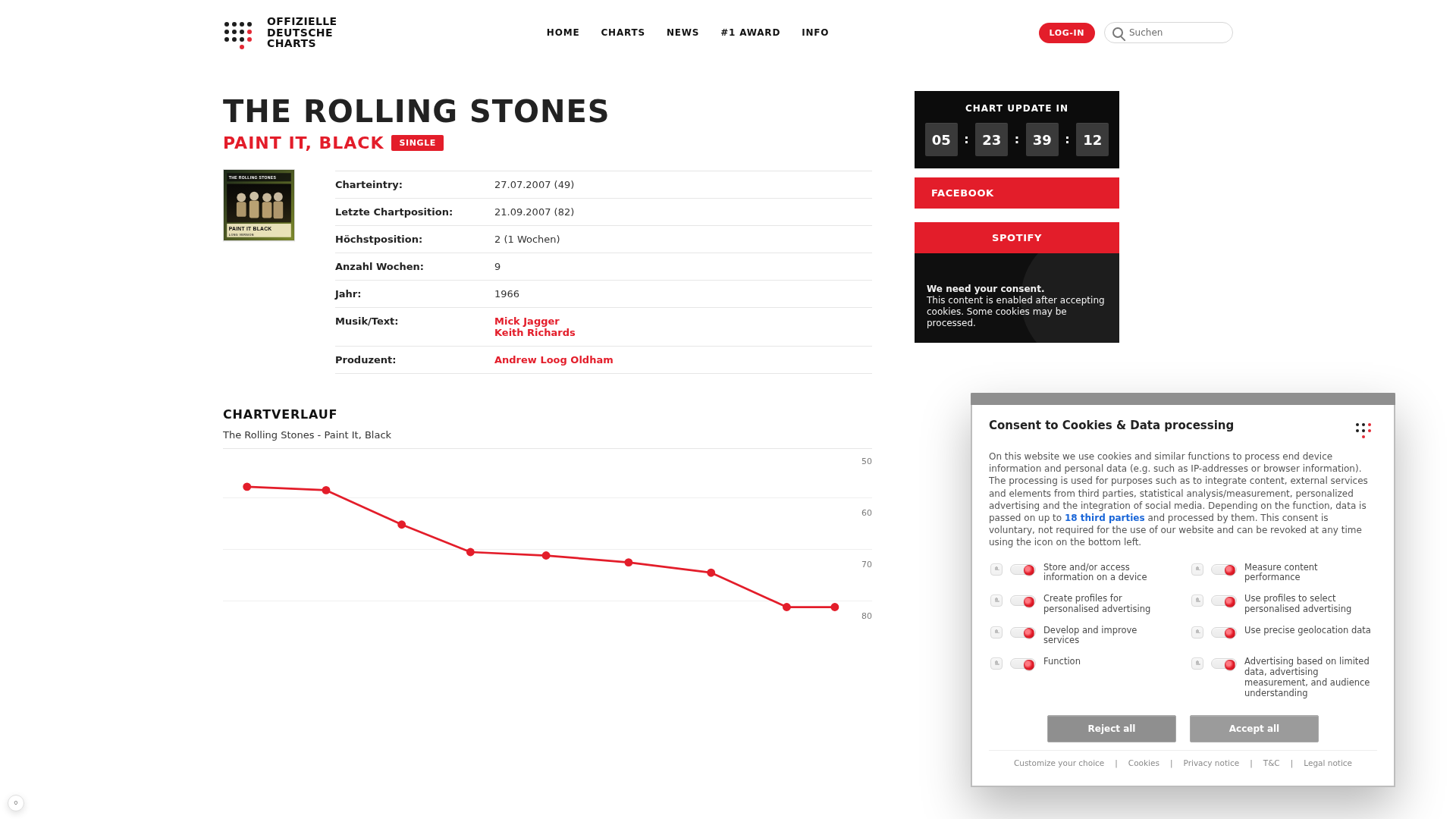} \\

\bottomrule
\end{tabular}
\renewcommand{\arraystretch}{1.0}
\endgroup
\vspace{-2mm}
\caption{Qualitative comparison on iterative rounds generated by GPT-5.2. Naïve self-evolution exhibits unstable refinement with oscillatory edits across rounds. In round 5, although it fixes the toggle color and top-bar length, it distorts the cookie modal's scale. This error is repeatedly modified in later rounds but remains unresolved. In contrast, \ours follows a more structured repair trajectory guided by selected rubrics, progressively addressing the cookie modal, top bar, line-chart scale, and toggle color. Although \ours still exhibits visual repair coupling when the chart-scale adjustment temporarily misplaces the cookie modal in round 4, it corrects the issue in the next round, leading to more controlled and recoverable gains.}
\label{tab:appendix-case-qualitative-rounds}
\end{table*}

%

%% file: latex/sections/tables/gpt_rubrics_qwen_example.tex
\begin{table*}[t]
\centering
\small
\begingroup
\setlength{\tabcolsep}{3pt}
\renewcommand{\arraystretch}{0.9}
\setlength{\fboxsep}{0pt}
\setlength{\fboxrule}{0.5pt}

\begin{tabular}{@{}R{0.09\linewidth} | Z{0.2\linewidth} Z{0.2\linewidth} | Y{0.25\linewidth} Y{0.25\linewidth} @{}}
\toprule
Model & GT & Previous & GPT-5.4 Rubric & Self Rubric \\
\midrule
Qwen3-VL-32B-Instruct &
\imgbox{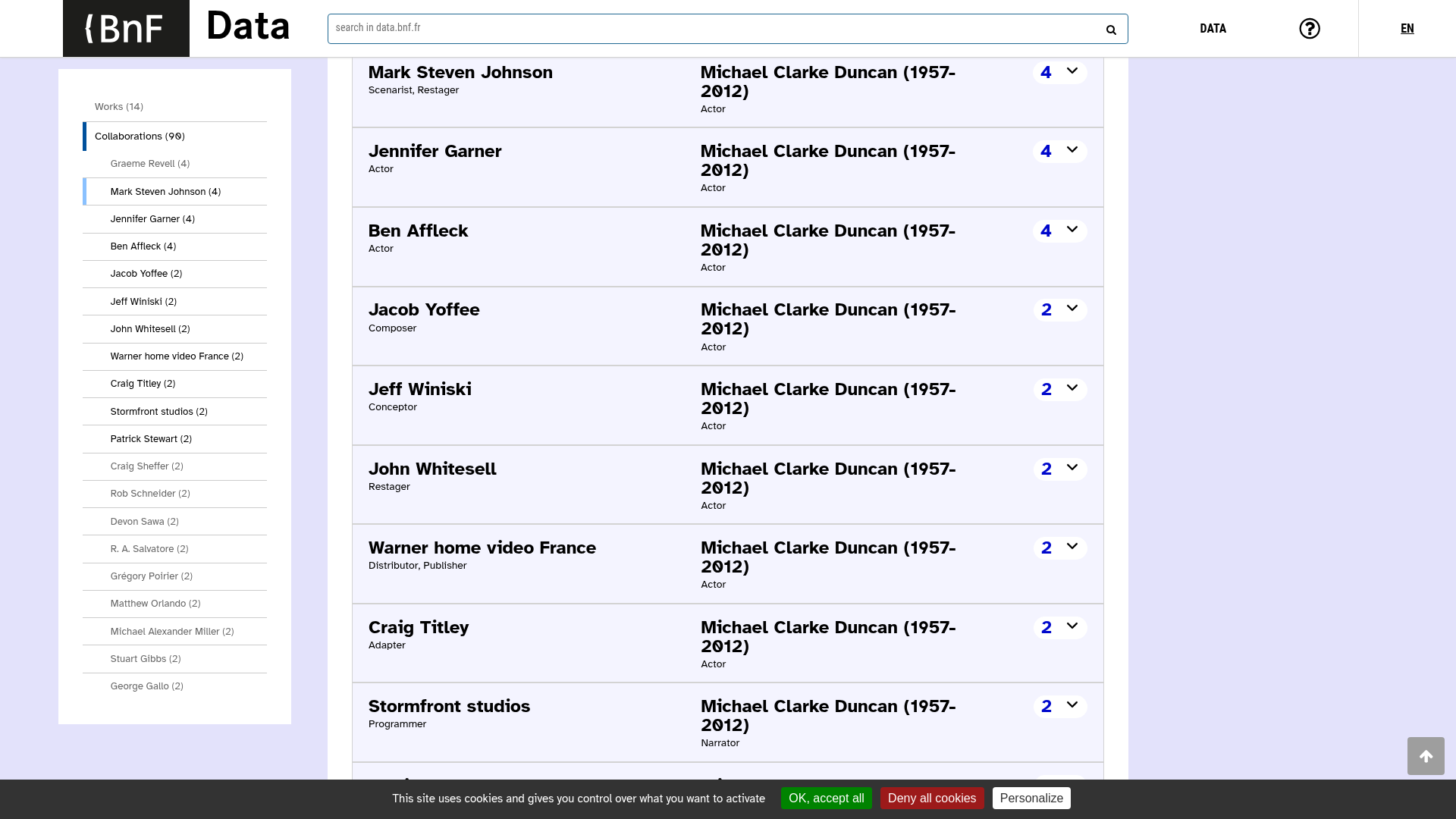} &
\imgbox{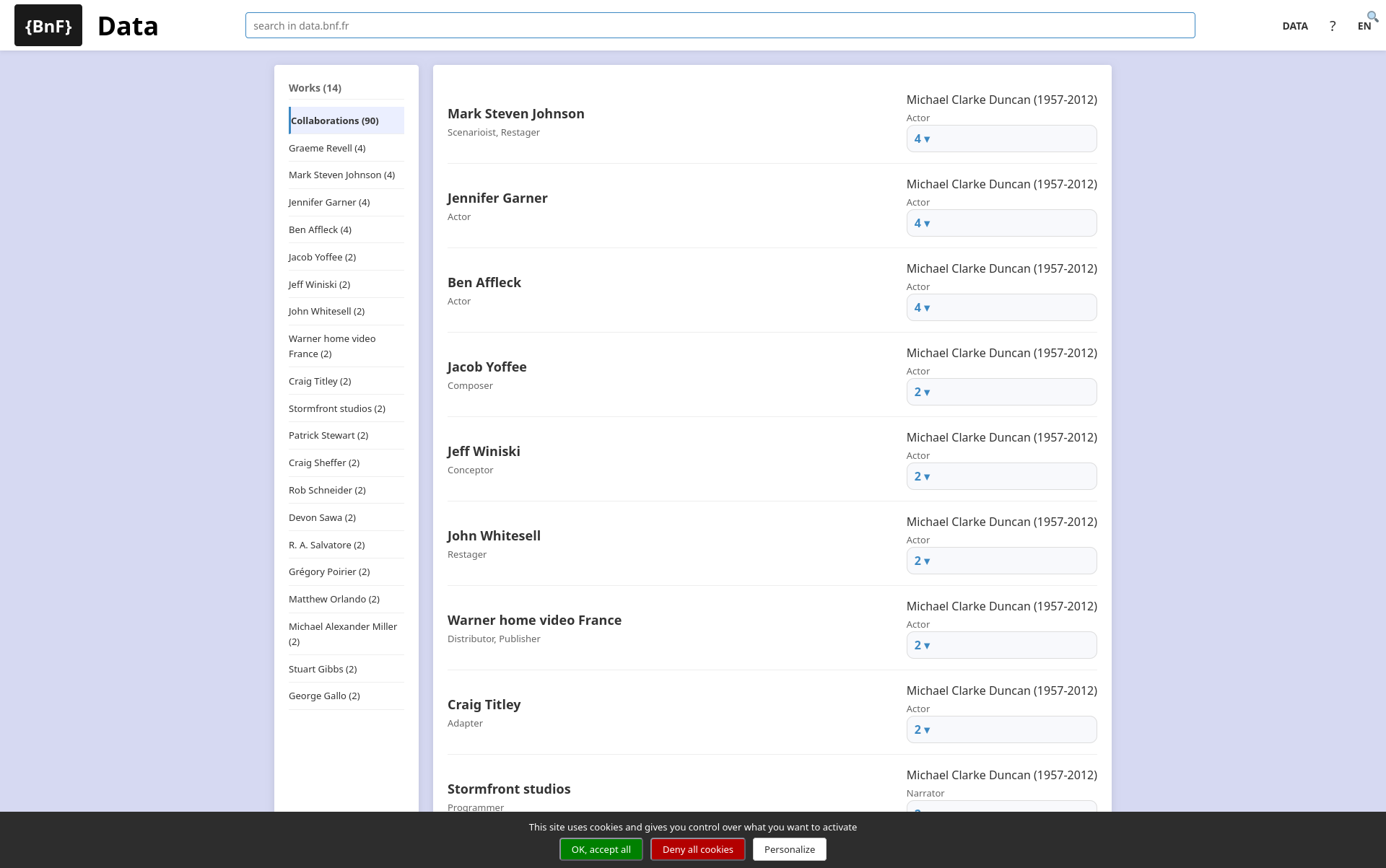} &
\rubriccell{Main content container framing}{styling\_visual}{The central results area in the reference is a large flat light-gray panel with thin borders and no card shadow, while the current attempt uses a smaller white card with rounded corners and shadow. Expand the main panel vertically and horizontally, remove the shadow/radius, and use a subtle border so the list feels like a full-width table region.}
& \rubriccell{Correct count selector styling and layout}{styling\_visual}{%
The count selectors (e.g., '4 v') in the main content are currently rendered as rounded rectangular buttons with borders and hover effects, which do not match the target UI. In the target, they appear as small, compact inline elements with a blue number and a downward arrow inside a minimal white circle with a thin border. Update the '.count-selector' CSS to remove the background, border, and padding, and instead use a small circular container with a thin border and blue text.%
} \\

\midrule

Qwen3.5-9B & 
\imgbox{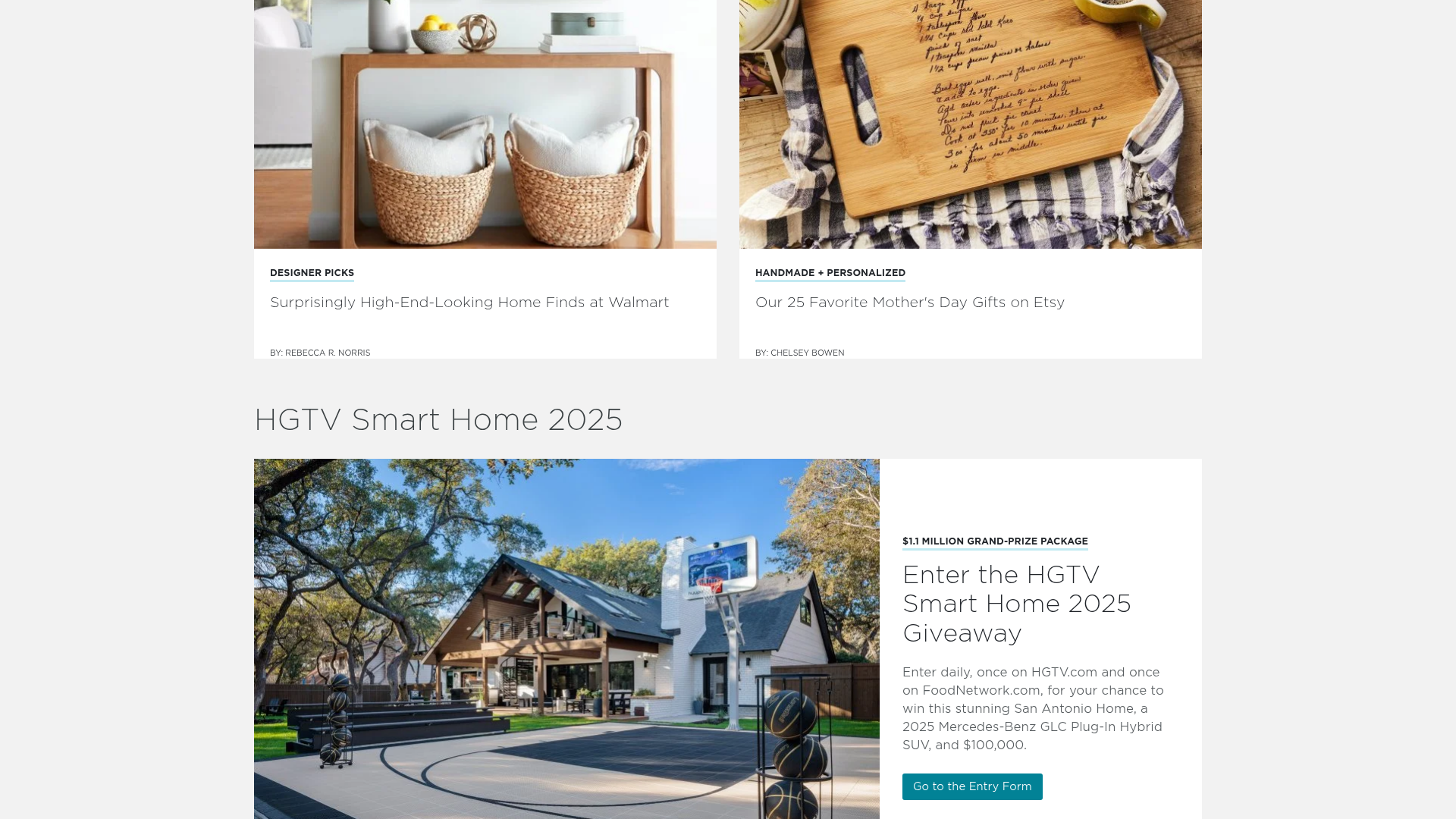} &
\imgbox{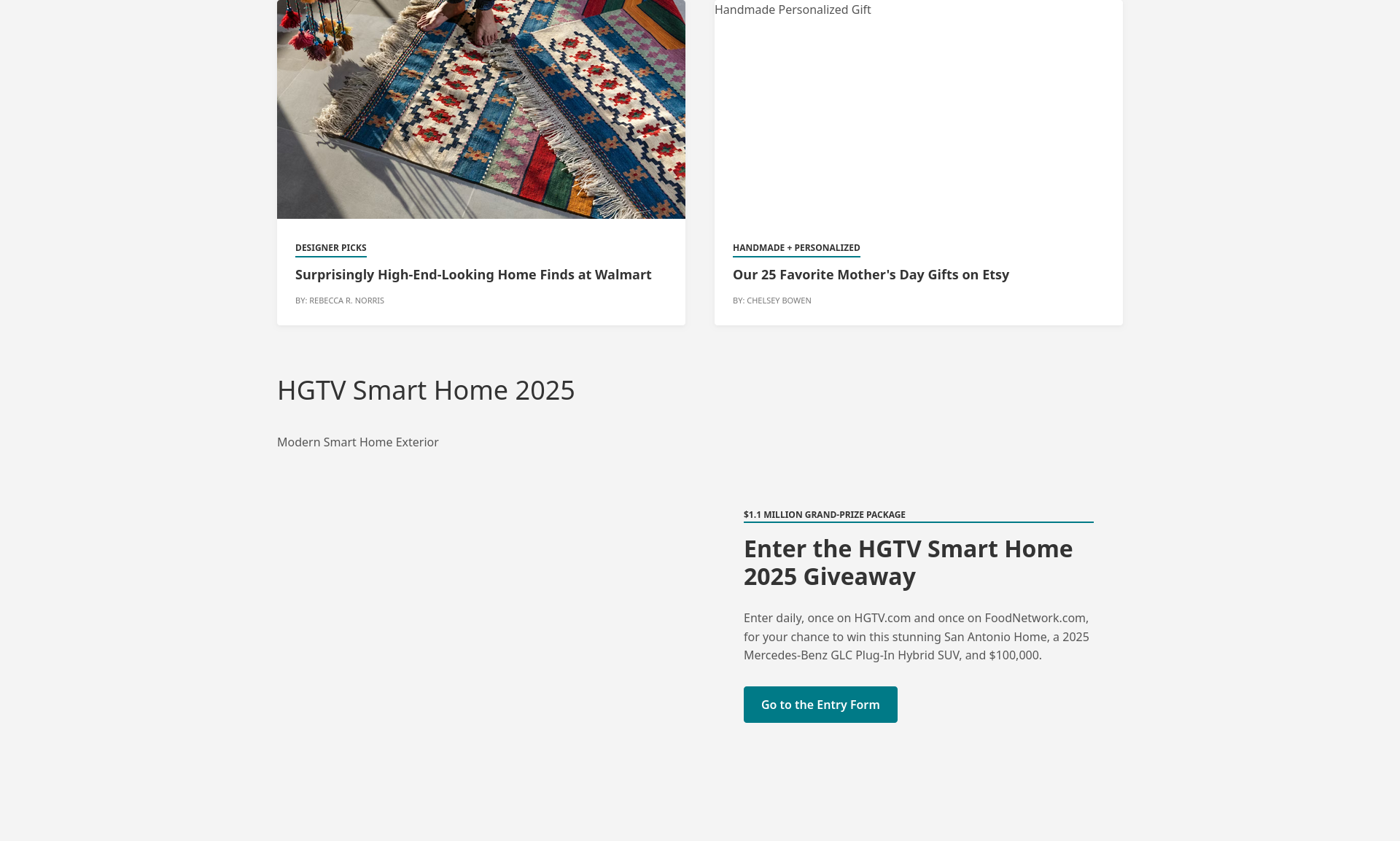} &
\rubriccell{Missing hero image in the lower-left feature area}{completeness}{%
The large Smart Home photo is absent in the rendered attempt, leaving a blank area with only the image alt text visible. Ensure the lower section uses a reliably loading image and places it as a large left column visual, matching the reference where the house photo fills most of the left half of the feature row.%
}
& \rubriccell{Incorrect Category Label Styling}{styling\_visual}{%
In the target screenshot, the category labels (e.g., 'DESIGNER PICKS') are bold, uppercase sans-serif text with a thin, light blue underline. The current attempt applies a thick, dark teal border-bottom to these labels, which creates a heavy, incorrect visual style compared to the subtle underline in the reference.%
} \\

\midrule

Qwen3.6-35B-A3B & 
\imgbox{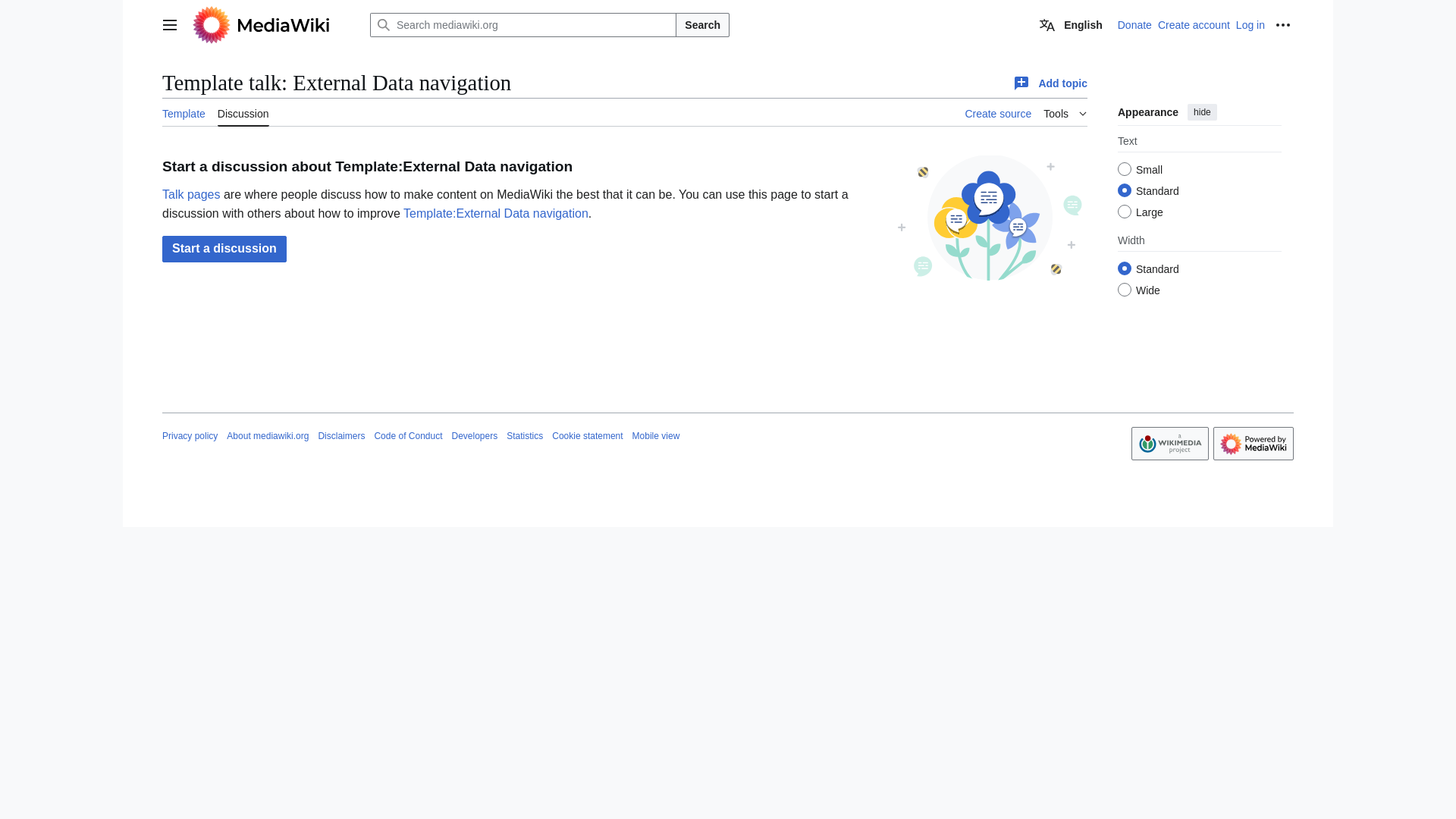} &
\imgbox{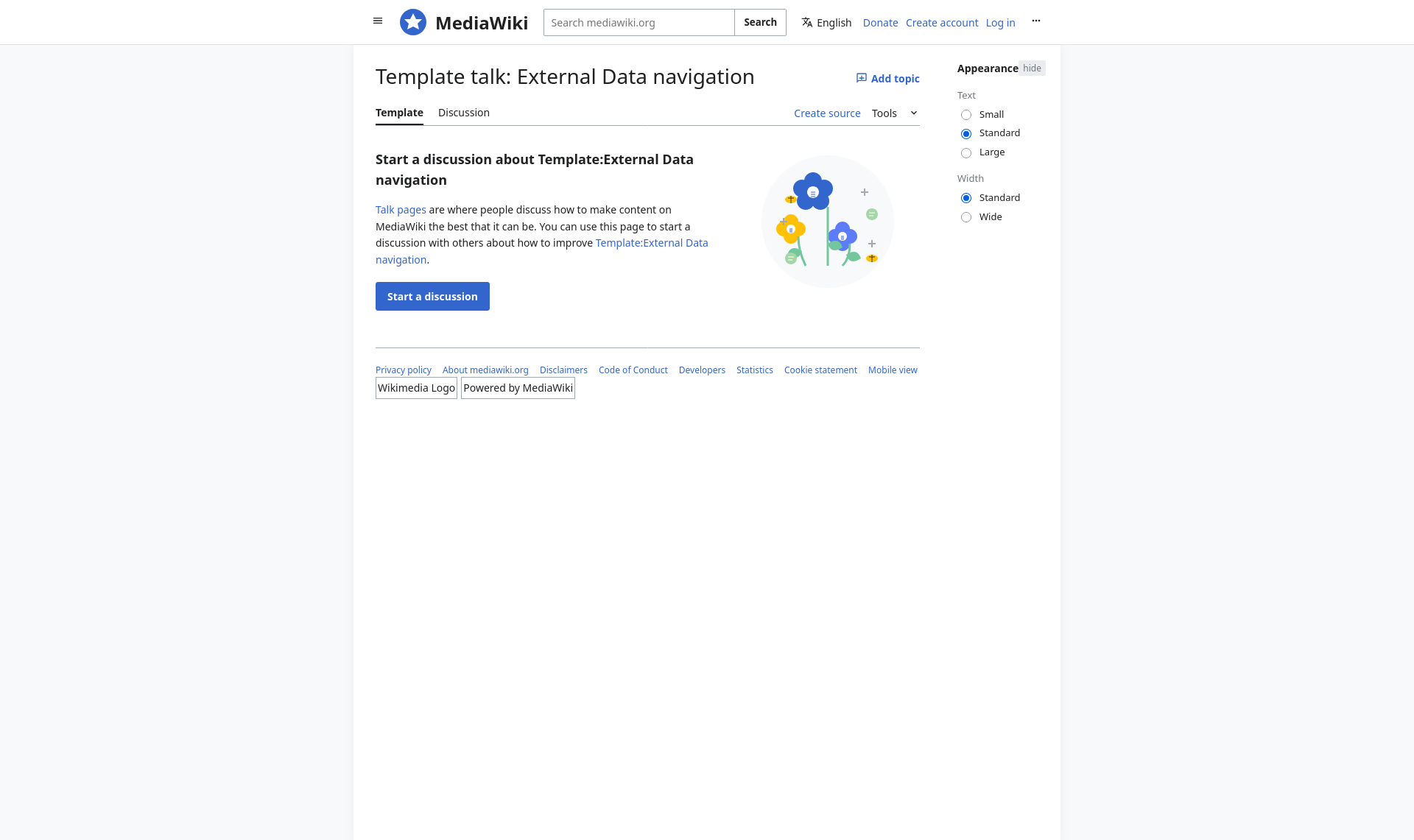} &
\rubriccell{Full-width centered page shell}{layout\_geometry}{%
The current page is constrained to a narrow 960px column and leaves a large empty gray area on the right, while the reference uses a much wider centered white canvas that spans most of the viewport. Increase the main page wrapper width to roughly 1100--1140px, center it within the light gray background, and keep the content and right appearance panel inside that single wide white area.%
}
& \rubriccell{Tab Active State Styling}{styling\_visual}{%
The 'Template' tab is currently styled as the active tab with a bold font and bottom border. In the target screenshot, the 'Discussion' tab is the active one. The active class should be moved to 'Discussion' and 'Template' should be inactive.%
} \\

\bottomrule
\end{tabular}
\renewcommand{\arraystretch}{1.0}
\endgroup
\vspace{-2mm}
\caption{Selected qualitative examples comparing GPT-generated and self-generated rubrics for Qwen self-evolution.
For each Qwen code improver, we show the target screenshot, the initial rendered output, and the selected rubric used for the next refinement step.
Rubrics from GPT-5.4 tend to identify higher-level visual-repair targets, such as page framing, missing major visual regions, or global layout structure.
In contrast, self-generated rubrics often focus on more local implementation details, such as selector styling, underline thickness, or active-tab states.
}
\label{tab:appendix-case-qwen-gpt-rubrics-examples}
\end{table*}

%% file: latex/sections/x_prompt_for_generation.tex
\section{Prompts for Self-Evolution}
\label{appendix:prompt-for-rubse}

We include the detailed prompt for naïve self-evolution in Table~\ref{tab:appendix-prompt-selfEvolve}. For \ours, the prompts for rubric generation, rubric selection, and rubric-guided single-step self-evolution are provided in Tables~\ref{tab:appendix-prompt-rubrics-gen}, \ref{tab:appendix-prompt-rubrics-select}, and \ref{tab:appendix-prompt-rubrics-improve}, respectively.

\begin{table}[!h]
\begin{tcolorbox}[colback=gray!5!white,colframe=gray!75!black,title=Naïve Self-Evolution]

\small

You are an expert web developer specializing in HTML and CSS.

\vspace{0.2em}

The user will provide:

\vspace{0.15em}

\begin{itemize}[leftmargin=*,topsep=2pt,itemsep=-1pt]
    \item A target webpage screenshot (ground-truth UI).
    \item A screenshot/render of a previous HTML+CSS attempt.
    \item The previous attempt's HTML+CSS source.
\end{itemize}

\vspace{0.2em}

Your task is to improve the previous HTML+CSS so that the rendered page matches the target screenshot more closely.

\vspace{0.2em}

Output format requirements:

\vspace{0.15em}

\begin{enumerate}[leftmargin=*,topsep=2pt,itemsep=-1pt]
    \item First, output a \texttt{<think>...</think>} block.
    \begin{itemize}[leftmargin=*,topsep=1pt,itemsep=-1pt]
        \item Compare the previous attempt against the target screenshot.
        \item Identify mismatches and what to change: layout, spacing, typography, colors, alignment, and missing elements.
        \item This is internal analysis; do not include any code here.
    \end{itemize}
    \item Then, output exactly one \verb|```html```| code block.
    \begin{itemize}[leftmargin=*,topsep=1pt,itemsep=-1pt]
        \item The code block must contain a single complete HTML document.
        \item Include all CSS inside \texttt{<style>} tags.
        \item Do not use JavaScript.
        \item Do not reference any external files, libraries, images, or fonts.
        \item Use a fixed 1920$\times$1080 viewport/aspect ratio. Set the root layout to exactly 1920px wide and 1080px tall, and avoid responsive rules that change this aspect ratio.
    \end{itemize}
\end{enumerate}

\vspace{0.2em}

\textbf{Critical:} Do not output anything other than the \texttt{<think>} block followed immediately by the \verb|```html```| block. No extra text before, between, or after them.

\vspace{0.2em}

The first image is the target screenshot, the second image is the previous rendered UI. The HTML text is the previous attempt to improve.

\end{tcolorbox}
\vspace{-4mm}
\caption{Prompt for naïve self-evolution.}
\label{tab:appendix-prompt-selfEvolve}
\end{table}

\begin{table}[!h]
\begin{tcolorbox}[colback=gray!5!white,colframe=gray!75!black,title=\ours: Rubric Generation]

\small

You are an expert UI evaluator and web developer.

\vspace{0.2em}

The user will provide:

\vspace{0.15em}

\begin{itemize}[leftmargin=*,topsep=2pt,itemsep=-1pt]
    \item A target webpage screenshot (ground-truth UI).
    \item A screenshot/render of a previous HTML+CSS attempt.
    \item The previous attempt's HTML+CSS source.
    \item Optionally, a JSON array of previously-addressed rubrics from earlier rounds; these are history items to avoid.
\end{itemize}

\vspace{0.2em}

Your task is to produce a list of actionable, example-specific rubrics that identify the most important ways the HTML/CSS should be improved so its rendered page matches the target screenshot more closely. If rubric history is provided, generate only new rubrics for remaining mismatches and do not restate, rephrase, or slightly vary historical rubrics.

\vspace{0.2em}

Each rubric must be assigned one type from this fixed set:

\vspace{0.15em}

\begin{itemize}[leftmargin=*,topsep=2pt,itemsep=-1pt]
    \item \texttt{layout\_geometry}: page structure, major regions, alignment, relative positioning, and element sizes.
    \item \texttt{spacing\_density}: padding, margins, gaps, line-heights, and overall visual density.
    \item \texttt{typography\_text}: font sizes, font weights, text hierarchy, text placement, and line breaks.
    \item \texttt{styling\_visual}: colors, backgrounds, borders, corner radius, shadows, dividers, and button or card styling.
    \item \texttt{completeness}: missing visible UI elements, hallucinated components, and extraneous elements.
\end{itemize}

\vspace{0.2em}

Rubric design requirements:

\vspace{0.15em}

\begin{itemize}[leftmargin=*,topsep=2pt,itemsep=-1pt]
    \item Each rubric must be concrete, checkable, and specific to this example.
    \item Clearly describe what is currently mismatched and how it should be corrected.
    \item Ground every rubric in observable evidence from the provided images and HTML.
    \item Avoid generic or reusable UI advice.
    \item Produce 3--8 rubrics, ordered roughly by importance.
\end{itemize}

\vspace{0.2em}

Output only a single JSON array. Each element must be a JSON object with exactly three keys: \texttt{title}, \texttt{description}, and \texttt{type}. Do not include commentary, markdown, code blocks, or text outside the JSON array.

\end{tcolorbox}
\vspace{-4mm}
\caption{Prompt for rubric generation in \ours.}
\label{tab:appendix-prompt-rubrics-gen}
\end{table}

\begin{table}[!h]
\begin{tcolorbox}[colback=gray!5!white,colframe=gray!75!black,title=\ours Rubric Selection]

\small

You are an expert UI evaluator and web developer.

\vspace{0.2em}

The user will provide:

\vspace{0.15em}

\begin{itemize}[leftmargin=*,topsep=2pt,itemsep=-1pt]
    \item A target webpage screenshot (ground-truth UI).
    \item A screenshot/render of a previous HTML+CSS attempt.
    \item The previous attempt's HTML+CSS source.
    \item A JSON array of candidate rubrics, each with \texttt{title}, \texttt{description}, and \texttt{type}.
\end{itemize}

\vspace{0.2em}

Your task is to select the single most critical rubric from the candidate list: the one that, if addressed, would produce the greatest improvement in visual similarity between the rendered page and the target screenshot.

\vspace{0.2em}

Selection criteria:

\vspace{0.15em}

\begin{itemize}[leftmargin=*,topsep=2pt,itemsep=-1pt]
    \item Prioritize issues that are visually prominent and immediately apparent.
    \item Prefer rubrics where fixing the issue requires a targeted, well-scoped HTML/CSS change.
    \item Consider visual impact relative to implementation effort.
    \item Do not select a rubric that is already well-resolved in the previous attempt.
\end{itemize}

\vspace{0.2em}

Output only a single JSON object with exactly three keys: \texttt{title}, \texttt{description}, and \texttt{type}. The object must be copied exactly from the candidate list. Do not include commentary, markdown, code blocks, or text outside the JSON object.

\end{tcolorbox}
\vspace{-4mm}
\caption{Prompt for rubric selection in \ours.}
\label{tab:appendix-prompt-rubrics-select}
\end{table}


\begin{table}[!h]
\begin{tcolorbox}[colback=gray!5!white,colframe=gray!75!black,title=\ours: Rubric-Guided Improvement]

\small

You are an expert web developer specializing in HTML and CSS.

\vspace{0.2em}

The user will provide:

\vspace{0.15em}

\begin{itemize}[leftmargin=*,topsep=2pt,itemsep=-1pt]
    \item A target webpage screenshot (ground-truth UI).
    \item A screenshot/render of a previous HTML+CSS attempt.
    \item The previous attempt's HTML+CSS source.
    \item A JSON object containing the selected improvement rubric.
\end{itemize}

\vspace{0.2em}

Your task is to improve the previous HTML+CSS so that the rendered page matches the target screenshot more closely.

\vspace{0.2em}

Output format requirements:

\vspace{0.15em}

\begin{enumerate}[leftmargin=*,topsep=2pt,itemsep=-1pt]
    \item First, output a \texttt{<think>...</think>} block.
    \begin{itemize}[leftmargin=*,topsep=1pt,itemsep=-1pt]
        \item Use the provided rubric as a checklist.
        \item Compare the previous attempt against the target screenshot.
        \item Identify the top mismatches and propose concrete HTML/CSS changes.
        \item This is internal analysis; do not include any code here.
    \end{itemize}
    \item Then, output exactly one \verb|```html```| code block.
    \begin{itemize}[leftmargin=*,topsep=1pt,itemsep=-1pt]
        \item The code block must contain a single complete HTML document.
        \item Include all CSS inside \texttt{<style>} tags.
        \item Do not use JavaScript.
        \item Do not reference any external files, libraries, images, or fonts.
        \item Use a fixed 1920$\times$1080 viewport/aspect ratio. Set the root layout to exactly 1920px wide and 1080px tall, and avoid responsive rules that change this aspect ratio.
    \end{itemize}
\end{enumerate}

\vspace{0.2em}

\textbf{Critical:} Do not output anything other than the \texttt{<think>} block followed immediately by the \verb|```html```| block. No extra text before, between, or after them.

\end{tcolorbox}
\vspace{-4mm}
\caption{Prompt for single-step rubric-guided code improvement in \ours.}
\label{tab:appendix-prompt-rubrics-improve}
\end{table}

%% file: latex/sections/x_additional_discussion.tex
\section{Additional Discussion}
\label{appendix:addition_discussion}

\paragraph{Relationship to General-Purpose Coding Agents.}
Modern general-purpose coding agents, such as Claude Code~\citep{claude_code}, Codex~\citep{openai_codex}, and OpenHands~\citep{wang2025openhands}, provide execution environments that define actions available to the model, including file access, command execution, screenshot inspection, and iterative code editing. RubSE operates at a different and complementary level: it defines an explicit, model-agnostic policy that specifies how rendered feedback is structured, prioritized, and retained to guide visual repair through its typed rubric taxonomy and \textsc{Evolve--Select--History} procedure. In our experiments, we implement it through constrained, role-separated calls to both open-source and frontier models, without requiring a general-purpose agent harness. This design isolates the policy-level contribution: rubric-guided repair is compared with unstructured self-evolution under the same model and execution setting.

\paragraph{Artifact Use and Licensing.}
We use UI2Code under its MIT license and Design2Code under its ODC-By license. These benchmarks consist of publicly available UI screenshots or rendered webpages; we do not collect new user data. For models, we use the Qwen3-VL series under its Apache-2.0 license, while proprietary frontier models, including GPT and Claude, are accessed through their respective API/service terms. All artifacts are used only for academic evaluation, and we do not redistribute third-party datasets, model weights, or proprietary model outputs.

\paragraph{Documentation of Artifacts.}
We report the benchmark and model details in Section~\ref{sec:experiment-settings}. These benchmarks provide screenshots as visual inputs for VLMs to generate HTML/CSS for webpage reconstruction. Since our evaluation focuses on UI screenshots and webpage-to-code generation, demographic attributes and linguistic phenomena are not applicable.

%% file: acl_latex.bbl
\begin{thebibliography}{63}
\providecommand{\natexlab}[1]{#1}

\bibitem[{Aky{\"u}rek et~al.(2025)Aky{\"u}rek, Gosai, Zhang, Gupta, Jeong, Gunjal, Rabbani, Mazzone, Randolph, Meymand et~al.}]{akyurek2025prbench}
Afra~Feyza Aky{\"u}rek, Advait Gosai, Chen Bo~Calvin Zhang, Vipul Gupta, Jaehwan Jeong, Anisha Gunjal, Tahseen Rabbani, Maria Mazzone, David Randolph, Mohammad~Mahmoudi Meymand, et~al. 2025.
\newblock Prbench: Large-scale expert rubrics for evaluating high-stakes professional reasoning.
\newblock \emph{arXiv preprint arXiv:2511.11562}.

\bibitem[{{Anthropic}(2025{\natexlab{a}})}]{claude_code}
{Anthropic}. 2025{\natexlab{a}}.
\newblock Claude code.
\newblock \url{https://code.claude.com/docs}.

\bibitem[{{Anthropic}(2025{\natexlab{b}})}]{anthropic2025claudesonnet45}
{Anthropic}. 2025{\natexlab{b}}.
\newblock Introducing claude sonnet 4.5.
\newblock \url{https://www.anthropic.com/news/claude-sonnet-4-5}.

\bibitem[{Arora et~al.(2025)Arora, Wei, Hicks, Bowman, Qui{\~n}onero-Candela, Tsimpourlas, Sharman, Shah, Vallone, Beutel et~al.}]{arora2025healthbench}
Rahul~K Arora, Jason Wei, Rebecca~Soskin Hicks, Preston Bowman, Joaquin Qui{\~n}onero-Candela, Foivos Tsimpourlas, Michael Sharman, Meghan Shah, Andrea Vallone, Alex Beutel, et~al. 2025.
\newblock Healthbench: Evaluating large language models towards improved human health.
\newblock \emph{arXiv preprint arXiv:2505.08775}.

\bibitem[{Azzolini et~al.(2025)Azzolini, Bai, Brandon, Cao, Chattopadhyay, Chen, Chu, Cui, Diamond, Ding et~al.}]{azzolini2025cosmos}
Alisson Azzolini, Junjie Bai, Hannah Brandon, Jiaxin Cao, Prithvijit Chattopadhyay, Huayu Chen, Jinju Chu, Yin Cui, Jenna Diamond, Yifan Ding, et~al. 2025.
\newblock Cosmos-reason1: From physical common sense to embodied reasoning.
\newblock \emph{arXiv preprint arXiv:2503.15558}.

\bibitem[{Bai et~al.(2025)Bai, Cai, Chen, Chen, Chen, Cheng, Deng, Ding, Gao, Ge et~al.}]{bai2025qwen3}
Shuai Bai, Yuxuan Cai, Ruizhe Chen, Keqin Chen, Xionghui Chen, Zesen Cheng, Lianghao Deng, Wei Ding, Chang Gao, Chunjiang Ge, et~al. 2025.
\newblock Qwen3-vl technical report.
\newblock \emph{arXiv preprint arXiv:2511.21631}.

\bibitem[{Beltramelli(2018)}]{beltramelli2018pix2code}
Tony Beltramelli. 2018.
\newblock pix2code: Generating code from a graphical user interface screenshot.
\newblock In \emph{Proceedings of the ACM SIGCHI symposium on engineering interactive computing systems}, pages 1--6.

\bibitem[{Bhathal and Gupta(2025)}]{bhathal2025websight}
Tanvir Bhathal and Asanshay Gupta. 2025.
\newblock Websight: A vision-first architecture for robust web agents.
\newblock \emph{arXiv preprint arXiv:2508.16987}.

\bibitem[{Cal{\`o} and De~Russis(2026)}]{calo2026webui}
Tommaso Cal{\`o} and Luigi De~Russis. 2026.
\newblock Webui-95: A large-scale dataset of normalized web interfaces via ui-to-code generation.
\newblock In \emph{Proceedings of the Extended Abstracts of the 2026 CHI Conference on Human Factors in Computing Systems}, pages 1--5.

\bibitem[{Chen et~al.(2025{\natexlab{a}})Chen, Li, Wang, Jiang, Liu, Lu, Huang, Byeon, Le, Ehrlich et~al.}]{chen2026eagle}
Guo Chen, Zhiqi Li, Shihao Wang, Jindong Jiang, Yicheng Liu, Lidong Lu, De-An Huang, Wonmin Byeon, Matthieu Le, Max Ehrlich, et~al. 2025{\natexlab{a}}.
\newblock Eagle 2.5: Boosting long-context post-training for frontier vision-language models.
\newblock \emph{Advances in Neural Information Processing Systems}, 38:91077--91100.

\bibitem[{Chen et~al.(2023)Chen, Xiong, Wu, Liu, Hu, Chen, Chen, Liu, and Huang}]{chen2023gpt}
Ruibo Chen, Tianyi Xiong, Yihan Wu, Guodong Liu, Zhengmian Hu, Lichang Chen, Yanshuo Chen, Chenxi Liu, and Heng Huang. 2023.
\newblock Gpt-4 vision on medical image classification--a case study on covid-19 dataset.
\newblock \emph{arXiv preprint arXiv:2310.18498}.

\bibitem[{Chen et~al.(2025{\natexlab{b}})Chen, Ding, Zhang, Chen, Du, Sun, and Chen}]{chen2025designcoder}
Yunnong Chen, Shixian Ding, YingYing Zhang, Wenkai Chen, Jinzhou Du, Lingyun Sun, and Liuqing Chen. 2025{\natexlab{b}}.
\newblock Designcoder: Hierarchy-aware and self-correcting ui code generation with large language models.
\newblock \emph{arXiv preprint arXiv:2506.13663}.

\bibitem[{Chen et~al.(2024)Chen, Wu, Wang, Su, Chen, Xing, Zhong, Zhang, Zhu, Lu et~al.}]{chen2024internvl}
Zhe Chen, Jiannan Wu, Wenhai Wang, Weijie Su, Guo Chen, Sen Xing, Muyan Zhong, Qinglong Zhang, Xizhou Zhu, Lewei Lu, et~al. 2024.
\newblock Internvl: Scaling up vision foundation models and aligning for generic visual-linguistic tasks.
\newblock In \emph{Proceedings of the IEEE/CVF conference on computer vision and pattern recognition}, pages 24185--24198.

\bibitem[{Deng et~al.(2026)Deng, Yao, and Zhang}]{deng2026visrefiner}
Jie Deng, Kaichun Yao, and Libo Zhang. 2026.
\newblock Visrefiner: Learning from visual differences for screenshot-to-code generation.
\newblock \emph{arXiv preprint arXiv:2602.05998}.

\bibitem[{Ge et~al.(2025)Ge, Liu, Ye, Li, and Wang}]{ge2025advancing}
Tong Ge, Yashu Liu, Jieping Ye, Tianyi Li, and Chao Wang. 2025.
\newblock Advancing vision-language models in front-end development via data synthesis.
\newblock \emph{arXiv preprint arXiv:2503.01619}.

\bibitem[{Gou et~al.(2024)Gou, Shao, Gong, Yang, Duan, Chen et~al.}]{gou2024critic}
Zhibin Gou, Zhihong Shao, Yeyun Gong, Yujiu Yang, Nan Duan, Weizhu Chen, et~al. 2024.
\newblock Critic: Large language models can self-correct with tool-interactive critiquing.
\newblock In \emph{International Conference on Learning Representations}, volume 2024, pages 57734--57811.

\bibitem[{Gui et~al.(2025{\natexlab{a}})Gui, Li, Wan, Shi, Zhang, Chen, Su, Chen, Wu, Zhou et~al.}]{gui2025webcode2m}
Yi~Gui, Zhen Li, Yao Wan, Yemin Shi, Hongyu Zhang, Bohua Chen, Yi~Su, Dongping Chen, Siyuan Wu, Xing Zhou, et~al. 2025{\natexlab{a}}.
\newblock Webcode2m: A real-world dataset for code generation from webpage designs.
\newblock In \emph{Proceedings of the ACM on Web Conference 2025}, pages 1834--1845.

\bibitem[{Gui et~al.(2025{\natexlab{b}})Gui, Li, Zhang, Wang, Lv, Jiang, Liu, Chen, Wan, Zhang et~al.}]{gui2025latcoder}
Yi~Gui, Zhen Li, Zhongyi Zhang, Guohao Wang, Tianpeng Lv, Gaoyang Jiang, Yi~Liu, Dongping Chen, Yao Wan, Hongyu Zhang, et~al. 2025{\natexlab{b}}.
\newblock Latcoder: Converting webpage design to code with layout-as-thought.
\newblock In \emph{Proceedings of the 31st ACM SIGKDD Conference on Knowledge Discovery and Data Mining V. 2}, pages 721--732.

\bibitem[{Gui et~al.(2025{\natexlab{c}})Gui, Wan, Li, Zhang, Chen, Zhang, Su, Chen, Zhou, Jiang et~al.}]{gui2025uicopilot}
Yi~Gui, Yao Wan, Zhen Li, Zhongyi Zhang, Dongping Chen, Hongyu Zhang, Yi~Su, Bohua Chen, Xing Zhou, Wenbin Jiang, et~al. 2025{\natexlab{c}}.
\newblock Uicopilot: Automating ui synthesis via hierarchical code generation from webpage designs.
\newblock In \emph{Proceedings of the ACM on Web Conference 2025}, pages 1846--1855.

\bibitem[{Gunjal et~al.(2025)Gunjal, Wang, Lau, Nath, He, Liu, and Hendryx}]{gunjal2025rubrics}
Anisha Gunjal, Anthony Wang, Elaine Lau, Vaskar Nath, Yunzhong He, Bing Liu, and Sean Hendryx. 2025.
\newblock Rubrics as rewards: Reinforcement learning beyond verifiable domains.
\newblock \emph{arXiv preprint arXiv:2507.17746}.

\bibitem[{Guo et~al.(2025)Guo, Yang, Zhang, Song, Wang, Zhu, Xu, Zhang, Ma, Bi et~al.}]{guo2025deepseek}
Daya Guo, Dejian Yang, Haowei Zhang, Junxiao Song, Peiyi Wang, Qihao Zhu, Runxin Xu, Ruoyu Zhang, Shirong Ma, Xiao Bi, et~al. 2025.
\newblock Deepseek-r1: Incentivizing reasoning capability in llms via reinforcement learning.
\newblock \emph{arXiv preprint arXiv:2501.12948}.

\bibitem[{Huang et~al.(2023)Huang, Gu, Hou, Wu, Wang, Yu, and Han}]{huang2023large}
Jiaxin Huang, Shixiang Gu, Le~Hou, Yuexin Wu, Xuezhi Wang, Hongkun Yu, and Jiawei Han. 2023.
\newblock Large language models can self-improve.
\newblock In \emph{Proceedings of the 2023 conference on empirical methods in natural language processing}, pages 1051--1068.

\bibitem[{Hurst et~al.(2024)Hurst, Lerer, Goucher, Perelman, Ramesh, Clark, Ostrow, Welihinda, Hayes, Radford et~al.}]{hurst2024gpt}
Aaron Hurst, Adam Lerer, Adam~P Goucher, Adam Perelman, Aditya Ramesh, Aidan Clark, AJ~Ostrow, Akila Welihinda, Alan Hayes, Alec Radford, et~al. 2024.
\newblock Gpt-4o system card.
\newblock \emph{arXiv preprint arXiv:2410.21276}.

\bibitem[{Jiang et~al.(2025)Jiang, Zheng, Wan, Han, Wang, Lyu, and Yue}]{jiang2025screencoder}
Yilei Jiang, Yaozhi Zheng, Yuxuan Wan, Jiaming Han, Qunzhong Wang, Michael~R Lyu, and Xiangyu Yue. 2025.
\newblock Screencoder: Advancing visual-to-code generation for front-end automation via modular multimodal agents.
\newblock \emph{arXiv preprint arXiv:2507.22827}.

\bibitem[{Khattab et~al.(2023)Khattab, Singhvi, Maheshwari, Zhang, Santhanam, Vardhamanan, Haq, Sharma, Joshi, Moazam et~al.}]{khattab2023dspy}
Omar Khattab, Arnav Singhvi, Paridhi Maheshwari, Zhiyuan Zhang, Keshav Santhanam, Sri Vardhamanan, Saiful Haq, Ashutosh Sharma, Thomas~T Joshi, Hanna Moazam, et~al. 2023.
\newblock Dspy: Compiling declarative language model calls into self-improving pipelines.
\newblock \emph{arXiv preprint arXiv:2310.03714}.

\bibitem[{Li et~al.(2024)Li, Zhang, Guo, Zhang, Li, Zhang, Zhang, Zhang, Li, Liu et~al.}]{li2024llava}
Bo~Li, Yuanhan Zhang, Dong Guo, Renrui Zhang, Feng Li, Hao Zhang, Kaichen Zhang, Peiyuan Zhang, Yanwei Li, Ziwei Liu, et~al. 2024.
\newblock Llava-onevision: Easy visual task transfer.
\newblock \emph{arXiv preprint arXiv:2408.03326}.

\bibitem[{Lin et~al.(2025)Lin, Zhou, Zhao, Wan, Ma, Gao, and Li}]{lin2025webuibench}
Zhiyu Lin, Zhengda Zhou, Zhiyuan Zhao, Tianrui Wan, Yilun Ma, Junyu Gao, and Xuelong Li. 2025.
\newblock Webuibench: a comprehensive benchmark for evaluating multimodal large language models in webui-to-code.
\newblock In \emph{Findings of the Association for Computational Linguistics: ACL 2025}, pages 15780--15797.

\bibitem[{Liu et~al.(2026)Liu, Chen, Chen, Xiong, Zheng, and Huang}]{liu2026prepare}
Chenxi Liu, Yanshuo Chen, Ruibo Chen, Tianyi Xiong, Tong Zheng, and Heng Huang. 2026.
\newblock Prepare reasoning language models for multi-agent debate with self-debate reinforcement learning.
\newblock \emph{arXiv preprint arXiv:2601.22297}.

\bibitem[{Liu et~al.(2025{\natexlab{a}})Liu, Xiong, Chen, Chen, Wu, Guo, Zhou, and Huang}]{liu2025modality}
Chenxi Liu, Tianyi Xiong, Yanshuo Chen, Ruibo Chen, Yihan Wu, Junfeng Guo, Tianyi Zhou, and Heng Huang. 2025{\natexlab{a}}.
\newblock Modality-balancing preference optimization of large multimodal models by adversarial negative mining.
\newblock \emph{arXiv preprint arXiv:2506.08022}.

\bibitem[{Liu et~al.(2023)Liu, Li, Wu, and Lee}]{liu2023visual}
Haotian Liu, Chunyuan Li, Qingyang Wu, and Yong~Jae Lee. 2023.
\newblock Visual instruction tuning.
\newblock \emph{Advances in neural information processing systems}, 36:34892--34916.

\bibitem[{Liu et~al.(2025{\natexlab{b}})Liu, Xu, Yu, Hong, Yang, Zhao, and Wang}]{liu2025openrubrics}
Tianci Liu, Ran Xu, Tony Yu, Ilgee Hong, Carl Yang, Tuo Zhao, and Haoyu Wang. 2025{\natexlab{b}}.
\newblock Openrubrics: Towards scalable synthetic rubric generation for reward modeling and llm alignment.
\newblock \emph{arXiv preprint arXiv:2510.07743}.

\bibitem[{Ni et~al.(2025{\natexlab{a}})Ni, Cai, Chen, Liang, Lyu, Deng, Zou, Nie, Yuan, Yue et~al.}]{ni2025viscoder2}
Yuansheng Ni, Songcheng Cai, Xiangchao Chen, Jiarong Liang, Zhiheng Lyu, Jiaqi Deng, Kai Zou, Ping Nie, Fei Yuan, Xiang Yue, et~al. 2025{\natexlab{a}}.
\newblock Viscoder2: Building multi-language visualization coding agents.
\newblock \emph{arXiv preprint arXiv:2510.23642}.

\bibitem[{Ni et~al.(2025{\natexlab{b}})Ni, Nie, Zou, Yue, and Chen}]{ni2025viscoder}
Yuansheng Ni, Ping Nie, Kai Zou, Xiang Yue, and Wenhu Chen. 2025{\natexlab{b}}.
\newblock Viscoder: Fine-tuning llms for executable python visualization code generation.
\newblock \emph{arXiv preprint arXiv:2506.03930}.

\bibitem[{{OpenAI}(2025{\natexlab{a}})}]{openai_codex}
{OpenAI}. 2025{\natexlab{a}}.
\newblock Introducing codex.
\newblock \url{https://openai.com/index/introducing-codex/}.

\bibitem[{{OpenAI}(2025{\natexlab{b}})}]{openai2025gpt52}
{OpenAI}. 2025{\natexlab{b}}.
\newblock Introducing gpt-5.2.
\newblock \url{https://openai.com/index/introducing-gpt-5-2/}.

\bibitem[{{OpenAI}(2026)}]{openai2026gpt54}
{OpenAI}. 2026.
\newblock Introducing gpt-5.4.
\newblock \url{https://openai.com/index/introducing-gpt-5-4/}.

\bibitem[{Rezaei et~al.(2025)Rezaei, Vacareanu, Wang, Wang, Liu, He, and Aky{\"u}rek}]{rezaei2025online}
MohammadHossein Rezaei, Robert Vacareanu, Zihao Wang, Clinton Wang, Bing Liu, Yunzhong He, and Afra~Feyza Aky{\"u}rek. 2025.
\newblock Online rubrics elicitation from pairwise comparisons.
\newblock \emph{arXiv preprint arXiv:2510.07284}.

\bibitem[{Shao et~al.(2025)Shao, Asai, Shen, Ivison, Kishore, Zhuo, Zhao, Park, Finlayson, Sontag et~al.}]{shao2025dr}
Rulin Shao, Akari Asai, Shannon~Zejiang Shen, Hamish Ivison, Varsha Kishore, Jingming Zhuo, Xinran Zhao, Molly Park, Samuel~G Finlayson, David Sontag, et~al. 2025.
\newblock Dr tulu: Reinforcement learning with evolving rubrics for deep research.
\newblock \emph{arXiv preprint arXiv:2511.19399}.

\bibitem[{Si et~al.(2025)Si, Zhang, Li, Yang, Liu, and Yang}]{si2025design2code}
Chenglei Si, Yanzhe Zhang, Ryan Li, Zhengyuan Yang, Ruibo Liu, and Diyi Yang. 2025.
\newblock Design2code: Benchmarking multimodal code generation for automated front-end engineering.
\newblock In \emph{Proceedings of the 2025 Conference of the Nations of the Americas Chapter of the Association for Computational Linguistics: Human Language Technologies (Volume 1: Long Papers)}, pages 3956--3974.

\bibitem[{Soselia et~al.(2023)Soselia, Saifullah, and Zhou}]{soselia2023learning}
Davit Soselia, Khalid Saifullah, and Tianyi Zhou. 2023.
\newblock Learning ui-to-code reverse generator using visual critic without rendering.
\newblock \emph{arXiv preprint arXiv:2305.14637}.

\bibitem[{Srivastava et~al.(2025)Srivastava, Singh, Madhavan, Patil, Addepalli, Suggala, Aravamudhan, Sharma, Laha, Raghuveer et~al.}]{srivastava2025robust}
Pragya Srivastava, Harman Singh, Rahul Madhavan, Gandharv Patil, Sravanti Addepalli, Arun Suggala, Rengarajan Aravamudhan, Soumya Sharma, Anirban Laha, Aravindan Raghuveer, et~al. 2025.
\newblock Robust reward modeling via causal rubrics.
\newblock \emph{arXiv preprint arXiv:2506.16507}.

\bibitem[{Wang et~al.(2025{\natexlab{a}})Wang, Li, Song, Xu, Tang, Zhuge, Pan, Song, Li, Singh et~al.}]{wang2025openhands}
Xingyao Wang, Boxuan Li, Yufan Song, Frank~F Xu, Xiangru Tang, Mingchen Zhuge, Jiayi Pan, Yueqi Song, Bowen Li, Jaskirat Singh, et~al. 2025{\natexlab{a}}.
\newblock Openhands: An open platform for ai software developers as generalist agents.
\newblock In \emph{International Conference on Learning Representations}, volume 2025, pages 65882--65919.

\bibitem[{Wang et~al.(2025{\natexlab{b}})Wang, Li, Yang, Zhang, Liu, Xiong, and Huang}]{wang2025llava}
Xiyao Wang, Chunyuan Li, Jianwei Yang, Kai Zhang, Bo~Liu, Tianyi Xiong, and Furong Huang. 2025{\natexlab{b}}.
\newblock Llava-critic-r1: Your critic model is secretly a strong policy model.
\newblock \emph{arXiv preprint arXiv:2509.00676}.

\bibitem[{Wang et~al.(2026)Wang, Zang, Han, Bu, Zhou, Jin, and Wang}]{wang2026unified}
Yibin Wang, Yuhang Zang, Feng Han, Jiazi Bu, Yujie Zhou, Cheng Jin, and Jiaqi Wang. 2026.
\newblock Unified personalized reward model for vision generation.
\newblock \emph{arXiv preprint arXiv:2602.02380}.

\bibitem[{Wang et~al.(2025{\natexlab{c}})Wang, Jung, Lu, Diao, Evans, Zeng, Molchanov, Choi, Kautz, and Dong}]{wang2025profbench}
Zhilin Wang, Jaehun Jung, Ximing Lu, Shizhe Diao, Ellie Evans, Jiaqi Zeng, Pavlo Molchanov, Yejin Choi, Jan Kautz, and Yi~Dong. 2025{\natexlab{c}}.
\newblock Profbench: Multi-domain rubrics requiring professional knowledge to answer and judge.
\newblock \emph{arXiv preprint arXiv:2510.18941}.

\bibitem[{Wu et~al.(2024)Wu, Schoop, Leung, Barik, Bigham, and Nichols}]{wu2024uicoder}
Jason Wu, Eldon Schoop, Alan Leung, Titus Barik, Jeffrey~P Bigham, and Jeffrey Nichols. 2024.
\newblock Uicoder: Finetuning large language models to generate user interface code through automated feedback.
\newblock In \emph{Proceedings of the 2024 Conference of the North American Chapter of the Association for Computational Linguistics: Human Language Technologies (Volume 1: Long Papers)}, pages 7511--7525.

\bibitem[{Xiong et~al.(2025{\natexlab{a}})Xiong, Ge, Li, Zhang, Kulkarni, Wang, He, Zhu, Liu, Chen et~al.}]{xiong2025multi}
Tianyi Xiong, Yi~Ge, Ming Li, Zuolong Zhang, Pranav Kulkarni, Kaishen Wang, Qi~He, Zeying Zhu, Chenxi Liu, Ruibo Chen, et~al. 2025{\natexlab{a}}.
\newblock Multi-crit: Benchmarking multimodal judges on pluralistic criteria-following.
\newblock \emph{arXiv preprint arXiv:2511.21662}.

\bibitem[{Xiong et~al.(2026)Xiong, Wang, Liu, Dong, Li, Huang, Kautz, and Yu}]{xiong2026phycritic}
Tianyi Xiong, Shihao Wang, Guilin Liu, Yi~Dong, Ming Li, Heng Huang, Jan Kautz, and Zhiding Yu. 2026.
\newblock Phycritic: Multimodal critic models for physical ai.
\newblock \emph{arXiv preprint arXiv:2602.11124}.

\bibitem[{Xiong et~al.(2025{\natexlab{b}})Xiong, Wang, Guo, Ye, Fan, Gu, Huang, and Li}]{xiong2025llava}
Tianyi Xiong, Xiyao Wang, Dong Guo, Qinghao Ye, Haoqi Fan, Quanquan Gu, Heng Huang, and Chunyuan Li. 2025{\natexlab{b}}.
\newblock Llava-critic: Learning to evaluate multimodal models.
\newblock In \emph{Proceedings of the Computer Vision and Pattern Recognition Conference}, pages 13618--13628.

\bibitem[{Xu et~al.(2025)Xu, Yang, Hong, Pan, Fan, Wang, Gu, Xu, and Tang}]{xu2025webvia}
Mingde Xu, Zhen Yang, Wenyi Hong, Lihang Pan, Xinyue Fan, Yan Wang, Xiaotao Gu, Bin Xu, and Jie Tang. 2025.
\newblock Webvia: A web-based vision-language agentic framework for interactive and verifiable ui-to-code generation.
\newblock \emph{arXiv preprint arXiv:2511.06251}.

\bibitem[{Yang et~al.(2026)Yang, Hong, Xu, Fan, Wang, Cheng, Gu, and Tang}]{yang2025ui2code}
Zhen Yang, Wenyi Hong, Mingde Xu, Xinyue Fan, Weihan Wang, Jiele Cheng, Xiaotao Gu, and Jie Tang. 2026.
\newblock Ui2code$^{N}$: Ui-to-code generation as interactive visual optimization.
\newblock In \emph{Proceedings of the 43rd International Conference on Machine Learning}.

\bibitem[{Yifei et~al.(2025)Yifei, Chang, Malaviya, and Yatskar}]{yifei2025researchqa}
Li~S Yifei, Allen Chang, Chaitanya Malaviya, and Mark Yatskar. 2025.
\newblock Researchqa: Evaluating scholarly question answering at scale across 75 fields with survey-mined questions and rubrics.
\newblock \emph{arXiv preprint arXiv:2509.00496}.

\bibitem[{Yu et~al.(2025)Yu, Wang, Wang, Huang, Ma, He, Cai, Chen, Huang, Zhao et~al.}]{yu2025minicpm}
Tianyu Yu, Zefan Wang, Chongyi Wang, Fuwei Huang, Wenshuo Ma, Zhihui He, Tianchi Cai, Weize Chen, Yuxiang Huang, Yuanqian Zhao, et~al. 2025.
\newblock Minicpm-v 4.5: Cooking efficient mllms via architecture, data, and training recipe.
\newblock \emph{arXiv preprint arXiv:2509.18154}.

\bibitem[{Yuan et~al.(2025)Yuan, Mang, Chen, Wan, Liu, Xu, Huang, Wang, Jiao, and He}]{yuan2025curing}
Youliang Yuan, Qiuyang Mang, Jingbang Chen, Hong Wan, Xiaoyuan Liu, Junjielong Xu, Jen-tse Huang, Wenxuan Wang, Wenxiang Jiao, and Pinjia He. 2025.
\newblock Curing miracle steps in llm mathematical reasoning with rubric rewards.
\newblock \emph{arXiv preprint arXiv:2510.07774}.

\bibitem[{Yun et~al.(2024)Yun, Lin, Thushara, Bhat, Wang, Jiang, Deng, Wang, Tao, Li et~al.}]{yun2024web2code}
Sukmin Yun, Haokun Lin, Rusiru Thushara, Mohammad~Q Bhat, Yongxin Wang, Zutao Jiang, Mingkai Deng, Jinhong Wang, Tianhua Tao, Junbo Li, et~al. 2024.
\newblock Web2code: A large-scale webpage-to-code dataset and evaluation framework for multimodal llms.
\newblock \emph{Advances in neural information processing systems}, 37:112134--112157.

\bibitem[{Zhang et~al.(2025{\natexlab{a}})Zhang, Li, Xu, Liu, Liu, Zhou, Deng, Wu, Huang, Li et~al.}]{zhang2025artifactsbench}
Chenchen Zhang, Yuhang Li, Can Xu, Jiaheng Liu, Ao~Liu, Changzhi Zhou, Ken Deng, Dengpeng Wu, Guanhua Huang, Kejiao Li, et~al. 2025{\natexlab{a}}.
\newblock Artifactsbench: Bridging the visual-interactive gap in llm code generation evaluation.
\newblock \emph{arXiv preprint arXiv:2507.04952}.

\bibitem[{Zhang et~al.(2025{\natexlab{b}})Zhang, Lei, Li, Wang, Liu, Yang, Li, Wang, Yang, Wu et~al.}]{zhang2025critic}
Di~Zhang, Jingdi Lei, Junxian Li, Xunzhi Wang, Yujie Liu, Zonglin Yang, Jiatong Li, Weida Wang, Suorong Yang, Jianbo Wu, et~al. 2025{\natexlab{b}}.
\newblock Critic-v: Vlm critics help catch vlm errors in multimodal reasoning.
\newblock In \emph{Proceedings of the IEEE/CVF Conference on Computer Vision and Pattern Recognition}, pages 9050--9061.

\bibitem[{Zhang et~al.(2025{\natexlab{c}})Zhang, Hu, Upasani, Ma, Hong, Kamanuru, Rainton, Wu, Ji, Li et~al.}]{zhang2025agentic}
Qizheng Zhang, Changran Hu, Shubhangi Upasani, Boyuan Ma, Fenglu Hong, Vamsidhar Kamanuru, Jay Rainton, Chen Wu, Mengmeng Ji, Hanchen Li, et~al. 2025{\natexlab{c}}.
\newblock Agentic context engineering: Evolving contexts for self-improving language models.
\newblock \emph{arXiv preprint arXiv:2510.04618}.

\bibitem[{Zhang et~al.(2023)Zhang, Lu, Wang, Yan, Yan, Qin, Wang, Yan, Wang, and Petzold}]{zhang2023gpt}
Xinlu Zhang, Yujie Lu, Weizhi Wang, An~Yan, Jun Yan, Lianke Qin, Heng Wang, Xifeng Yan, William~Yang Wang, and Linda~Ruth Petzold. 2023.
\newblock Gpt-4v (ision) as a generalist evaluator for vision-language tasks.
\newblock \emph{arXiv preprint arXiv:2311.01361}.

\bibitem[{Zhao et~al.(2025{\natexlab{a}})Zhao, Zhang, Lin, Li, Wu, Zhang, and Wei}]{zhao2025pyvision}
Shitian Zhao, Haoquan Zhang, Shaoheng Lin, Ming Li, Qilong Wu, Kaipeng Zhang, and Chen Wei. 2025{\natexlab{a}}.
\newblock Pyvision: Agentic vision with dynamic tooling.
\newblock \emph{arXiv preprint arXiv:2507.07998}.

\bibitem[{Zhao et~al.(2025{\natexlab{b}})Zhao, Jiang, Zeng, Chen, Qiu, Huang, Zhong, Zheng, Cao, and Ma}]{zhao2025vincicoder}
Xuanle Zhao, Deyang Jiang, Zhixiong Zeng, Lei Chen, Haibo Qiu, Jing Huang, Yufeng Zhong, Liming Zheng, Yilin Cao, and Lin Ma. 2025{\natexlab{b}}.
\newblock Vincicoder: Unifying multimodal code generation via coarse-to-fine visual reinforcement learning.
\newblock \emph{arXiv preprint arXiv:2511.00391}.

\bibitem[{Zheng et~al.(2023)Zheng, Chiang, Sheng, Zhuang, Wu, Zhuang, Lin, Li, Li, Xing et~al.}]{zheng2023judging}
Lianmin Zheng, Wei-Lin Chiang, Ying Sheng, Siyuan Zhuang, Zhanghao Wu, Yonghao Zhuang, Zi~Lin, Zhuohan Li, Dacheng Li, Eric Xing, et~al. 2023.
\newblock Judging llm-as-a-judge with mt-bench and chatbot arena.
\newblock \emph{Advances in neural information processing systems}, 36:46595--46623.

\bibitem[{Zhou et~al.(2025)Zhou, Zhao, Hou, Sun, Chen, and Wang}]{zhou2025declarui}
Ting Zhou, Yanjie Zhao, Xinyi Hou, Xiaoyu Sun, Kai Chen, and Haoyu Wang. 2025.
\newblock Declarui: Bridging design and development with automated declarative ui code generation.
\newblock \emph{Proceedings of the ACM on Software Engineering}, 2(FSE):219--241.

\end{thebibliography}
